%% file: acl_latex.tex
\documentclass[11pt]{article}
\usepackage[preprint]{acl}
\usepackage{times,latexsym}
\usepackage[T1]{fontenc}
\usepackage[utf8]{inputenc}
\usepackage{microtype,inconsolata}
\usepackage{amsmath,amssymb,booktabs,array,graphicx,xcolor}
\usepackage{colortbl}
\usepackage{listings}
\usepackage{mmstyle}
\input{data/release/facts.tex}

\title{\raisebox{-0.32em}[0.95em][0.2em]{\includegraphics[height=1.4em]{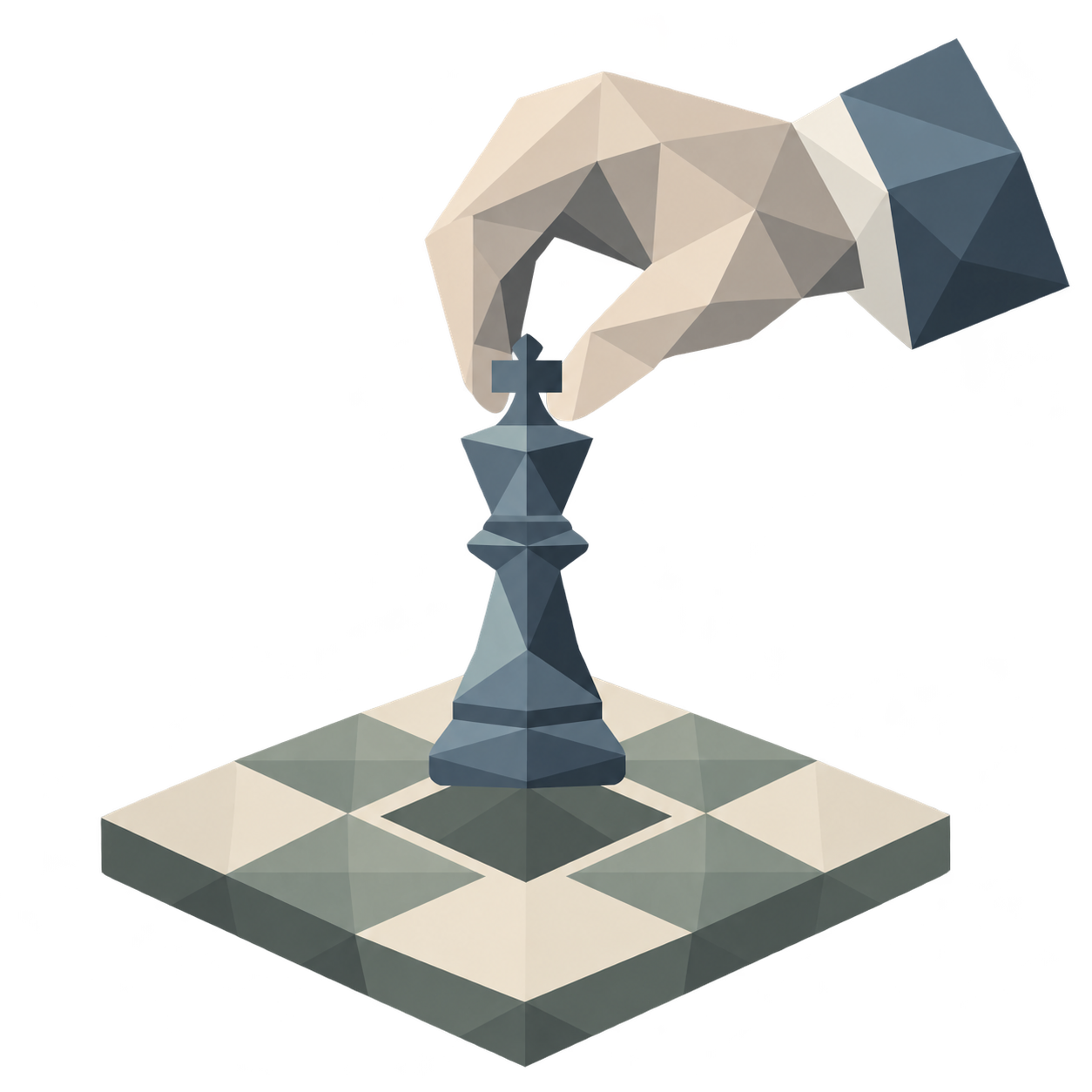}}\hspace{0.3em}MMDM: Measuring Massive Multitask Decision Making}
\author{Junhao Shen$^{1*}$ \quad Qing Zong$^{2*}$ \quad Yue Guo$^3$ \\ \textbf{Zhanqiu Zhang}$^{3\dagger}$ \quad \textbf{Yiwen Guo}$^{4\dagger}$ \quad \textbf{Hong Cheng}$^1$ \\
$^1$The Chinese University of Hong Kong \quad $^2$Hong Kong University of Science and Technology \\
$^3$LIGHTSPEED \quad $^4$Independent Researcher \\
\texttt{shen.junhao@outlook.com} \quad \texttt{qzong@cse.ust.hk} \\
\texttt{\{zqzhang27,guoyiwen89\}@gmail.com} \quad \texttt{hcheng@se.cuhk.edu.hk}\\
Project Page: \url{mmdm-benchmark.github.io}
}

\begin{document}
\makeatletter
\let\mmdmoriginalmaketitle\@maketitle
\def\@maketitle{\mmdmoriginalmaketitle\input{sections/teaser}}
\makeatother
\maketitle

\renewcommand\thefootnote{}
\footnotetext{$\dagger$: Corresponding author. $*$: Equal contribution.}
\footnotetext{}

\input{sections/abstract}
\input{sections/intro}

\input{sections/related_work}
\input{sections/benchmark}
\input{sections/experiment}
\input{sections/conclusion}
\input{sections/limitations}

\input{sections/ethics}
\bibliography{references}
\clearpage
\appendix
\setcounter{table}{0}
\setcounter{figure}{0}
\setcounter{equation}{0}
\renewcommand{\thetable}{A\arabic{table}}
\renewcommand\thefigure{A\arabic{figure}}
\renewcommand\theequation{A\arabic{equation}}
\raggedbottom
\makeatletter
\setlength{\@fptop}{0pt}
\setlength{\@fpsep}{18pt}
\setlength{\@fpbot}{0pt plus 1fil}
\setlength{\@dblfptop}{0pt}
\setlength{\@dblfpsep}{18pt}
\setlength{\@dblfpbot}{0pt plus 1fil}
\makeatother
\input{sections/appendix}
\input{sections/examples}

\end{document}

%% file: data/release/facts.tex
\newcommand{\NumExamples}{5,448}
\newcommand{\NumCategories}{16}

%% file: sections/teaser.tex
\begin{minipage}{\textwidth}
\centering
\includegraphics[width=\textwidth]{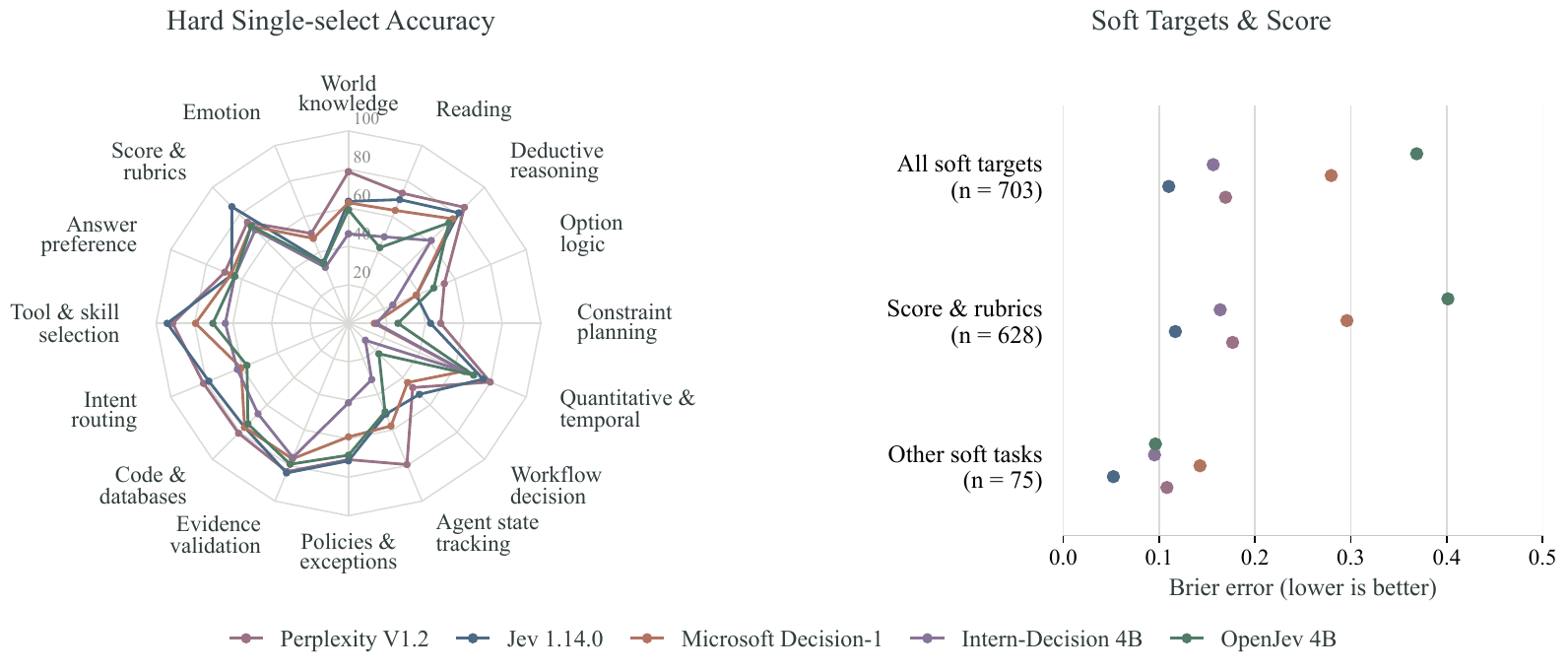}
\captionof{figure}{MMDM capability profiles for five representative decision models. The left panel shows Hard Single-select accuracy across 16 task categories (\%, higher is better). The right panel shows soft Brier error over all soft targets, soft Score targets, and other soft tasks (lower is better).}
\label{fig:teaser}\vspace{7pt}
\end{minipage}

%% file: sections/abstract.tex
\begin{abstract}
System-1 Decision Models directly make structured probabilistic decisions over query-specified options given a state in a single forward pass, without autoregressively generating reasoning trajectories or textual responses. 
Recent evaluations cover diverse tasks but often rely on existing benchmarks, where prior exposure to test material can obscure generalization.
Moreover, real-world decision making often involves long contexts, large option spaces, and complex logical dependencies, challenges that existing benchmarks do not adequately capture together.
Thus, we introduce \benchmarkname{} (Massive Multitask Decision Making), a benchmark of \NumExamples{} high-quality questions across \NumCategories{} task categories, combining newly constructed tasks with source-grounded recompositions and rewrites for single-select, multi-select, and soft-label decision making in fresh contexts. It covers realistic scenarios such as agent state tracking, tool and skill selection, preference judgments, and rubric scoring, with option spaces ranging from 2 to 1,024 and inputs extending to 64,817 tokens.
With \benchmarkname{}, we conduct a comprehensive quantitative evaluation of 28 decision models, showing that many perform well on common knowledge yet struggle with interdependent options, constrained decisions, and multi-select tasks. For example, Perplexity Decider-V1.2 reaches 78.75\% Hard Single-select Accuracy on world knowledge, but only 53.95\% on option logic and 47.95\% on constraint planning, while the best Hard Multi-select Accuracy across all models is 36.81\%. These results highlight joint reasoning, constraint-aware selection, and complete multi-select recovery as key priorities for decision-model development, with MMDM providing a testbed for measuring this progress.
\end{abstract}

%% file: sections/intro.tex
\section{Introduction}
System-1 Decision Models make structured probabilistic decisions over query-specified options given a textual state in a single forward pass~\citep{typesafeai2026jev,microsoft2026decision}, which support diverse applications, including text and intent classification, model and tool selection, and rubric-based quality scoring~\citep{ling2026jev}.
Therefore, a rigorous and comprehensive evaluation of these models must account for the diversity and complexities of decision making in real-world scenarios.

Recent benchmarks have assessed decision models across diverse tasks and calibration criteria~\citep{duan2026jeval,solka2026decisionbench,standhartinger2026jevbench}. However, many existing evaluations draw heavily on widely circulated public benchmarks, making test-set contamination a central concern for benchmark validity~\citep{oren2024contamination}. Moreover, real-world decisions often combine long contexts, large option spaces, and complex logical dependencies. For example, a practically useful model needs to track the current agent state, determine whether an earlier approval remains valid, and select an eligible tool from a large catalog with heterogeneous prerequisites. These settings call for an evaluation that combines task breadth with new instances and explicit tests of these capabilities.

In this paper, we introduce \textbf{MMDM} (Massive Multitask Decision Making), a benchmark of \NumExamples{} questions for evaluating text-based System-1 Decision Models across realistic decision scenarios. It combines broad task coverage with explicit tests of context understanding, option discrimination, and logical consistency. Table~\ref{tab:benchmarkcomparison} compares its scope with existing decision benchmarks.
\input{tables/benchmark-comparison}

Specifically, MMDM evaluates \textbf{Hard Single-select}, \textbf{Hard Multi-select}, and \textbf{Soft-label} questions, requiring a unique correct choice, the complete valid set, or a prediction matching the reference distribution over outcomes. These answer types capture practical requirements such as choosing the next actions, identifying all eligible tools, and assigning probabilities to rubric scores under uncertainty. Decisions can depend on agent states, action prerequisites, or relationships among options. In benchmark construction, source answers and contextual evidence ground rewrites of states, queries and distractors to reduce memorization-induced score inflation~\citep{zhu2024inference,zhao2025mmlucf}.
We combine these adaptations with recompositions and newly constructed tasks. The resulting \NumCategories{} categories span knowledge and comprehension, logic and constraints, agent and workflow decisions, and preference and rubric judgments. With 2--1,024 options and complete inputs of up to 64,817 tokens, MMDM also tests these capabilities across varied context lengths and option spaces.

As shown in Figure~\ref{fig:teaser}, our evaluation of 28 decision models reveals substantial differences across capabilities. Models that perform well on world knowledge still struggle with interdependent options, decisions subject to constraints, and complete multi-select recovery. These capabilities matter in agent workflows, where choices interact, actions have prerequisites, and several tools or skills may be valid for the current goal. Specifically, Perplexity Decider-V1.2~\citep{perplexityai2026decider} achieves 78.75\% Hard Single-select accuracy on world knowledge, compared with 53.95\% on option logic and 47.95\% on constraint planning. WeMM-JEV 27B~\citep{wemmjevteam2026wemm} achieves the highest Hard Single-select accuracy of 76.00\%, while the highest Hard Multi-select accuracy is 36.81\%, highlighting the challenge of recovering all valid options. Model rankings also differ between hard-answer accuracy and soft-distribution error, making both necessary to characterize decision quality.

Overall, our contributions are threefold. First, we introduce a multitask decision benchmark that combines fresh decision instances with long contexts, large option spaces, and interdependent options. Second, we provide a unified evaluation protocol for single-select correctness, multi-select completeness, and soft-label distributional fidelity, together with an open dataset and toolkit. Third, our evaluation reveals capability gaps that are not captured by previous decision-model benchmarks, including failures in joint consistency and complete multi-select recovery, providing concrete directions for future model development.

%% file: tables/benchmark-comparison.tex
\begin{table*}[t]
\centering
\begingroup
\fontsize{9}{11}\selectfont
\setlength{\tabcolsep}{3pt}
\renewcommand{\arraystretch}{1.4}
\newcommand{\benchyes}{\textcolor[HTML]{597A68}{\ensuremath{\checkmark}}}
\newcommand{\benchno}{\textcolor[HTML]{AE6A63}{\ensuremath{\boldsymbol{\times}}}}
\newcommand{\benchns}{N/A}
\begin{tabular}{@{}>{\raggedright\arraybackslash}p{.36\textwidth}
>{\centering\arraybackslash}p{.09\textwidth}
>{\centering\arraybackslash}p{.12\textwidth}
>{\centering\arraybackslash}p{.105\textwidth}
>{\centering\arraybackslash}p{.105\textwidth}
>{\centering\arraybackslash}p{.07\textwidth}
>{\centering\arraybackslash}p{.07\textwidth}@{}}
\toprule
\rowcolor{exampleBlue}
\textbf{Benchmark} & \shortstack{\textbf{Max.}\\\textbf{options}} & \shortstack{\textbf{Context}\\\textbf{length}} & \shortstack{\textbf{New}\\\textbf{instances}} & \shortstack{\textbf{Multi-select}} & \shortstack{\textbf{Soft}\\\textbf{targets}} & \shortstack{\textbf{Score}\\\textbf{tasks}} \\
\midrule
\mbox{JEVal~\citep{duan2026jeval}} & 255 & 82,678 & \benchno & \benchno & \benchno & \benchyes \\
\mbox{DecisionBench 1.0~\citep{solka2026decisionbench}} & 255 & 35,435 & \benchno & \benchno & \benchno & \benchyes \\
\mbox{JevBench~\citep{standhartinger2026jevbench}} & \benchns & \benchns & \benchyes & \benchno & \benchyes & \benchyes \\
\mbox{DecideBench 1.1~\citep{yong2026decidebench}} & 6 & 229 & \benchyes & \benchno & \benchno & \benchno \\
\mbox{Jev Evaluation~\citep{deusser2026evaluating}} & 151 & \benchns & \benchno & \benchno & \benchno & \benchyes \\
\mbox{ReflexBench 1.0~\citep{navarro2026reflexbench}} & 255 & \benchns & \benchno & \benchno & \benchno & \benchyes \\
\midrule
\rowcolor{exampleSage}
\textbf{MMDM (ours)} & 1,024 & 64,817 & \benchyes & \benchyes & \benchyes & \benchyes \\
\bottomrule
\end{tabular}
\endgroup
\caption{Comparison of the main evaluation sets of existing decision benchmarks. Context length is the maximum complete input in Qwen3.5-4B tokens; N/A indicates an undisclosed exact value. New instances denotes newly authored or semantically recomposed questions beyond direct reuse or format-only adaptation of existing benchmarks. Multi-select evaluates the complete valid set from individual option rankings; soft targets use non-one-hot references. \textcolor[HTML]{597A68}{$\checkmark$}/\textcolor[HTML]{AE6A63}{$\times$}: the defined feature is documented/not documented in the main benchmark. MMDM combines all four properties with 1,024-way decisions and long inputs (Detailed distribution is shown in Table~\ref{fig:inputcoverage}).}
\label{tab:benchmarkcomparison}
\end{table*}

%% file: sections/related_work.tex
\section{Related Work}
\paragraph{System-1 Decision Models.}
System-1 Decision Models produce structured probabilities over request-defined options in a single inference step. 
Jev introduces a typed decision interface and Reinforcement Learning for Calibrated Decisions, making probability quality a central training objective~\citep{typesafeai2026jev}.
After this closed-source model, open implementations explore different ways to obtain these decisions from language-model backbones. SemIf reads logits for option tokens directly and supports shared-prefix reuse across queries~\citep{lee2026semif}. OpenJev learns a shared decision head over request-defined options, using supervised learning followed by reinforcement learning~\citep{shen2026openjev}. Intern-Decision and H2O-Lightning extend typed decisions to multimodal inputs~\citep{zhu2026intern,h2oai2026lightning}. Microsoft Decision-1 post-trains Qwen3.5-9B for single-pass scoring in routing, verification, and workflow control~\citep{microsoft2026decision}.
Together, these approaches develop efficient readout, task adaptation, and probability calibration. Their growing use in software and agent pipelines makes it essential to evaluate whether they preserve decision quality when the relevant evidence, prerequisites, and options interact.

\paragraph{Benchmarks for Decision Making.}
JevBench establishes an early comparison of typed decision quality, calibration, latency, and cost~\citep{standhartinger2026jevbench}. DecisionBench broadens applied coverage to 43 tasks with runtime-defined options~\citep{solka2026decisionbench}, while ReflexBench separated raw model quality from workflow-policy effects~\citep{navarro2026reflexbench}. DecideBench subsequently introduced 400 contrastive decisions to test sensitivity to answer-changing context~\citep{yong2026decidebench}. Jev Evaluation examined 37 existing datasets with frozen templates~\citep{deusser2026evaluating}, and JEVal expanded systematic evaluation across 36 datasets and ten domains~\citep{duan2026jeval}.
These tool kits improve coverage and comparability, but their main evaluations leave complete multi-select decisions underexplored. Additionally, broad source reuse also leaves a need for fresh decision instances~\cite{oren2024contamination}. 
Complementary resources target individual capabilities. AgentBench evaluates interactive agents~\citep{liu2024agentbench}. ToolBench evaluates tool-use tasks~\citep{qin2024toolllm}. LongBench and LongBench v2 studies long-context understanding in challenging realistic tasks~\citep{bai2024longbench,bai2025longbenchv2}. MuSiQue evaluates multi-hop evidence composition~\citep{trivedi2022musique}. They provide valuable capability tests, while a common assessment of option selection and probability allocation across these settings remains necessary. 
MMDM addresses this gap by combining fresh decision scenarios, long inputs, and large option spaces with single-select, multi-select, and soft-label objectives.

%% file: sections/benchmark.tex
\section{The MMDM Benchmark}
\label{sec:construction}
MMDM evaluates decision capabilities across diverse tasks through a shared state--query--options interface. First, we formalize the decision problems, answer types and evaluation objectives (Section~\ref{sec:formulation}). Then, we describe data collection and construction (Section~\ref{sec:routes}). Finally, we present task coverage and dataset statistics (Section~\ref{sec:coverage}).

\subsection{Task Formulation}
\label{sec:formulation}
\label{sec:softconstruction}
\label{sec:metrics}
\textbf{Decision Instance.}
A decision instance consists of a \textbf{state} $s$, a \textbf{query} $q$, and an ordered set of \textbf{options} $\mathcal{O}$. The state provides the evidence and current situation, the query specifies the decision to make, and each option $o_j=(a_j,c_j)$ contains a name $a_j$ and an optional criterion $c_j$ describing its meaning or applicability. A decision model $f_\theta$ maps these inputs to an option distribution
\begin{equation}
 x=(s,q,\mathcal{O}),\qquad p=f_\theta(x)\in\Delta^{K-1},
 \label{eq:decision}
\end{equation}
where $\mathcal{O}=(o_1,\ldots,o_K)$ and $\Delta^{K-1}$ is the probability simplex, so $p_j\geq0$ and $\sum_{j=1}^{K}p_j=1$. Options are provided in each request, allowing their names, meanings, and number to vary beyond a fixed output vocabulary. Criteria may refer to other options, requiring the model to interpret the option set jointly with the state and query.

\textbf{Answer Types.}
Each instance has a reference distribution $\pi\in\Delta^{K-1}$ with one of three answer semantics. We use $\ind[\cdot]$ for the indicator function.

\emph{Hard Single-select} specifies one correct option $y$, such as the only tool that meets a request's requirements. The model should assign this option the highest probability. The reference places all probability mass on it,
\begin{equation}
 \pi_j=\ind[j=y].
 \label{eq:hardtargets}
\end{equation}
\emph{Hard Multi-select} specifies a complete set $G$ of equally valid options, such as several tools that can each fulfill the same request. Every member must rank above every invalid option. Equal reference mass represents their equal validity,
\begin{equation}
 \pi_j=\frac{\ind[j\in G]}{|G|}.
 \label{eq:multitarget}
\end{equation}
\emph{Soft-label} specifies how likely each outcome is under the uncertainty described in the state. For example, a tool call may succeed or fail because the service is intermittently available. If the stated service conditions imply a 70\% success probability, the reference assigns 0.7 to \emph{success} and 0.3 to \emph{failure}. The model should match the probabilities of all outcomes, including less likely ones. For outcomes $\omega\in\Omega$ and a rule $r$ mapping each outcome to an option index, the reference is
\begin{equation}
 \pi_j=\sum_{\omega\in\Omega}P(\omega\mid x)\ind[r(\omega)=j].
 \label{eq:softtarget}
\end{equation}
Scoring tasks use ordered options with numeric values, such as levels of report completeness. A known score gives a hard single-select reference; uncertain outcomes give a soft distribution over levels. The expected score summarizes this distribution on the rubric's numeric scale. Appendix~\ref{app:softtargets} details label derivation.

\begin{samepage}
\textbf{Evaluation Metrics.} 
Per-question correctness for the two hard-answer tracks is
\begin{align}
 a_{\rm single}(p,y)&=\ind\!\left[p_y>\max_{j\ne y}p_j\right],\label{eq:singleacc}\\
 a_{\rm multi}(p,G)&=\ind\!\left[\min_{j\in G}p_j>\max_{j\notin G}p_j\right].\label{eq:multiacc}
\end{align}
\end{samepage}
Hard Single-select and Hard Multi-select Accuracy average these correctness values over their respective tracks, counting failed requests and boundary ties as incorrect. Higher accuracy is better.

Separation--Balance F-score (SB-F) combines separation from incorrect options with balanced support for the correct set. For $N=|G|$, let $\mu=N^{-1}\sum_{i\in G}p_i$ be the mean correct-option probability and $b=\max_{j\notin G}p_j$ the largest incorrect-option probability, with $b=0$ if every option is correct. The separation component is
\begin{equation}
 S(p,G)=\frac{1}{N}\sum_{i\in G}\frac{\max(p_i-b,0)}{\max(p_i,\mu)+b}.
 \label{eq:sbf-separation}
\end{equation}
A summand is zero when its denominator is zero. Each correct option earns credit only above $b$, and $\mu$ prevents negligible probability from earning full credit. Correct-set balance is
\begin{equation}
 B(p,G)=\frac{\left(\sum_{i\in G}p_i\right)^2}{N\sum_{i\in G}p_i^2},
 \label{eq:sbf-balance}
\end{equation}
with $B(p,G)=0$ when the correct set receives no mass. Their product defines
\begin{equation}
 \operatorname{SB\!-\!F}(p,G)=S(p,G)\,B(p,G).
 \label{eq:sbf}
\end{equation}
SB-F lies in $[0,1]$, with higher values better. 
We average per-question scores over all valid multi-select responses to show the performance. 

Soft Brier measures error across the full reference distribution. Score nMAE measures expected-score error relative to the rubric range, using the numeric levels $v_j$. These per-question errors are
\begin{align}
 \operatorname{Brier}(p,\pi)&=\sum_{j=1}^{K}(p_j-\pi_j)^2,\label{eq:brier}\\
 e_{\rm score}(p,\pi,v)&=\frac{\left|\sum_{j=1}^{K}v_j(p_j-\pi_j)\right|}{\max_j v_j-\min_j v_j}.\label{eq:scoreerror}
\end{align}
Soft Brier and Score nMAE average Equations~\eqref{eq:brier} and~\eqref{eq:scoreerror} over valid native responses in their respective tracks. Lower values are better for both. 

\subsection{Data Collection and Construction}
\label{sec:routes}
\textbf{Collection.} 
We select public datasets and agent trajectories that support a concrete decision with explicit options and a checkable answer. Sources span knowledge, reading, reasoning, preferences, code, and agent interactions. For example, QuALITY~\citep{pang2022quality} and LongBench v2~\citep{bai2025longbenchv2} supply document-grounded questions, BIRD~\citep{li2023bird} supplies database problems, and ComplexFuncBench~\citep{zhong2025complexfuncbench} and ToolTalk~\citep{farn2023tooltalk} supply interaction histories. Source questions and annotations provide the basis for adaptation and verification.

\textbf{Construction.} 
We combine source-grounded adaptation with semantic recomposition and newly constructed tasks. For rewritten source instances, we retain the original answers and supporting evidence while reformulating the state, query, and distractors. Semantic and option-quality checks select the resulting instances. This strategy builds on evidence that benchmark rewriting can mitigate memorization-induced score inflation~\citep{zhu2024inference,zhao2025mmlucf}. Recompositions introduce decision questions grounded in source databases or agent histories. Executable mechanisms construct interacting options, constrained plans, state transitions, and rubric outcomes with computable answers. Across all three routes, distractors are designed around identifiable failure modes, such as applying an expired rule, confusing entities, or overlooking the prerequisite.

\textbf{Validation and Selection.} 
We verify answers using source evidence, executable checks, and model review appropriate to each construction route. Five-model majority screening is followed by independent human expert review of borderline cases. Both stages apply the same quality standards for distractors, answer leakage, duplicates, and source overlap. An option-only probe masks the state and query; instances with unresolved shortcuts are excluded. Appendix~\ref{app:selection} details the review procedure and selection rules.

\subsection{Task Coverage and Dataset Statistics}
\label{sec:coverage}
\begin{figure}[t]
\centering\includegraphics[width=\columnwidth]{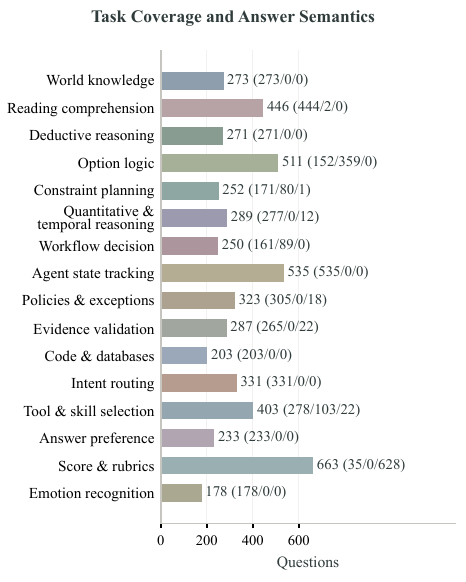}
\caption{Task coverage and answer semantics. The colors of the 16 task categories are fixed across figures. Labels show the total count followed by (Hard Single-select / Hard Multi-select / Soft-label) counts.}
\label{fig:distribution}
\end{figure}
MMDM contains 5,448 questions across 16 task categories, comprising 4,112 Hard Single-select, 633 Hard Multi-select, and 703 Soft-label questions. The 663 Score questions include 35 with hard levels and 628 with soft targets. The collection covers 62 registered source identifiers and 80 category--subcategory combinations. 
Option sets range from 2 to 1,024 options, including 94 questions with more than 255 options. Complete inputs include the state, query, option names, and criteria. We measure complete inputs with the Qwen3.5-4B tokenizer from the Qwen family~\citep{yang2025qwen3}. Their median length is 827 tokens, the 95th percentile is 13,318, and the maximum is 64,817. These ranges support evaluation from common short decisions to long-document reasoning and large-catalog selection.

The categories connect knowledge and language understanding with decisions about evidence, rules, actions, and assessments. They cover world knowledge and reading comprehension; deductive, quantitative, and option-dependent reasoning; agent state tracking, tool and skill selection, and workflows; and preference, rubric, and emotion judgments. Figure~\ref{fig:distribution} summarizes all categories and their answer types. 
Appendix~\ref{app:sources} further elaborate each category and option-count distributions; Appendix~\ref{app:examples} presents representative questions and model responses.

%% file: sections/experiment.tex
\section{Experiment}
\label{sec:evaluation}
\input{tables/current-overall}
\input{tables/current-category-all}

\subsection{Experimental Setup}
\label{sec:setup}
\textbf{Models.} We evaluate 28 System-1 decision models and use Random as a heuristic reference. These models return option probabilities directly through a decision-scoring head or a single decoding step, without generated reasoning or adaptive thinking. They include hosted APIs and locally deployed checkpoints. Each model uses its supported context window and option limit, preserving the capabilities available in its released checkpoint or API. Appendix~\ref{app:modelsettings} details model selection, deployment settings, and the Random reference.

\textbf{Protocol and Metrics.} The evaluator preserves the query and all option names and criteria, deterministically removing the middle of the state when the context budget is exceeded. Independent option sets exceeding a model's option cap are divided into groups that fit the cap. Each request scores all options in a group together. From each group, the highest-probability option advances for single-select, and up to $|G|$ highest-probability options advance for multi-select. This reduction repeats until the retained options fit one final request. Reference cardinality and gold identities are hidden from the model. Option-dependent and distributional tasks keep the entire option set without splitting; inputs exceeding the model's option cap count as capacity failures. Grouped outputs contribute to hard selection metrics; full-distribution metrics use native responses.
We use the metrics defined in Section~\ref{sec:formulation}. Hard accuracies retain every question in the corresponding track, including failed requests. SB-F, Soft Brier, and Score nMAE use native predictions. 
Appendix~\ref{app:protocol} specifies token budgets, grouping, and failure accounting.

\subsection{Overall Performance}
\label{sec:overall}

\finding{Similar single-select accuracy can accompany very different probability quality.} In Table~\ref{tab:overall}, WeMM-JEV and KnowLine differ by less than one percentage point in single-select accuracy. Yet KnowLine has nearly half the Soft Brier error and less than one-third the Score nMAE of WeMM-JEV. A model that chooses the right option frequently can therefore still be a poor fit for applications that use its probabilities or expected scores.

\finding{Complete multi-select identification and probability support remain weak.} The highest multi-select accuracy in Table~\ref{tab:overall}  is only 36.81\%, and every model has an SB-F below 0.20. This means that current models often fail to recover the complete set of valid options and to assign probability appropriately across them. Thus, improving multi-select decisions requires progress not only in finding all correct options, but also in representing their relative support reliably.

\finding{Expected-score accuracy and distribution fidelity favor different models.} 
Jev achieves the lowest Soft Brier, while KnowLine achieves the lowest Score nMAE. GPT-6 Luna Decisions also has a lower Score nMAE than Microsoft Decision-1 despite a higher Soft Brier. This difference arises because the expected score compresses an entire probability distribution into a single value, allowing probability errors at different score levels to cancel out. Therefore, reporting both metrics shows whether a model predicts a useful final score and whether the probabilities behind that score are themselves accurate.

\begin{figure*}[t]
\centering\includegraphics[width=0.95\textwidth]{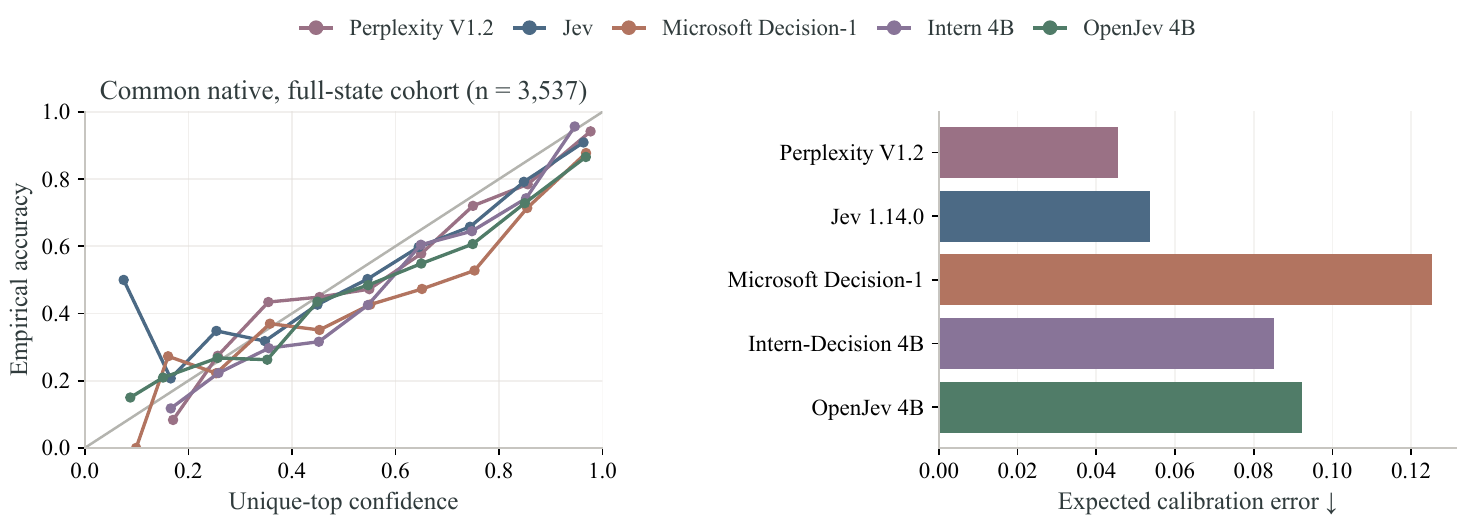}
\caption{Confidence calibration on 3,537 shared single-select questions with native, full-state predictions and unique maxima. The left panel groups predictions into ten equal-width confidence bins. The gray diagonal marks perfect calibration. Above it, accuracy exceeds confidence (underconfidence); below it, confidence exceeds accuracy (overconfidence). The right panel reports ECE ($\downarrow$) on the same questions.}
\label{fig:reliability}
\end{figure*}

\begin{figure*}[t]
\centering\includegraphics[width=0.95\textwidth]{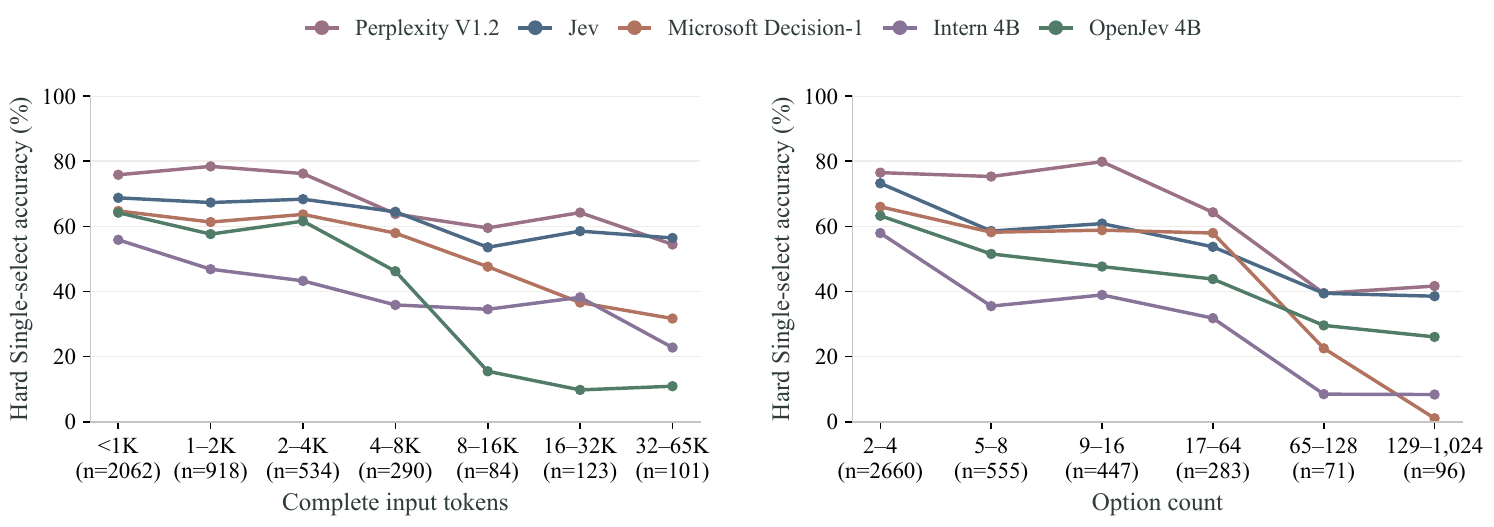}
\caption{Hard Single-select Accuracy across complete-input length and option-count bins. Counts refer to benchmark items, with failures included. Each bin retains its released task composition and model-specific context and option handling. The curves describe performance under these deployment conditions.}
\label{fig:capacity}
\end{figure*}

\input{tables/current-case-studies}

\subsection{Performance by Task Category}
\label{sec:capabilities}
\finding{Models struggle when several conditions must hold at once.} 
WeMM-JEV leads Deductive Reasoning in Table~\ref{tab:categoryall}, yet scores below 60\% on Option Logic, Constraint Planning, and Workflow Decision. KnowLine shows the same contrast, with 86.0\% on Deductive Reasoning but 54.6\% on Option Logic and 48.5\% on Constraint Planning. In these tasks, an option may look correct on its own but conflict with another option, a rule, or an earlier action. MMDM separates these requirements from deduction, exposing a weakness in joint consistency that the broad knowledge-and-reasoning score of JEVal does not isolate.

\finding{Similar overall accuracy can hide very different strengths across tasks.}  
Table~\ref{tab:categoryall} shows that WeMM-JEV performs better on world knowledge and emotion recognition, while KnowLine is stronger on workflow decisions and policy handling. Models with lower overall scores can still lead particular categories. Quyet achieves the highest accuracy on Evidence Validation, while Mapika leads Emotion Recognition. These differences show why category-level results matter when selecting a model for a specific workload, since a higher overall score does not guarantee better performance on the capabilities that workload actually requires.

\finding{Strong routing performance does not guarantee strong agent decision making.} 
GPT-6 Luna Decisions outperforms Jev and Microsoft Decision-1 on Intent Routing, but performs worse on Tool and Skill Selection, Agent State Tracking, and Workflow Decision. The largest gap appears in Workflow Decision, which requires the model to account for prerequisites and action order. This shows that routing correctly does not necessarily mean the model can also choose the right action from the current agent state or coordinate a multi-step workflow.

\subsection{Confidence Calibration}
\label{sec:calibration}
\finding{Overconfidence appears at different operating ranges.} 
Figure~\ref{fig:reliability} compares five representative models using reliability curves and expected calibration error (ECE), where lower is better. Intern-Decision is overconfident in the middle of its range. OpenJev is closer to the diagonal there, but its highest-confidence bin averages about 0.97 confidence and 0.87 accuracy. Thus, this difference matters in deployment because the same confidence threshold may be reliable for one model but overly aggressive for another one when making decision.

\finding{Higher selection accuracy does not ensure better calibration.} 
OpenJev has higher single-select accuracy than Intern-Decision in Table~\ref{tab:overall}, yet Figure~\ref{fig:reliability} gives it a larger ECE. Perplexity and Jev achieve the lowest ECE among the five representatives, while Microsoft Decision-1 has the largest error and remains below the diagonal through much of the moderate-to-high confidence range. The curves and aggregate errors together show why model selection should account for both how often a model is correct and how reliably its confidence reflects that correctness.

\subsection{Context and Option-Space Scaling}
\label{sec:capacity}
\finding{Model rankings change between long inputs and large option sets.} The two panels of Figure~\ref{fig:capacity} reveal different scaling profiles. OpenJev outperforms Intern-Decision in every option-count bin, but falls below it once complete inputs exceed 8K tokens. At 8K--16K tokens, their accuracies are 15.5\% and 34.5\%, respectively; on 129--1,024 options, they are 26.0\% and 8.3\%. This reversal distinguishes performance on long-input workloads from performance on large-option workloads. MMDM measures both axes because success on one does not predict the model ordering on the other.

\finding{Large catalogs expose differences hidden by compact choices.} In Figure~\ref{fig:capacity}, Jev and Microsoft Decision-1 differ by about seven percentage points on 2--4 options, but by more than 37 points on 129--1,024 options. Perplexity retains the highest accuracy in the largest option bin, although its advantage over Jev is much smaller there than on 9--16 options. Across the released workloads and the protocol in Section~\ref{sec:setup}, the curves reveal both severe losses and changing relative advantages as the option space grows. A single aggregate accuracy would conceal these differences in the scale of decisions each model can handle.

\subsection{Case Studies}
\label{sec:cases}
\finding{Models can accept a workflow or report that conflicts with other evidence.} 
In Figure~\ref{fig:cases}(a), Perplexity V1.2 selects a workflow that approves a document before editing it. Publication requires approval of the final revision, so those later edits invalidate the certificate. Jev and Microsoft Decision-1 select a valid ordering. In Figure~\ref{fig:cases}(b), Hazel's report is false under the unique consistent interpretation, yet all three models rank it above at least one true report. The two cases expose errors in checking an option against history or other options, even when the relevant conditions are explicit.

\finding{Models struggle to translate a simple record-level defect into the correct score distribution.} 
In Figure~\ref{fig:cases}(c), eight of nine records satisfy every rubric criterion, while one record lacks a review stamp. A three-record sample receives the lower score only when it contains that record, which happens with probability one-third. However, Perplexity V1.2 and Jev assign more probability to the lower score, while Microsoft Decision-1 predicts the correct most likely score but still assigns the probabilities incorrectly. This example shows that even a simple scoring rule can lead to errors in both the predicted score and its probability distribution.

%% file: tables/current-overall.tex
\begin{table*}[t]\centering
\fontsize{9}{10}\selectfont\setlength{\tabcolsep}{3.5pt}\renewcommand{\arraystretch}{1.0}
\resizebox{1.0\textwidth}{!}{\begin{tabular*}{\textwidth}{@{\extracolsep{\fill}}l | r | rr | rr@{}}\toprule
\textbf{Decision Model} & \shortstack{Single-select Acc.$\uparrow$} & \shortstack{Multi-select Acc.$\uparrow$} & SB-F$\uparrow$ & \shortstack{Soft Brier$\downarrow$} & \shortstack{Score nMAE$\downarrow$} \\ \midrule
\multicolumn{6}{c}{\textit{Heuristic reference}} \\ \midrule
Random (uniform) & 25.35 & 3.28 & 0.0000 & 0.2051 & 0.1624 \\
\midrule
\multicolumn{6}{c}{\textit{Closed-source decision models}} \\ \midrule
Jev 1.14.0 & 67.17 & 22.12 & 0.1187 & \mmdmbestcell{0.1101} & 0.0926 \\
Microsoft Decision-1 & 61.36 & 21.96 & 0.1186 & 0.2795 & 0.1836 \\
Liquid-d1-Ultra & 73.32 & 24.33 & 0.1624 & 0.1321 & \mmdmsecondcell{0.0883} \\
Liquid-d1 & 62.33 & 21.96 & 0.1228 & 0.2827 & 0.1401 \\
GPT-6 Luna Decisions & 56.57 & 12.01 & 0.0886 & 0.3333 & 0.1588 \\
\midrule
\multicolumn{6}{c}{\textit{Open-source decision models}} \\ \midrule
WeMM-JEV 27B & \mmdmbestcell{76.00} & \mmdmbestcell{36.81} & 0.1946 & 0.2327 & 0.2203 \\
KnowLine 31B Gen1 & \mmdmsecondcell{75.07} & 32.39 & 0.1777 & \mmdmsecondcell{0.1233} & \mmdmbestcell{0.0624} \\
Perplexity Decider-V1.2 & 74.42 & \mmdmsecondcell{33.81} & 0.1907 & 0.1694 & 0.1641 \\
Autotrust JEV 27B-VL & 67.07 & 20.22 & 0.1268 & 0.3405 & 0.2128 \\
Decisio-Gemma-4-31B & 64.64 & 15.32 & 0.1026 & 0.1653 & 0.1034 \\
Clef & 64.18 & 16.43 & 0.0987 & 0.1422 & 0.1092 \\
Perplexity Decider-V1 & 62.38 & 21.48 & 0.1291 & 0.1555 & 0.0999 \\
Perplexity Decider-Flash-V1.2 & 62.06 & 16.27 & 0.0965 & 0.2836 & 0.1674 \\
H2O-Lightning 4B & 58.29 & 20.54 & 0.1031 & 0.3482 & 0.2684 \\
Decisio-Gemma-4-12B & 57.71 & 18.17 & 0.0962 & 0.1627 & 0.1045 \\
Quyet-1.0-Large & 57.44 & 16.59 & 0.0890 & 0.1804 & 0.1079 \\
OpenJev 4B & 57.20 & 17.38 & 0.1085 & 0.3687 & 0.3106 \\
Clef-Omni & 56.91 & 14.85 & 0.0827 & 0.2761 & 0.1704 \\
Mapika Decider 4B & 56.49 & 14.06 & 0.0835 & 0.2131 & 0.1918 \\
Clef-Flash & 55.74 & 14.85 & 0.0843 & 0.3553 & 0.2586 \\
Decisio-Qwen3.6-35B-A3B & 53.33 & 16.11 & 0.1196 & 0.2554 & 0.1397 \\
Intern-Decision 4B & 49.03 & 11.06 & 0.0896 & 0.1564 & 0.1449 \\
JevK5 4B & 48.69 & 10.43 & 0.0800 & 0.2291 & 0.1770 \\
Strands-Decider 2B & 44.65 & 8.21 & 0.0580 & 0.2052 & 0.1689 \\
SemIf 4B & 42.39 & 6.32 & 0.0750 & 0.3402 & 0.1243 \\
Intern-Decision 2B & 38.76 & 6.00 & 0.0585 & 0.2011 & 0.1519 \\
Intern-Decision 0.8B & 32.17 & 3.95 & 0.0399 & 0.2051 & 0.1531 \\
NanoJev 0.6B & 26.68 & 7.11 & 0.0390 & 0.3061 & 0.2732 \\
\bottomrule\end{tabular*}}
\caption{Overall results. Accuracies are percentages. SB-F uses model-specific valid native multi-select outputs. Sample counts and fixed-cohort SB-F rankings appear in Table~\ref{tab:sbf}. \colorbox{mmdmBestRose}{\textbf{Best}} and \colorbox{mmdmSecondBlue}{\underline{second-best}} accuracy and error values are marked across all models.}\label{tab:overall}
\end{table*}

%% file: tables/current-category-all.tex
\begin{table*}[t]\centering
\setlength{\tabcolsep}{2.2pt}\renewcommand{\arraystretch}{1.16}
\resizebox{1.0\textwidth}{!}{\begin{tabular}{l | r | rrrrrrrrrrrrrrrr}\toprule
 & Overall & \multicolumn{16}{c}{Hard Single-select Accuracy by task category} \\
\cmidrule(lr){2-2}\cmidrule(lr){3-18}
\textbf{Decision Model} & \textbf{ALL} & \textbf{WKN} & \textbf{RDC} & \textbf{DED} & \textbf{OPL} & \textbf{CPL} & \textbf{QTR} & \textbf{WFD} & \textbf{AST} & \textbf{PEX} & \textbf{EVD} & \textbf{CDB} & \textbf{IRT} & \textbf{TSS} & \textbf{APF} & \textbf{SCR} & \textbf{EMR} \\ \midrule
\multicolumn{18}{c}{\textit{Heuristic reference}} \\ \midrule
Random (uniform) & 25.3 & 30.7 & 30.1 & 33.5 & 25.7 & 4.1 & 30.7 & 4.4 & 10.4 & 35.6 & 46.3 & 31.9 & 28.5 & 12.8 & 37.7 & 29.5 & 9.2 \\
\midrule
\multicolumn{18}{c}{\textit{Closed-source decision models}} \\ \midrule
Jev 1.14.0 & 67.2 & 63.4 & 69.6 & 81.2 & 38.2 & 42.7 & 75.8 & 52.2 & 51.0 & 71.5 & 84.2 & 76.4 & 78.5 & \mmdmbestcell{94.2} & 65.7 & \mmdmbestcell{85.7} & 34.3 \\
Microsoft Decision-1 & 61.4 & 62.6 & 63.5 & 76.8 & 38.2 & 13.5 & 65.0 & 43.5 & 57.8 & 59.0 & 76.2 & 76.4 & 60.4 & 79.5 & 66.1 & 71.4 & 47.8 \\
Liquid-d1-Ultra & 73.3 & 75.1 & \mmdmsecondcell{75.2} & 82.3 & 38.8 & 47.4 & \mmdmsecondcell{80.5} & \mmdmsecondcell{60.2} & 74.2 & 76.4 & 82.6 & \mmdmbestcell{88.2} & 75.2 & 89.6 & \mmdmsecondcell{72.5} & 77.1 & 39.9 \\
Liquid-d1 & 62.3 & 66.7 & 56.3 & 76.4 & 32.2 & 30.4 & 67.9 & 28.6 & 41.7 & 72.8 & 84.2 & 74.9 & 79.2 & 86.0 & 66.1 & 74.3 & 49.4 \\
GPT-6 Luna Decisions & 56.6 & 53.5 & 66.4 & 60.5 & 27.6 & 19.3 & 65.3 & 16.8 & 36.6 & 61.3 & 78.5 & 69.0 & \mmdmsecondcell{81.6} & 74.5 & 64.4 & 65.7 & 32.0 \\
\midrule
\multicolumn{18}{c}{\textit{Open-source decision models}} \\ \midrule
WeMM-JEV 27B & \mmdmbestcell{76.0} & \mmdmbestcell{81.0} & \mmdmbestcell{75.9} & \mmdmbestcell{86.3} & \mmdmbestcell{57.9} & 47.4 & \mmdmbestcell{82.7} & 49.1 & \mmdmbestcell{83.0} & 67.2 & 83.4 & 82.8 & 80.4 & \mmdmsecondcell{93.2} & 72.1 & 77.1 & \mmdmsecondcell{55.1} \\
KnowLine 31B Gen1 & \mmdmsecondcell{75.1} & 71.4 & 72.1 & \mmdmsecondcell{86.0} & \mmdmsecondcell{54.6} & \mmdmbestcell{48.5} & 78.0 & \mmdmbestcell{65.8} & \mmdmsecondcell{81.1} & \mmdmsecondcell{77.0} & 83.8 & \mmdmsecondcell{86.7} & 77.6 & 89.9 & \mmdmbestcell{74.2} & \mmdmbestcell{85.7} & 41.6 \\
Perplexity Decider-V1.2 & 74.4 & \mmdmsecondcell{78.8} & 73.2 & 85.2 & 53.9 & \mmdmsecondcell{48.0} & 79.8 & 47.2 & 79.4 & 70.8 & 83.4 & 80.8 & \mmdmsecondcell{81.6} & 91.4 & 69.5 & 74.3 & 50.6 \\
Autotrust JEV 27B-VL & 67.1 & 60.1 & 71.8 & 72.3 & 37.5 & 34.5 & 74.7 & 33.5 & 62.2 & 69.8 & 84.5 & 81.8 & 77.6 & 92.4 & 69.5 & 77.1 & 35.4 \\
Decisio-Gemma-4-31B & 64.6 & 60.8 & 69.1 & 65.3 & 28.9 & 35.7 & 69.7 & 14.9 & 64.5 & 71.8 & 78.5 & 77.8 & \mmdmbestcell{81.9} & 85.6 & 69.5 & \mmdmsecondcell{82.9} & 31.5 \\
Clef & 64.2 & 72.2 & 71.2 & 66.8 & 36.8 & 26.9 & 70.8 & 24.2 & 58.9 & 70.2 & 78.9 & 70.4 & 76.1 & 72.3 & 65.2 & 80.0 & 52.8 \\
Perplexity Decider-V1 & 62.4 & 61.5 & 63.1 & 68.3 & 30.9 & 28.1 & 65.7 & 29.8 & 62.1 & 59.7 & 81.9 & 78.8 & 77.9 & 77.0 & 64.8 & 77.1 & 37.1 \\
Perplexity Decider-Flash-V1.2 & 62.1 & 65.2 & 60.1 & 74.5 & 30.9 & 17.0 & 69.7 & 17.4 & 57.0 & 72.8 & 79.2 & 67.0 & 76.1 & 82.4 & 61.4 & 68.6 & 48.9 \\
H2O-Lightning 4B & 58.3 & 57.5 & 57.4 & 59.8 & 37.5 & 21.1 & 62.5 & 24.8 & 54.2 & 64.9 & 80.4 & 71.4 & 67.7 & 75.2 & 65.7 & 60.0 & 36.0 \\
Decisio-Gemma-4-12B & 57.7 & 61.2 & 59.5 & 56.8 & 27.6 & 19.3 & 70.0 & 24.2 & 42.6 & 64.9 & 78.1 & 72.9 & 77.3 & 79.5 & 63.1 & 68.6 & 28.7 \\
Quyet-1.0-Large & 57.4 & 71.1 & 64.9 & 68.3 & 28.9 & 28.7 & 73.6 & 3.7 & 28.2 & \mmdmbestcell{77.4} & \mmdmbestcell{89.1} & 82.8 & 71.3 & 51.1 & 67.8 & 74.3 & 21.9 \\
OpenJev 4B & 57.2 & 59.0 & 42.6 & 73.8 & 48.0 & 25.7 & 70.4 & 22.4 & 49.7 & 68.5 & 79.2 & 73.9 & 57.1 & 70.5 & 63.9 & 71.4 & 33.7 \\
Clef-Omni & 56.9 & 74.4 & 61.9 & 62.4 & 34.2 & 25.1 & 62.1 & 31.7 & 34.8 & 70.8 & 75.5 & 64.0 & 58.6 & 71.2 & 59.7 & 65.7 & 50.0 \\
Mapika Decider 4B & 56.5 & 60.1 & 54.3 & 69.4 & 30.9 & 19.3 & 61.0 & 14.3 & 40.4 & 59.0 & \mmdmsecondcell{84.9} & 74.4 & 64.4 & 70.5 & 58.8 & 74.3 & \mmdmbestcell{64.0} \\
Clef-Flash & 55.7 & 73.3 & 58.3 & 60.1 & 36.2 & 19.9 & 65.3 & 28.6 & 34.6 & 62.0 & 76.2 & 63.1 & 64.4 & 66.2 & 60.1 & 62.9 & 51.1 \\
Decisio-Qwen3.6-35B-A3B & 53.3 & 64.5 & 63.1 & 62.4 & 28.9 & 18.7 & 65.3 & 19.9 & 29.7 & 60.7 & 81.5 & 69.5 & 47.4 & 73.0 & 61.4 & 68.6 & 28.7 \\
Intern-Decision 4B & 49.0 & 46.5 & 48.6 & 60.9 & 25.0 & 14.6 & 65.3 & 12.4 & 31.6 & 41.3 & 75.5 & 66.5 & 62.5 & 64.0 & 63.9 & 68.6 & 31.5 \\
JevK5 4B & 48.7 & 57.9 & 52.3 & 70.1 & 30.3 & 19.9 & 61.7 & 6.8 & 26.9 & 50.5 & 79.6 & 72.4 & 43.5 & 61.9 & 58.4 & 68.6 & 15.7 \\
Strands-Decider 2B & 44.6 & 44.0 & 45.5 & 56.1 & 30.9 & 18.7 & 49.5 & 9.3 & 17.2 & 59.7 & 71.3 & 64.0 & 51.1 & 53.6 & 57.9 & 65.7 & 34.8 \\
SemIf 4B & 42.4 & 41.8 & 52.0 & 52.0 & 23.0 & 8.2 & 53.4 & 6.2 & 21.1 & 46.2 & 70.6 & 62.1 & 43.5 & 57.6 & 54.1 & 65.7 & 16.9 \\
Intern-Decision 2B & 38.8 & 37.7 & 41.0 & 50.2 & 27.6 & 10.5 & 55.6 & 4.3 & 15.3 & 41.3 & 61.9 & 58.6 & 36.9 & 52.2 & 48.9 & 71.4 & 30.9 \\
Intern-Decision 0.8B & 32.2 & 37.0 & 36.3 & 35.1 & 21.7 & 8.8 & 48.4 & 4.3 & 11.8 & 37.0 & 55.8 & 38.4 & 37.5 & 34.9 & 37.3 & 71.4 & 23.6 \\
NanoJev 0.6B & 26.7 & 30.0 & 30.2 & 40.2 & 19.7 & 2.3 & 24.2 & 6.2 & 11.0 & 39.7 & 47.9 & 24.6 & 39.9 & 10.1 & 45.5 & 40.0 & 13.5 \\
\bottomrule\end{tabular}}
\caption{Hard Single-select Accuracy (\%, $\uparrow$). ALL aggregates the 4,112 single-select questions. Colors mark the best and second-best model values as in Table~\ref{tab:overall}. Task codes are WKN (World knowledge); RDC (Reading comprehension); DED (Deductive reasoning); OPL (Option logic); CPL (Constraint planning); QTR (Quantitative and temporal reasoning); WFD (Workflow decision); AST (Agent state tracking); PEX (Policies and exceptions); EVD (Evidence validation); CDB (Code and databases); IRT (Intent routing); TSS (Tool and skill selection); APF (Answer preference); SCR (Score and rubrics); EMR (Emotion recognition).}\label{tab:categoryall}
\end{table*}

%% file: tables/current-case-studies.tex
\begin{figure*}[t]
\centering
\begin{minipage}[t]{.318\textwidth}
\small\raggedright
\colorbox{exampleSage}{\parbox{\dimexpr\linewidth-2\fboxsep}{\textbf{(a) Approval and later edits}}}\par\smallskip
Finch certifies the current content. Lumen and Fern modify it. Publication requires a certificate matching the final revision.\par\smallskip
\textbf{Valid.} edit, convert, update envelope, then approve and publish.\par
\textbf{Invalid.} approve first, then edit and convert.\par\smallskip
{\fontsize{8}{9.3}\selectfont\setlength{\tabcolsep}{2.5pt}\begin{tabular}{@{}lr@{}}\toprule
Model & Selected option \\\midrule
Perplexity V1.2 & \textcolor{exampleWrong}{Invalid (0.353)}\\
Jev & \textcolor{exampleCorrect}{Valid (0.520)}\\
Microsoft Decision-1 & \textcolor{exampleCorrect}{Valid (0.894)}\\\bottomrule
\end{tabular}}\par\smallskip
\textit{Later revisions invalidate an earlier approval.}
\end{minipage}\hfill
\begin{minipage}[t]{.318\textwidth}
\small\raggedright
\colorbox{exampleBlue}{\parbox{\dimexpr\linewidth-2\fboxsep}{\textbf{(b) Complete set identification}}}\par\smallskip
Eight reports refer to one another. For example, Hazel says Maple's report is false; Maple says Pine's or Willow's report is true.\par\smallskip
\textbf{Reference.} all seven reports except Hazel are true under the unique consistent assignment.\par\smallskip
{\fontsize{8}{9.3}\selectfont\setlength{\tabcolsep}{2.5pt}\begin{tabular}{@{}lrr@{}}\toprule
Model & $p_{\rm Hazel}$ & $\min p_{\rm true}$\\\midrule
Perplexity V1.2 & 0.083 & 0.075\\
Jev & 0.120 & 0.110\\
Microsoft Decision-1 & 0.075 & 0.067\\\bottomrule
\end{tabular}}\par\smallskip
\textit{All three rank a false report above a true one.}
\end{minipage}\hfill
\begin{minipage}[t]{.318\textwidth}
\small\raggedright
\colorbox{exampleClay}{\parbox{\dimexpr\linewidth-2\fboxsep}{\textbf{(c) A distribution over scores}}}\par\smallskip
Draw three of nine records. Eight pass all four checks; one lacks a review stamp. Each check earns one point only if every drawn record passes it.\par\smallskip
\textbf{Reference.} the incomplete record is drawn with probability $3/9$.\par\smallskip
{\fontsize{8}{9.3}\selectfont\setlength{\tabcolsep}{2.5pt}\begin{tabular}{@{}lrr@{}}\toprule
Distribution & Score 3 & Score 4\\\midrule
Reference & 0.333 & 0.667\\
Perplexity V1.2 & 0.649 & 0.331\\
Jev & 0.590 & 0.310\\
Microsoft Decision-1 & 0.191 & 0.754\\\bottomrule
\end{tabular}}\par\smallskip
\textit{Two models reverse the dominant score probabilities.}
\end{minipage}
\caption{Saved model responses to three decisions involving workflow prerequisites, interdependent options, and uncertain rubric outcomes. Inputs are condensed for display; complete examples are Examples~\ref{ex:workflow_decision}, \ref{ex:option_logic}, and~\ref{ex:score_rubric} in Appendix~\ref{app:examples}. Probabilities in (c) omit levels 0--2.}
\label{fig:cases}
\end{figure*}

%% file: sections/conclusion.tex
\section{Conclusion}
We introduced MMDM, a benchmark of 5,448 questions across 16 task categories for evaluating text-based System-1 Decision Models. With option sets ranging from 2 to 1,024 and inputs reaching approximately 64K tokens, MMDM tests decisions across both familiar requests and complex contexts. Its single-select, multi-select, and soft-label tasks assess whether models choose correctly, identify all valid options, and assign useful probabilities and scores.
Our evaluation of 28 decision models reveals clear gaps between these capabilities and exposes dimensions of decision quality that are not captured by previous benchmarks.
The strongest models vary across answer types and task categories. High single-select accuracy can accompany inaccurate probabilities, while deduction and routing strengths coexist with weaknesses in joint constraints and workflow decisions. Reliability curves and scaling profiles further distinguish confidence quality from accuracy and long-input performance from large-option performance. These findings direct progress toward checking interacting conditions, handling larger decision spaces, and producing probabilities that support reliable choices and scores. We release MMDM and its evaluation toolkit to make progress on these capabilities easy to measure and compare.

%% file: sections/limitations.tex
\section*{Limitations}
MMDM has three main limitations. First, it focuses on text-based decisions defined by a fixed state, query, and option set. Extending MMDM to multimodal inputs and interactive trajectories would broaden its coverage of real-world decision settings.
Second, multi-select questions are concentrated in option logic, constraint planning, workflows, and tool or skill selection, while most soft targets come from rubric scoring. These settings provide clear labels and verifiable target distributions, but broader evaluation of subjective judgments will require references that better capture variation across human assessments.
Third, MMDM combines newly constructed tasks with source-preserving adaptations, recompositions, and rewrites. Although recompositions and rewrites introduce new decision requirements and contexts, prior exposure to the underlying public source material may still occur.

%% file: sections/ethics.tex
\section*{Ethics Statement}
MMDM combines publicly available research data with newly constructed decision tasks. We cite the source datasets and document their contributions in Appendix~\ref{app:selection}. Source materials remain subject to their original terms of use. The same appendix describes model-assisted construction, label verification, and targeted human review, including independent assessment and adjudication of disagreements. Our experiments evaluate model predictions from recorded or simulated inputs; they do not execute the tools or workflows described in the questions. We release the benchmark and evaluation toolkit, with source attribution, to support transparent and reproducible research.

%% file: sections/appendix.tex
\begin{center}
{\Large\textbf{Appendices}}
\end{center}
\noindent The appendices are organized as follows.\\
$\bullet$ Appendix~\ref{app:sources} reports input coverage and defines the tasks.\\
$\bullet$ Appendix~\ref{app:selection} describes data sources, construction, label derivation, and quality control.\\
$\bullet$ Appendix~\ref{app:models} details decision models, the Random reference, capacity handling, and supplementary metrics.\\
$\bullet$ Appendix~\ref{app:matched} reports complete category results, matched cohorts, and reliability analyses.\\
$\bullet$ Appendix~\ref{app:repro} describes release artifacts and reproducibility.\\
$\bullet$ Appendix~\ref{app:examples} presents examples with saved model responses for all 16 categories.

\section{Input Coverage and Task Definitions}
\label{app:sources}
\textbf{Classification and Coverage.}
Table~\ref{tab:taxonomy} defines the 16 capabilities and links each one to its corresponding example in Appendix~\ref{app:examples}. A category identifies the capability a decision requires; the answer type specifies whether the target is one option, a complete valid set, or a distribution. 
We measure input length using the full text seen by the model, including the state, query, option names, and option criteria. Each example is capped at 65,535 tokens. Among the 5,448 examples, 3,155 are shorter than 1K tokens, while 356 exceed 8K tokens, including 107 longer than 32K. The number of options ranges from 2 to 1,024.
At inference time, the actual token usage can be slightly higher because model providers add their own request formatting and reserved tokens. 

\begin{figure*}[t]
\centering\includegraphics[width=\textwidth]{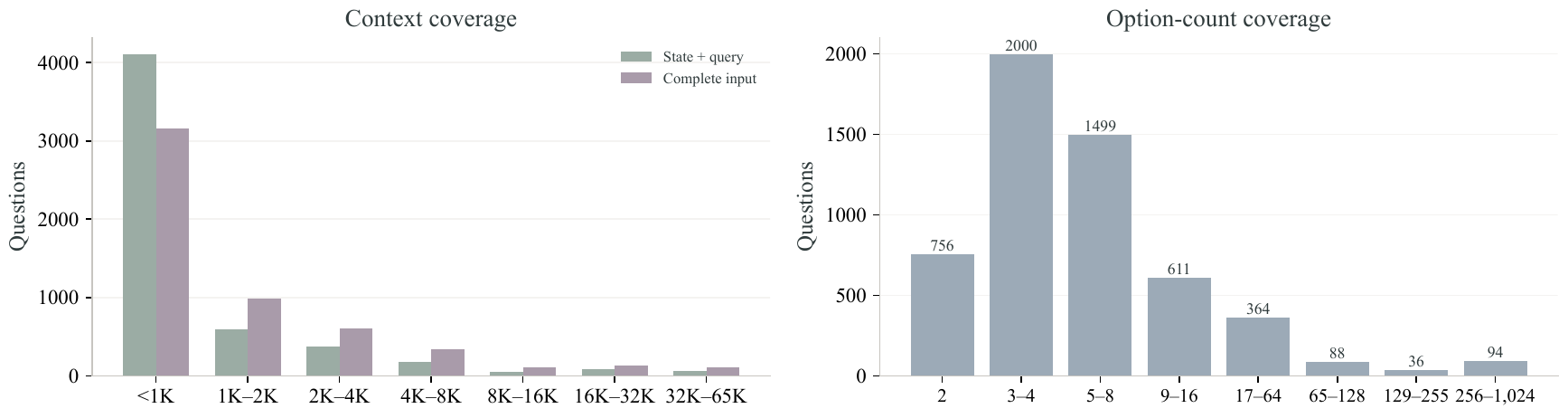}
\caption{Released input-length and option-count distributions. Complete input includes state, query, option names, and criteria under the common Qwen3.5-4B tokenizer. The state-plus-query view separates document content from catalog size.}
\label{fig:inputcoverage}
\end{figure*}
\input{tables/current-taxonomy}

\textbf{Decision Interfaces.}
Jev exposes three interfaces for different kinds of requests~\citep{typesafeai2026jev}. They specify what the model returns. The answer types in Section~\ref{sec:formulation} specify what makes a prediction correct and how it is evaluated.

\emph{Choice} compares named options. A support request might ask whether a ticket belongs to billing, shipping, or returns, with each option's criterion defining its responsibilities. The response contains the selected option and a probability for every option. In MMDM, this distribution supports evaluation against a single correct option, a set of equally valid options, or a soft reference, according to the task's answer type.

\emph{Noul} answers a yes/no query, such as whether a message requests a refund. It returns the probability of yes; values near one favor yes, values near zero favor no, and values near one-half express uncertainty between them. The remaining probability belongs to no, giving the two-option distribution used in evaluation. A definite reference answer is hard single-select, while a reference that quantifies uncertainty is soft-label.

\emph{Score} evaluates an input against ordered, descriptive rubric levels. For report completeness, these might range from missing required evidence to meeting every requirement. The response contains a probability for each level and their probability-weighted numeric score. MMDM evaluates both the level distribution and the expected score. Therefore, Scoring questions belong to the hard single-select or soft-label track, depending on whether the reference identifies one level or a distribution.

\section{Data Sources and Construction}
\label{app:selection}
\subsection{Source Collection and Adaptation}
Table~\ref{tab:publicsources} lists every public source represented in the release, its contribution, and the decision task obtained from it. We collect labeled questions, supporting documents, response comparisons, code and database problems, agent trajectories, and tool or skill descriptions. 
Table~\ref{tab:originalsources} describes the question families that we construct from explicit rules, state transitions, and outcome mechanisms. For source-grounded rewrites, the original answer and its supporting context serve as semantic anchors. We rewrite the state and query and rephrase or replace distractors, preserving the facts and relations needed to justify the answer. The semantic and option-quality checks in Appendix~\ref{app:qualityreview} verify that the revised input supports the reference answer and that incorrect options remain plausible but invalid.

\input{tables/current-public-sources}

\textbf{Knowledge, Language, and Document.}
Adaptations preserve task-relevant evidence and label semantics within the state--query--options interface. Knowledge and reading tasks retain the premises needed to distinguish the answer; multi-hop questions preserve the documents and links joining the required evidence. Entailment and moderation tasks express the source labels as named options with explicit meanings. Emotion tasks preserve the source text and the relevant emotion inventory. Preference tasks retain the compared responses and the criterion governing the comparison. A preference between responses remains a selection label; numerical annotations are used according to their original meaning.

\textbf{Code and Database.}
Program tasks preserve executable code and its input or specification. Database tasks provide the relevant schema and database conditions. In the BIRD adaptation~\citep{li2023bird}, we derive a decision query from the target SQL semantics and execute every SQL option against the source database. Validation checks the required operation, the resulting records, and whether distinct options produce distinguishable outputs. An option-only similarity probe checks whether a response can be selected by superficial resemblance among SQL statements without using the query, which rules out some naive samples.

\textbf{Retrieval and Freshness.}
Adaptive-RAG~\citep{jeong2024adaptive} supplies retrieval-strategy decisions, while RetrievalQA~\citep{zhang2024retrievalqa} supplies cases requiring a judgment about external retrieval. FreshQA~\citep{vu2024freshllms} supplies stable, time-sensitive, and false-premise questions. Its adaptation provides an explicit evidence date and decision date, and asks whether the available evidence suffices, should be refreshed, or contains a premise that should be rejected. The target is tied to the dated evidence presented in the state.

\textbf{Agent Histories and Skill Catalogs.}
At a decision point $t$, the state is the observed interaction prefix $h_{\leq t}$, and the query asks which tool or skill is appropriate next. The prefix ends before the target action. Options contain tool names and contracts or skill names and descriptions. The release includes 239 such decisions from ComplexFuncBench~\citep{zhong2025complexfuncbench} and ToolTalk~\citep{farn2023tooltalk}, comprising 163 tool selections and 76 skill selections. Two separate DeepSeek-V4-Pro~\citep{deepseekai2026deepseekv4} calls review the history and every option with labels hidden and option order changed. Acceptance requires agreement on present applicability and the complete valid set.

Catalog-based questions use task and environment descriptions from SkillsBench~\citep{li2026skillsbench}, retrieval annotations from ToolRet~\citep{shi2025toolret}, and skill descriptions from Graph of Skills~\citep{liu2026graphskills}. Generated requests introduce concrete goals and prerequisites grounded in these descriptions. An additional 31 questions are built from a collected directory of 10,798 skills with provenance preserved. For each question, we review every option for functional overlap and applicability to the current request; all valid options are included in the multi-select target. Original retrieval positives and packaged skills serve as reference candidates, while the expanded option set is independently reviewed to determine the labels.

\subsection{Original Task Construction}
\label{app:constructiondetails}
We construct new decision instances by specifying a mechanism, computing its reference answer, and rendering its states and conditions as natural language. The mechanism determines which details affect the decision. Distractors implement plausible mistakes, including an omitted prerequisite, an expired authorization, a stale artifact, a confused entity, or a missed exception. The final families and their validation procedures appear in Table~\ref{tab:originalsources}.

\textbf{Interdependent Options and Constrained Plans.}
Logic tasks define propositions whose truth values depend on other propositions. We enumerate satisfying assignments and check them with an independent truth-table implementation. Single-select questions ask for the unique complete interpretation; multi-select questions ask for all true propositions under that interpretation. Global interpretation rules are stated in the state, and interdependent statements appear in \texttt{option.criteria}. We check entity references, quantifier scope, and natural-language statements against the formal propositions.
Planning tasks evaluate complete proposed assignments against eligibility, scheduling, resource, and precedence constraints; all feasible plans remain valid options.

\textbf{State Transitions and Workflows.}
State tasks reconstruct the current situation from events and tool receipts. Their mechanisms track goal scope, artifact versions, committed or rolled-back updates, evidence readiness, and execution failures. Transactional histories are checked both by folding the event sequence and by independent SQLite replay. Workflow tasks compare complete action sequences under prerequisites and revision-sensitive certificates. A simulator tracks state and content versions, while a separate checker verifies action positions and required orderings. Workflow option sets contain at most 128 sequences. 

\textbf{Scenario and Rule Composition.}
Additional original tasks combine policy conditions, evidence, tools, practical planning, and scoring in concrete scenarios. Compositional rule programs introduce thresholds, exceptions, authority levels, document versions, and joins between pieces of evidence. Natural-language renderings preserve the facts and rules that determine the computed answer. For tasks with finite uncertain outcomes, the state supplies the underlying weights, transition probabilities, or sampling process used to derive the reference distribution.

\subsection{Answer Encoding and Label Derivation}
\label{app:softtargets}
The released input follows $x=(s,q,\mathcal{O})$ in Equation~\eqref{eq:decision}. Each option has a stable identifier $\iota_j$ that links its name and criterion to the reference label. The record also specifies its Choice, Noul, or Score primitive and, for Score, the numeric levels $v_1,\ldots,v_K$. The released field \texttt{question} stores the query, and \texttt{option.criteria} stores each option's description. Labels, categories, provenance, and verification traces are separate from the model-visible input.

\textbf{Hard Single-select and Hard Multi-select.}
A unique correct option $y$ is encoded as $\pi_j=\ind[j=y]$. A complete set of equally valid options $G$ is encoded as $\pi_j=\ind[j\in G]/|G|$. The latter target records known set membership, and evaluation tests whether every member ranks above every nonmember. Source labels, executable constraints, or agreement-based review determine these references according to the construction route. A deterministic rubric score uses a hard label at the corresponding numeric level.

\textbf{Soft-label and Score.}
For finite outcomes $\omega\in\Omega$ and an answer operator $r(\omega)$ mapping each outcome to an option index, the soft reference is
\begin{equation}
 \pi_j=\sum_{\omega\in\Omega}P(\omega\mid x)\ind[r(\omega)=j].
 \label{eq:soft}
\end{equation}
For Score, option $j$ carries the numeric rubric level $v_j$. The reference follows the uncertainty explicitly specified in the task. Weighted finite worlds, state transitions, and sampling without replacement provide computable outcome probabilities.

The main Score family evaluates a common rubric over a subset of records. Records carry attributes such as signed handoffs, attachments, current templates, and review markers. For a uniform sample $S$ of size $m$ from a population of $L$ records,
\begin{equation}
 \pi_j=\frac{\left|\{S:|S|=m,\ r(S)=j\}\right|}{\binom{L}{m}}.
\end{equation}
Every criterion is evaluated on the same sample, preserving dependencies among the checks. Exact enumeration is cross-checked by an independent combinatorial calculation. This family contributes 598 questions. Across all families, the 703 soft questions comprise 628 Score and 75 Choice or Noul questions. Every distribution records all levels, including zero-probability outcomes; Score also records the numeric scale. Example~\ref{ex:score_rubric} illustrates this construction.

\input{tables/current-original-sources}

\subsection{Cleaning, Correction, and Verification}
\label{app:qualityreview}
We initially collected 683,237 instances from public sources, retaining 669,135 after normalization and deduplication. We subsequently constructed 75,588 additional instances, producing a pool of 744,723 for selection. 

\textbf{Representation and Semantic Checks.}
We check unique item and option identities, complete option references, valid score scales, normalized nonnegative labels, and agreement between answer types and references. Source-parent and request-overlap checks compare related instances with registered training, replay, and evaluation records. 
Answers are checked against source evidence, execution, independent formal computation, or blind review. Repairs undergo the corresponding check again; unresolved failures are excluded.

\textbf{Option Quality.}
We further check whether the options can be distinguished from the information available in the state and query, rather than from superficial cues. Distractors should reflect realistic failure modes, such as overlooking a prerequisite or choosing a tool with similar functionality but incompatible requirements. We also mask the state and query and ask whether the answer can still be guessed from option names or descriptions alone. Options with obvious shortcuts, copied answers, or weak distinctions are removed. For SQL tasks, we additionally check whether different queries are semantically equivalent and whether surface similarity reveals the answer.

\textbf{Multi-model Screening.}
Candidate instances are screened by DeepSeek~\citep{deepseekai2026deepseekv4}, GPT-6-Astra~\citep{openai2026gpt6astra} with high reasoning effort, Gemini-3.8-Flash~\citep{googledeepmind2026gemini38flash}, Qwen3.8-Max~\citep{qwenteam2026qwen38max}, and GLM-5.2~\citep{zai2026glm52}. Each model independently judges whether an instance meets the representation, semantic, and option-quality standards above. Four or five accept votes support automatic inclusion; a 3--2 vote sends the instance to human review, and fewer than three accept votes exclude it. This stage accepted 4,715 instances automatically and referred 928 borderline cases for human review.

\textbf{Independent Human Review.}
Two reviewers independently assessed all 928 borderline cases under a common standard operating procedure (SOP). The SOP applies the representation, semantic, and option-quality standards above, using source evidence, verification traces, and option-only probe results from the earlier checks. Reviewers verify that the state and query support the reference answer, that all valid options are represented, and that distractors reflect plausible errors rather than superficial cues. Their initial decisions agreed on 97.3\% of cases; a third reviewer adjudicated the remaining disagreements. Corrections underwent the corresponding semantic or executable checks before inclusion. Combining automatically accepted and human-reviewed instances produced the final 5,448 questions.

\input{tables/appendix-current-models}

\section{Models and Evaluation Details}
\label{app:models}
\subsection{Decision Models and Random Reference}
\label{app:modelsettings}
Table~\ref{tab:models} lists the selected decision models, release references, and deployment settings. Selection requires direct option-probability output through a decision-scoring head or a single decoding step, without generated reasoning or adaptive thinking.
The run settings record model versions, tokenizers, context budgets, service and output reserves, option caps, code versions, and grouping seeds. Decisio entries identify both the inference implementation and base checkpoint. Limits reflect the model and serving interface. 
\input{tables/current-category-distributions}
\input{tables/appendix-current-matched}
\input{tables/appendix-current-set}
\input{tables/appendix-current-set-modes}

Random supplies a heuristic point of reference by assigning equal probability to all $K$ options. We report analytical expected accuracies of $1/K$ for single-select and $1/\binom{K}{|G|}$ for multi-select under uniform random tie-breaking. Its uniform distribution is used to compute SB-F, Soft Brier, and Score nMAE. Random has SB-F zero on every released multi-select question, since each contains at least one incorrect option.

\subsection{Protocol and Metrics}
\label{app:protocol}
\label{app:metrics}
\textbf{Context and Option Handling.}
For context budget $B$, service allowance $H$, and output reserve $R$, the adapter preserves the query, all options, and their criteria. It keeps approximately equal token spans from the start and end of the state and recounts the serialized request against $B-H-R$. Truncation preserves Unicode boundaries and inserts an omission marker. Runs record the provider or proxy tokenizer. Fixed-field overflow and provider rejection count as failures; grouped calls share the prepared state.
All requests follow the protocol in Section~\ref{sec:setup}. A supported catalog is evaluated in one call. An independent catalog above the option cap is reduced recursively, with each call presenting all options in its group alongside the same prepared state. The highest-probability option in each group advances for hard single-select. For hard multi-select, the evaluator uses $N=|G|$ both as the evaluation cutoff and as the maximum number retained from each group; neither $N$ nor the reference option identities enter model requests. The metric evaluates complete-set ordering at this reference cardinality. Option-dependent and distributional tasks retain the complete option set without splitting. If that set exceeds a model's option cap, the request is recorded as a capacity failure. 

\textbf{Implementation and Averaging.}
Hard accuracies include all questions and count boundary ties, failed requests, and missing outputs as incorrect. The implementation treats probability differences of at most $10^{-12}$ as ties. SB-F, Soft Brier, and Score nMAE use valid native outputs over the complete option set. Each Score error is normalized by that item's rubric range.

\textbf{Confidence Calibration.}
We also measure how well single-select confidence predicts correctness using expected calibration error (ECE)~\citep{guo2017calibration}. For $L$ predictions partitioned into ten equal-width confidence bins $\mathcal{B}_b$,
\begin{equation}
 \operatorname{ECE}=\sum_{b=1}^{10}\frac{|\mathcal{B}_b|}{L}\left|\bar a_b-\bar u_b\right|,
 \label{eq:ece}
\end{equation}
where $\bar a_b$ is mean correctness and $\bar u_b$ is mean maximum option probability in bin $b$; empty bins contribute zero. Lower ECE indicates closer agreement between confidence and accuracy. Matched calibration analyses use the same single-select questions with valid, native, full-state predictions and unique maxima across the compared models.

\textbf{Unified Model Output.}
Every decision model returns a probability distribution over the options supplied in its request. The answer-type annotation and reference number of correct options are not included in the model input. Thus, single-select, multi-select, and soft-label questions share the same prediction interface, without an additional prompt specifying the target answer type or set size. Reference labels are only used by the evaluator

\input{tables/current-sbf}

\section{Additional Experimental Results}
\label{app:matched}
\subsection{Results by Task Category}
Table~\ref{tab:categorydistributions} extends Table~\ref{tab:categoryall} to every category with multi-select or soft-label questions. The two blocks distinguish exact multi-select accuracy from distributional error. Jev achieves the lowest Soft Brier in three of the six categories, while the strongest multi-select results vary across models and tasks.

\subsection{Matched-Cohort Analysis}
Table~\ref{tab:matched} compares the five representative models on shared questions. For each metric, we retain questions for which all five models produced valid, native predictions using the complete input. Calibration also requires a unique highest-probability option. With the questions and input handling held fixed, Perplexity V1.2 leads both hard-answer metrics, while Jev has the lowest Soft Brier and Score nMAE. The contrast between selection accuracy and probability quality therefore remains on shared inputs.
\input{tables/current-bootstrap}

\subsection{Multi-select Results}
Table~\ref{tab:setcategories} focuses on the four substantial multi-select categories for the same five models. All five perform best on Tool and Skill Selection, where options describe independently applicable capabilities. Option Logic, Constraint Planning, and Workflow Decision instead require resolving dependencies or validating complete plans, and their accuracies remain much lower. This contrast helps explain why a model's overall multi-select score depends on the task composition.

Table~\ref{tab:setmodes} breaks down the same protocol by whether the complete option set fits the model's option cap. Accuracy remains low on catalogs requiring recursive reduction. These larger catalogs differ across models because their caps differ, so Table~\ref{tab:matched} complements this breakdown with 523 shared multi-select questions evaluated in one call with the complete state and option set. The strongest result on this common cohort is 36.90\%, showing that the multi-select difficulty persists without option reduction. SB-F complements exact accuracy by measuring the strength and balance of support for the correct set.

\input{tables/current-multiselect-diagnostics}

\input{tables/current-context-diagnostics}

\subsection{Multi-select Separation and Balance}
\label{app:sbfresults}
We further examine the probability distributions behind these multi-select results. Table~\ref{tab:sbf} reports SB-F, separation, and balance on 284 questions shared by all 28 decision models, with results also broken down by category.

\textbf{Weak Separation.} 
WeMM-JEV and Perplexity V1.2 have the highest SB-F (0.1905 and 0.1900, respectively), but excel on different tasks. WeMM-JEV is stronger on Option Logic and Constraint Planning, while Perplexity performs better on Workflow Decision and Tool and Skill Selection. We average each metric over questions for each model. The medians across models are 0.113 for separation and 0.862 for balance, while median correct-option coverage is only 37.8\%. Models therefore often distribute probability evenly among correct options without separating them from incorrect ones.
Jev illustrates this pattern, with average balance of 0.915 and separation of 0.137. OpenJev has stronger separation but lower balance. Jev's category results make the distinction clearer. Its SB-F falls from 0.465 on Tool and Skill Selection to 0.022 on Option Logic, where balance remains high (0.973) but separation is low (0.023). The main difficulty in these questions is distinguishing interdependent true statements from false ones.

\textbf{Coverage and Probability Margins.} 
Table~\ref{tab:multidiagnostics} examines these errors on the same 284 questions. A valid top choice satisfies $\max_{i\in G}p_i>b$, where $b$ is the largest incorrect-option probability. Coverage measures the fraction of correct options above $b$, averaged across questions. Probability differences of at most $10^{-12}$ count as ties. GPT-6 Luna Decisions and Jev rank a valid option first on 184 and 181 questions, respectively, but place the entire correct set above all distractors on only 36 and 57. Recognizing one valid option thus frequently leaves other valid options behind.

Even when the full set is correctly ordered, the probability margin can be narrow. We define this as $b\geq\tfrac12\min_{i\in G}p_i$, meaning that the strongest distractor receives at least half the probability of the weakest correct option. This occurs in 20 of Luna's 36 successes and 43 of WeMM-JEV's 110. Across all 28 models, the median proportion is 45.6\%. SB-F captures this weakness in probability separation even when exact accuracy counts the prediction as correct.

\subsection{Task Mechanisms and Input Scale}
\label{app:taskdiagnostics}
\textbf{State Reconstruction.} 
We compare three types of state-tracking question, each with 16 options, using native predictions from all five representative models on complete inputs. Jev and Perplexity answer all 47 artifact-version questions correctly, identifying outputs with the required source version, validation, and approval. Their accuracies fall to 8.3\% and 16.7\% on 24 transactional-state questions, which require applying an ordered sequence of transactions under revision, lock, and capacity constraints. On 100 goal-scope questions, which ask which instructions apply given their owner, workspace, and revision, Jev reaches 26.0\% and Perplexity 75.0\%. Median input lengths are 2,836, 4,252, and 1,800 tokens, respectively. 

\textbf{Length and Decision Requirements.} 
Table~\ref{tab:context-diagnostics} compares long-document comprehension and multi-hop reasoning within Reading Comprehension category, using the same input-length ranges. All questions have four options, and both Jev and Perplexity receive the full state. The two reading tasks have different accuracies within each length range, with each task represented by its own set of questions. 

The large-option comparisons cover Tool and Skill Selection and Constraint Planning, holding option count and inference mode fixed. For the 512-option skill-selection and plan-comparison questions, option names and criteria account for 99.4\% and 97.1\% of input tokens, respectively. Thus, long inputs can require extensive option checking as well as state comprehension. The comparisons separate finding an applicable skill from checking whether a complete plan satisfies its constraints.

\subsection{Task-Conditioned Probability Diagnostics}
\label{app:probabilitydiagnostics}
\textbf{Confidence Across Tasks.} We examine predictions with confidence at least 0.9 among the 3,537 shared questions in the reliability analysis. Jev is almost equally confident on retrieval strategy (mean 0.969 over 123 predictions) and contextual emotion (0.968 over 50), yet accuracy differs sharply, at 97.6\% and 50.0\%, respectively. OpenJev answers all 55 high-confidence SQL questions correctly, compared with 63.2\% of its 38 high-confidence emotion questions. Thus, the same confidence threshold corresponds to different error rates across tasks.
To capture this variation, we compute ECE separately within each task category using the same ten confidence bins, then weight the results by question count. This prevents overconfidence in one task from cancelling underconfidence in another. ECE rises from 0.054 to 0.106 for Jev, from 0.085 to 0.110 for Intern-Decision, and from 0.092 to 0.116 for OpenJev. Jev illustrates the opposing errors, with confidence exceeding accuracy by 38.6\% on Emotion Recognition but falling below it by 29.0\% on Workflow Decision.

\textbf{Expected Scores and Their Distributions.} We examine 596 sampled-rubric questions using native predictions from all five models on complete inputs. For predictions with nMAE at most 0.05, equivalent to an error of at most 0.2 points on the 0--4 scale, we measure distribution error using total variation, $\operatorname{TV}(p,\pi)=\frac12\sum_j|p_j-\pi_j|$. This is the probability mass that must be reassigned to match the reference. Jev meets this nMAE threshold on 190 questions and Intern-Decision on 111. Their mean TV errors remain 0.323 and 0.306, respectively. Thus, a close estimate of the expected score can still misrepresent the probabilities of individual score levels.
Each question scores a random subset of records against four criteria. The expected total score depends on the four individual pass probabilities, while the full score distribution also depends on which criteria pass together. Using the exact pass probabilities, we construct an approximation that treats the criteria as independent. It preserves the expected score but produces mean Brier and TV errors of 0.0127 and 0.0690 across the 596 questions. This illustrates why accurate expected scores and accurate score distributions require different information.

\section{Reproducibility}
\label{app:repro}
The dataset is available on \url{https://huggingface.co/datasets/shenjunhao/mmdm}, and the evaluation toolkit is on \url{https://github.com/mmdm-benchmark/mmdm}. The repository's quick-start guide provides installation steps and model configurations. The toolkit can download the dataset or read a local JSONL file, resume cached runs, and export predictions and metrics. The \url{https://mmdm-benchmark.github.io/} hosts the leaderboard and documentation.

%% file: tables/current-taxonomy.tex
\begin{table*}[t]\centering\small\setlength{\tabcolsep}{4pt}\renewcommand{\arraystretch}{1.16}
\begin{tabular}{>{\raggedright\arraybackslash}p{.24\textwidth}>{\raggedright\arraybackslash}p{.72\textwidth}}\toprule
Category & Capability and example \\ \midrule
\textbf{WKN} World knowledge & Apply physical, social, scientific, and general knowledge to a situated request. Options distinguish plausible causes, intentions, continuations, or facts. Example~\ref{ex:world_knowledge}. \\
\textbf{RDC} Reading comprehension & Identify and connect evidence within passages, long documents, or multiple documents. Includes answerability, cross-paragraph inference, and multi-hop evidence composition. Example~\ref{ex:reading_comprehension}. \\
\textbf{DED} Deductive reasoning & Derive what follows from stated premises and rules, including entailment, quantifier scope, and natural-language logical arguments. Example~\ref{ex:deductive_reasoning}. \\
\textbf{OPL} Option logic & Resolve propositions that refer to one another inside option criteria. Select either the unique complete interpretation or every true proposition under a consistent assignment. Example~\ref{ex:option_logic}. \\
\textbf{CPL} Constraint planning & Identify complete plans satisfying interacting eligibility, resource, scheduling, and precedence constraints. Several plans can be feasible. Example~\ref{ex:constraint_planning}. \\
\textbf{QTR} Quantitative and temporal reasoning & Apply practical arithmetic, table and financial evidence, temporal ordering, numeric thresholds, and freshness requirements to a decision. Example~\ref{ex:quantitative_temporal_reasoning}. \\
\textbf{WFD} Workflow decision & Compare complete ordered workflows under tool preconditions, artifact revisions, and certificate validity. Each option is a workflow of up to seven operations; catalogs contain at most 128 options. Example~\ref{ex:workflow_decision}. \\
\textbf{AST} Agent state tracking & Reconstruct goal scope, artifact versions, transaction effects, evidence readiness, and execution failures from events and receipts. Example~\ref{ex:agent_state_tracking}. \\
\textbf{PEX} Policies and exceptions & Apply a policy to a concrete case, resolving exceptions, scope, authority, and effective versions; includes context-sensitive moderation. Example~\ref{ex:policy_exception}. \\
\textbf{EVD} Evidence validation & Assess whether evidence supports, refutes, or leaves a claim unresolved, whether further retrieval is needed, and whether a response meets explicit requirements. Example~\ref{ex:evidence_validation}. \\
\textbf{CDB} Code and databases & Determine program behavior, compare implementations, or select SQL with the required semantics using code, schema, and database context. Example~\ref{ex:code_database}. \\
\textbf{IRT} Intent routing & Identify an intent or topic and select the appropriate domain or retrieval strategy from the options supplied for the request. Example~\ref{ex:intent_routing}. \\
\textbf{TSS} Tool and skill selection & Select the next applicable tool or skill from its name and description, given the current request or observed agent history. Multiple independently applicable tools or skills form a valid set. Example~\ref{ex:tool_skill_selection}. \\
\textbf{APF} Answer preference & Compare responses against correctness, relevance, reasoning quality, instruction following, or stated preference criteria. Example~\ref{ex:answer_preference}. \\
\textbf{SCR} Score and rubrics & Assign a rubric level or a distribution over levels from explicit evidence. Covers deterministic grading and uncertain outcomes under a specified sampling mechanism. Example~\ref{ex:score_rubric}. \\
\textbf{EMR} Emotion recognition & Identify the emotion conveyed by a text or interaction, including context-dependent and closely related emotional categories. Example~\ref{ex:emotion_recognition}. \\
\bottomrule\end{tabular}
\caption{Task definitions and direct links to worked examples with saved model responses in Appendix~\ref{app:examples}.}\label{tab:taxonomy}\end{table*}

%% file: tables/current-public-sources.tex
\begin{table*}[p]\centering\small\setlength{\tabcolsep}{4pt}\renewcommand{\arraystretch}{1.10}
\caption{Public sources used in the final release. Counts refer to released MMDM questions; codes follow Table~\ref{tab:categoryall}. Catalog sources ground new requests.}\label{tab:publicsources}
\begin{tabular}{>{\raggedright\arraybackslash}p{.34\textwidth}r>{\raggedright\arraybackslash}p{.20\textwidth}>{\raggedright\arraybackslash}p{.32\textwidth}}\toprule
\textbf{Source} & \textbf{Items} & \textbf{Categories} & \textbf{Adaptation role} \\ \midrule
Adaptive-RAG~\citep{jeong2024adaptive} & 267 & IRT & Route to a retrieval strategy \\
AG News~\citep{zhang2015character} & 18 & IRT & Topic or intent mapped to named options \\
ARC~\citep{clark2018arc} & 3 & WKN & Knowledge and contextual answer selection \\
BANKING77~\citep{casanueva2020banking} & 2 & IRT & Topic or intent mapped to named options \\
BFCL V4~\citep{patil2025bfcl} & 4 & TSS & Tool names and contracts as options \\
BIRD Mini-Dev~\citep{li2023bird} & 80 & CDB & SQL decisions with database evidence \\
BoolQ~\citep{clark2019boolq} & 52 & RDC & Document evidence and comprehension \\
CodeContests~\citep{li2022alphacode} & 45 & CDB & Code behavior and implementation decisions \\
ComplexFuncBench~\citep{zhong2025complexfuncbench} & 167 & TSS & Observed agent-history decision points \\
ContractNLI~\citep{koreeda2021contractnli} & 128 & PEX & Premise--claim or rule-based judgments \\
CosmosQA~\citep{huang2019cosmosqa} & 75 & WKN & Knowledge and contextual answer selection \\
CRUXEval~\citep{gu2024cruxeval} & 60 & CDB & Code behavior and implementation decisions \\
Emotion~\citep{saravia2018emotion} & 77 & EMR & Emotion labels mapped to options \\
Enron-Spam~\citep{metsis2006spam} & 15 & EVD & Safety, spam, or action-risk decisions \\
FinQA~\citep{chen2021finqa} & 6 & QTR & Practical arithmetic and table-based reasoning \\
FreshQA~\citep{vu2024freshllms} & 165 & QTR & Dated evidence: refresh or reject a premise \\
GoEmotions~\citep{demszky2020goemotions} & 101 & EMR & Emotion labels mapped to options \\
Graph of Skills~\citep{liu2026graphskills} & 47 & TSS & Skill catalog grounding for new requests \\
HellaSwag~\citep{zellers2019hellaswag} & 15 & WKN & Knowledge and contextual answer selection \\
HelpSteer2~\citep{wang2024helpsteer2} & 70 & APF & Compare responses under source criteria \\
HoVer~\citep{jiang2020hover} & 39 & RDC & Document evidence and comprehension \\
HumanEval+~\citep{liu2023evalplus} & 1 & CDB & Code behavior and implementation decisions \\
IF-Multi (verifiable instructions)~\citep{pyatkin2025generalizing} & 15 & EVD & Convert verifiable instruction constraints \\
JudgeBench~\citep{tan2025judgebench} & 36 & APF & Compare responses under source criteria \\
LogiQA 2.0~\citep{liu2023logiqa} & 141 & DED & Premise--claim or rule-based judgments \\
LongBench Pro~\citep{chen2026longbenchpro} & 3 & RDC & Document evidence and comprehension \\
LongBench v2~\citep{bai2025longbenchv2} & 164 & AST, CDB, RDC & Document evidence and comprehension \\
MASSIVE~\citep{fitzgerald2023massive} & 44 & IRT & Topic or intent mapped to named options \\
MathQA~\citep{amini2019mathqa} & 50 & QTR & Practical arithmetic and table-based reasoning \\
MMLU (auxiliary)~\citep{hendrycks2021mmlu} & 6 & WKN & Knowledge and contextual answer selection \\
MMLU-Pro~\citep{wang2024mmlupro} & 3 & WKN & Knowledge and contextual answer selection \\
MuSiQue~\citep{trivedi2022musique} & 18 & RDC & Document evidence and comprehension \\
OpenBookQA~\citep{mihaylov2018openbookqa} & 10 & WKN & Knowledge and contextual answer selection \\
PIQA~\citep{bisk2020piqa} & 28 & WKN & Knowledge and contextual answer selection \\
QASC~\citep{khot2020qasc} & 23 & WKN & Knowledge and contextual answer selection \\
QuALITY~\citep{pang2022quality} & 156 & RDC & Document evidence and comprehension \\
RetrievalQA~\citep{zhang2024retrievalqa} & 154 & EVD & Decide whether external retrieval is needed \\
\bottomrule
\multicolumn{4}{r}{\footnotesize\textit{Continued on next page}} \\
\end{tabular}\end{table*}
\begin{table*}[t]\centering\small\setlength{\tabcolsep}{4pt}\renewcommand{\arraystretch}{1.10}
\ContinuedFloat
\caption{Public sources used in the final release (continued).}\label{tab:publicsourcescontinued}
\begin{tabular}{>{\raggedright\arraybackslash}p{.34\textwidth}r>{\raggedright\arraybackslash}p{.20\textwidth}>{\raggedright\arraybackslash}p{.32\textwidth}}\toprule
\textbf{Source} & \textbf{Items} & \textbf{Categories} & \textbf{Adaptation role} \\ \midrule
RewardBench 2~\citep{malik2026rewardbench} & 68 & APF & Compare responses under source criteria \\
RM-Bench~\citep{liu2025rmbench} & 59 & APF & Compare responses under source criteria \\
RuleTaker~\citep{clark2020ruletaker} & 73 & DED & Premise--claim or rule-based judgments \\
SciQ~\citep{welbl2017sciq} & 12 & WKN & Knowledge and contextual answer selection \\
SkillsBench~\citep{li2026skillsbench} & 18 & TSS & Task/environment context and skill descriptions \\
SNLI~\citep{bowman2015snli} & 57 & DED & Premise--claim or rule-based judgments \\
SocialIQA~\citep{sap2019socialiqa} & 76 & WKN & Knowledge and contextual answer selection \\
Spider~\citep{yu2018spider} & 12 & CDB & SQL decisions with database evidence \\
TAT-QA~\citep{zhu2021tatqa} & 3 & QTR & Practical arithmetic and table-based reasoning \\
Tool-call risk~\citep{moore2026toolrisk} & 3 & TSS & Safety, spam, or action-risk decisions \\
ToolRet~\citep{shi2025toolret} & 14 & TSS & Tool names and contracts as options \\
ToolTalk~\citep{farn2023tooltalk} & 72 & TSS & Observed agent-history decision points \\
ToxicChat~\citep{lin2023toxicchat} & 65 & PEX & Safety, spam, or action-risk decisions \\
WinoGrande~\citep{sakaguchi2020winogrande} & 22 & WKN & Knowledge and contextual answer selection \\
XNLI~\citep{conneau2018xnli} & 38 & EVD & Premise--claim or rule-based judgments \\
\midrule\textbf{Total from public sources} & \textbf{2,880} & & \\
\bottomrule
\end{tabular}\end{table*}

%% file: tables/current-original-sources.tex
\begin{table*}[t]\centering\small\setlength{\tabcolsep}{4pt}\renewcommand{\arraystretch}{1.16}
\begin{tabular}{>{\raggedright\arraybackslash}p{.24\textwidth}r>{\raggedright\arraybackslash}p{.65\textwidth}}\toprule
Construction family & Items & Mechanism and reference derivation \\ \midrule
Scenario-based decision constructions & 260 & Concrete policy, evidence, tool, planning, and scoring scenarios; explicit finite outcome models for soft targets. \\
Compositional rule and evidence programs & 157 & Thresholds, exceptions, versions, evidence joins, and output constraints; computed targets and verified natural-language renderings. \\
Single-select logical interpretations & 152 & Enumerate assignments to mutually referring propositions; each option gives a complete interpretation. \\
Interdependent reports & 359 & Enumerate the consistent assignment and select its true propositions; cross-check with an independent truth-table implementation. \\
Constraint-based plans & 240 & Enumerate proposed schedules and resource assignments; evaluate every feasibility condition. \\
Version-sensitive workflows & 250 & Simulate prerequisites, edits, certificates, and action order; independently verify positions and version requirements. \\
Context-dependent state programs & 441 & Track goal scope, current artifacts, tool outcomes, evidence readiness, and error causes using event receipts. \\
Transactional state histories & 80 & Fold updates, commits, and rollbacks; cross-check the resulting state by independent SQLite replay. \\
Sampled rubric judgments & 598 & Evaluate rubric conditions over sampled records; obtain the full score distribution by exact enumeration and combinatorial verification. \\
Catalog-grounded skill requests & 31 & Generate requests grounded in a collected skill directory; review semantically close options and preserve skill provenance. \\
\midrule\textbf{Total original constructions} & \textbf{2,568} & \\
\bottomrule\end{tabular}
\caption{Original question families, named by their construction mechanism. Catalog-grounded requests retain the provenance of their skill descriptions.}\label{tab:originalsources}\end{table*}

%% file: tables/appendix-current-models.tex
\begin{table*}[t]\centering\scriptsize\setlength{\tabcolsep}{4pt}
\begin{tabular}{p{.47\textwidth}rrlrr}\toprule
Decision Model and release reference & Context budget & Cap & Counter & Native & Grouped \\ \midrule
Autotrust JEV 27B-VL~\citep{autotrustailab2026jev} & 65536 & 256 & provider & 5380 & 68 \\
Clef~\citep{cloudflare2026clef} & 65536 & 255 & proxy & 5354 & 94 \\
Clef-Flash~\citep{cloudflare2026clef} & 24576 & 255 & proxy & 5422 & 26 \\
Clef-Omni~\citep{cloudflare2026clef} & 65536 & 255 & proxy & 5354 & 94 \\
Decisio-Gemma-4-12B~\citep{gemmateam2026gemma4,roudaki2026decisio} & 32768 & 255 & provider & 5396 & 52 \\
Decisio-Gemma-4-31B~\citep{gemmateam2026gemma4,roudaki2026decisio} & 32768 & 255 & provider & 5396 & 52 \\
Decisio-Qwen3.6-35B-A3B~\citep{qwenteam2026qwen36,roudaki2026decisio} & 32768 & 255 & provider & 5396 & 52 \\
GPT-6 Luna Decisions~\citep{openai2026decisions} & Full input & 255 & provider & 5354 & 94 \\
H2O-Lightning 4B~\citep{h2oai2026lightning} & 32000 & 255 & provider & 5396 & 52 \\
Intern-Decision 0.8B~\citep{zhu2026intern} & 32768 & 62 & provider & 5186 & 262 \\
Intern-Decision 2B~\citep{zhu2026intern} & 32768 & 62 & provider & 5186 & 262 \\
Intern-Decision 4B~\citep{zhu2026intern} & 32768 & 62 & provider & 5186 & 262 \\
Jev 1.14.0~\citep{typesafeai2026jev} & 262144 & 255 & proxy & 5354 & 94 \\
JevK5 4B~\citep{serikbay2026jevk5} & 32768 & 16 & provider & 4908 & 540 \\
KnowLine 31B Gen1~\citep{pelaai2026knowline} & 262144 & 255 & provider & 5354 & 94 \\
Liquid-d1~\citep{liquidai2026d1} & 65536 & 255 & proxy & 5354 & 94 \\
Liquid-d1-Ultra~\citep{liquidai2026d1} & 32768 & 255 & proxy & 5396 & 52 \\
Mapika Decider 4B~\citep{marosi2026decider} & 32768 & 255 & provider & 5396 & 52 \\
Microsoft Decision-1~\citep{microsoft2026decision} & 32768 & 255 & proxy & 5396 & 52 \\
NanoJev 0.6B~\citep{chen2026nanojev} & 32768 & 100000 & provider & 5448 & 0 \\
OpenJev 4B~\citep{shen2026openjev} & 65536 & 100000 & provider & 5448 & 0 \\
Perplexity Decider-Flash-V1.2~\citep{perplexityai2026decider} & 262144 & 255 & proxy & 5354 & 94 \\
Perplexity Decider-V1~\citep{perplexityai2026decider} & 8192 & 255 & provider & 5448 & 0 \\
Perplexity Decider-V1.2~\citep{perplexityai2026decider} & 262144 & 255 & proxy & 5354 & 94 \\
Quyet-1.0-Large~\citep{nguyen2026quyet} & 8000 & 10 & provider & 4552 & 896 \\
SemIf 4B~\citep{lee2026semif} & 32768 & 16 & provider & 4908 & 540 \\
Strands-Decider 2B~\citep{strandsagents2026decider} & 32768 & 100000 & provider & 5448 & 0 \\
WeMM-JEV 27B~\citep{wemmjevteam2026wemm} & 262144 & 255 & provider & 5354 & 94 \\
\bottomrule\end{tabular}
\caption{Evaluated decision models. \emph{Full input} indicates no evaluator-side truncation. Context budgets and option caps are run settings; the full run identity pins service/model versions, tokenizers, adapters, and reserves. A cap of 100,000 denotes the configured effectively uncapped native path, not a measured architectural limit.}\label{tab:models}\end{table*}

%% file: tables/current-category-distributions.tex
\begin{table*}[t]\centering
\fontsize{8}{9.5}\selectfont\setlength{\tabcolsep}{3pt}\renewcommand{\arraystretch}{1.05}
\resizebox{\textwidth}{!}{\begin{tabular}{l | rrrrr | rrrrrr}\toprule
 & \multicolumn{5}{c|}{\textbf{Hard Multi-select Accuracy (\%, $\uparrow$)}} & \multicolumn{6}{c}{\textbf{Soft Brier ($\downarrow$)}} \\
\cmidrule(lr){2-6}\cmidrule(lr){7-12}
\textbf{Decision Model} & \textbf{RDC} & \textbf{OPL} & \textbf{CPL} & \textbf{WFD} & \textbf{TSS} & \textbf{CPL} & \textbf{QTR} & \textbf{PEX} & \textbf{EVD} & \textbf{TSS} & \textbf{SCR} \\ \midrule
\multicolumn{12}{c}{\textit{Heuristic reference}} \\ \midrule
Random (uniform) & 6.67 & 4.40 & 0.61 & 0.59 & 3.71 & 0.0026 & 0.0821 & 0.0334 & 0.1039 & 0.0816 & 0.2206 \\
\midrule
\multicolumn{12}{c}{\textit{Closed-source decision models}} \\ \midrule
Jev 1.14.0 & \mmdmbestcell{100.00} & 5.01 & 22.50 & 21.35 & 80.58 & 0.0425 & \mmdmbestcell{0.0453} & 0.0374 & \mmdmbestcell{0.0275} & 0.0937 & \mmdmbestcell{0.1170} \\
Microsoft Decision-1 & \mmdmsecondcell{50.00} & 10.86 & 8.75 & 16.85 & 74.76 & 0.2496 & 0.1475 & 0.1133 & 0.1371 & 0.1647 & 0.2959 \\
Liquid-d1-Ultra & \mmdmbestcell{100.00} & 6.13 & 18.75 & 29.21 & \mmdmsecondcell{86.41} & 0.0073 & 0.1663 & 0.0948 & 0.0851 & 0.1584 & 0.1334$^{626}$ \\
Liquid-d1 & 0.00 & 7.52 & 21.25 & 19.10 & 75.73 & 0.1689 & 0.2351 & 0.1080 & 0.1886 & 0.2124 & 0.2946$^{625}$ \\
GPT-6 Luna Decisions & \mmdmsecondcell{50.00} & 3.06 & 6.25 & 6.74 & 51.46 & 0.1994 & 0.2015 & 0.1112 & 0.1562$^{20}$ & 0.1348$^{21}$ & 0.3548$^{627}$ \\
\midrule
\multicolumn{12}{c}{\textit{Open-source decision models}} \\ \midrule
WeMM-JEV 27B & \mmdmbestcell{100.00} & \mmdmbestcell{20.61} & 26.25 & \mmdmbestcell{47.19} & \mmdmbestcell{91.26} & 0.1950 & 0.1477 & 0.1102 & 0.1227 & 0.2204 & 0.2422 \\
KnowLine 31B Gen1 & \mmdmbestcell{100.00} & 14.48 & \mmdmbestcell{35.00} & \mmdmsecondcell{40.45} & 84.47 & 0.1332 & 0.1466 & 0.0672 & 0.0682 & 0.1217 & \mmdmsecondcell{0.1265} \\
Perplexity Decider-V1.2 & \mmdmbestcell{100.00} & \mmdmsecondcell{19.78} & 25.00 & 37.08 & 85.44 & 0.3949 & 0.0737 & 0.0776 & \mmdmsecondcell{0.0632} & 0.1840 & 0.1767$^{626}$ \\
Autotrust JEV 27B-VL & \mmdmbestcell{100.00} & 9.47 & 8.75 & 13.48 & 70.87 & 0.1783 & 0.2352 & 0.0428 & 0.1658 & 0.1331 & 0.3647 \\
Decisio-Gemma-4-31B & \mmdmbestcell{100.00} & 5.57 & 17.50 & 10.11 & 50.49 & 0.1046 & 0.3715 & 0.3445 & 0.2181 & 0.2452 & 0.1516 \\
Clef & \mmdmbestcell{100.00} & 5.01 & 15.00 & 8.99 & 62.14 & 0.0438 & 0.1101 & 0.1418 & 0.1916 & 0.1686 & 0.1403 \\
Perplexity Decider-V1 & \mmdmsecondcell{50.00} & 10.86 & 17.50 & 5.62 & 74.76 & 0.0357 & 0.2022 & 0.0715 & 0.0671 & 0.1062 & 0.1621 \\
Perplexity Decider-Flash-V1.2 & \mmdmbestcell{100.00} & 3.90 & 5.00 & 11.24 & 70.87 & 0.0744 & 0.1575 & \mmdmbestcell{0.0331} & 0.1470 & 0.0937 & 0.3050$^{625}$ \\
H2O-Lightning 4B & \mmdmsecondcell{50.00} & 11.42 & 12.50 & 10.11 & 66.99 & 0.0103 & 0.1128 & 0.1219 & 0.2102 & 0.1349 & 0.3720 \\
Decisio-Gemma-4-12B & \mmdmbestcell{100.00} & 4.46 & 22.50 & 10.11 & 67.96 & 0.0169 & 0.2140 & 0.1612 & 0.1034 & 0.1493 & 0.1646 \\
Quyet-1.0-Large & \mmdmsecondcell{50.00} & 4.18 & \mmdmsecondcell{30.00} & 14.61 & 50.49 & 0.2222 & 0.1660 & 0.1563 & 0.1003$^{21}$ & 0.1375$^{19}$ & 0.1853 \\
OpenJev 4B & 0.00 & 5.29 & 18.75 & 12.36 & 63.11 & 0.0070 & 0.2168 & 0.0423 & 0.1066 & \mmdmbestcell{0.0681} & 0.4012 \\
Clef-Omni & \mmdmbestcell{100.00} & 4.18 & 16.25 & 7.87 & 55.34 & 0.0470 & 0.2112 & 0.0780 & 0.2457$^{20}$ & 0.1504$^{21}$ & 0.2886$^{625}$ \\
Mapika Decider 4B & \mmdmsecondcell{50.00} & 5.57 & 8.75 & 8.99 & 51.46 & 0.1338 & 0.1100 & 0.1128 & 0.1517 & 0.1366 & 0.2229 \\
Clef-Flash & \mmdmbestcell{100.00} & 3.90 & 12.50 & 10.11 & 57.28 & 0.0286 & 0.2906 & 0.0722 & 0.1677 & 0.1227 & 0.3799 \\
Decisio-Qwen3.6-35B-A3B & \mmdmbestcell{100.00} & 6.13 & 8.75 & 11.24 & 59.22 & 0.2216 & 0.1718 & 0.1177 & 0.1222 & 0.1128 & 0.2706 \\
Intern-Decision 4B & \mmdmsecondcell{50.00} & 4.46 & 5.00 & 4.49 & 43.69 & 0.0258 & 0.1743 & 0.0690 & 0.0883 & \mmdmsecondcell{0.0836} & 0.1637 \\
JevK5 4B & \mmdmsecondcell{50.00} & 5.01 & 7.50 & 1.12 & 38.83 & 0.0251 & 0.1283 & 0.0916 & 0.0773 & 0.1304 & 0.2441 \\
Strands-Decider 2B & \mmdmsecondcell{50.00} & 3.34 & 7.50 & 1.12 & 31.07 & 0.0819 & 0.1483 & 0.0719 & 0.0822 & 0.1078 & 0.2181 \\
SemIf 4B & \mmdmsecondcell{50.00} & 3.06 & 2.50 & 0.00 & 25.24 & \mmdmsecondcell{0.0026} & 0.2521 & 0.0952 & 0.1107 & 0.1829 & 0.3630 \\
Intern-Decision 2B & \mmdmsecondcell{50.00} & 3.34 & 1.25 & 0.00 & 23.30 & \mmdmbestcell{0.0001} & 0.1347 & 0.0559 & 0.0954 & 0.1117 & 0.2137 \\
Intern-Decision 0.8B & \mmdmsecondcell{50.00} & 3.06 & 0.00 & 1.12 & 11.65 & 0.0037 & \mmdmsecondcell{0.0727} & \mmdmsecondcell{0.0362} & 0.0987 & 0.0909 & 0.2205 \\
NanoJev 0.6B & 0.00 & 10.86 & 0.00 & 0.00 & 5.83 & 0.0616 & 0.1626 & 0.0585 & 0.1885 & 0.2123 & 0.3237 \\
\bottomrule\end{tabular}}
\caption{Category results for multi-select and soft-label questions. RDC denotes Reading Comprehension; OPL, Option Logic; CPL, Constraint Planning; WFD, Workflow Decision; TSS, Tool and Skill Selection; QTR, Quantitative and Temporal Reasoning; PEX, Policies and Exceptions; EVD, Evidence Validation; and SCR, Score and Rubrics. Multi-select totals are 2, 359, 80, 89, and 103 in column order; soft-label totals are 1, 12, 18, 22, 22, and 628. Soft Brier uses valid native outputs; superscripts give counts below the category total. Colors mark the best and second-best model values within each column, as in Table~\ref{tab:overall}.}\label{tab:categorydistributions}
\end{table*}

%% file: tables/appendix-current-matched.tex
\begin{table*}[t]\centering\small
\begin{tabular}{lrrrrrr}\toprule
Decision Model & \shortstack{Hard Single-select\\Acc.$\uparrow$} & \shortstack{Hard Multi-select\\Acc.$\uparrow$} & Soft Brier$\downarrow$ & Score nMAE$\downarrow$ & SB-F$\uparrow$ & ECE$\downarrow$ \\ \midrule
Perplexity V1.2 & 76.66 & 36.90 & 0.1694 & 0.1641 & 0.1990 & 0.046 \\
Jev 1.14.0 & 68.73 & 22.75 & 0.1101 & 0.0927 & 0.1198 & 0.054 \\
Microsoft Decision-1 & 64.27 & 24.67 & 0.2794 & 0.1836 & 0.1260 & 0.125 \\
Intern-Decision 4B & 51.70 & 11.85 & 0.1565 & 0.1450 & 0.0845 & 0.085 \\
OpenJev 4B & 61.82 & 18.93 & 0.3685 & 0.3105 & 0.1183 & 0.092 \\
\bottomrule\end{tabular}
\caption{Results on shared native, full-state inputs. Cohorts contain 3,706 single-select, 523 multi-select, 701 soft-label, and 661 Score questions; ECE uses 3,537 single-select questions with unique maxima. Accuracies are percentages. Each metric uses the same questions across models; SB-F uses the 523-question multi-select cohort.}\label{tab:matched}\end{table*}

%% file: tables/appendix-current-set.tex
\begin{table}[!t]\centering\fontsize{8}{9.5}\selectfont\setlength{\tabcolsep}{3pt}
\begin{tabular}{lrrrr}\toprule \textbf{Decision Model} & \textbf{OPL} & \textbf{CPL} & \textbf{WFD} & \textbf{TSS} \\ \midrule
Perplexity V1.2 & 19.78 & 25.00 & 37.08 & 85.44 \\
Jev 1.14.0 & 5.01 & 22.50 & 21.35 & 80.58 \\
Microsoft Decision-1 & 10.86 & 8.75 & 16.85 & 74.76 \\
Intern-Decision 4B & 4.46 & 5.00 & 4.49 & 43.69 \\
OpenJev 4B & 5.29 & 18.75 & 12.36 & 63.11 \\
\bottomrule\end{tabular}
\caption{Hard Multi-select Accuracy (\%, $\uparrow$) for representative models. OPL denotes Option Logic (359 questions); CPL, Constraint Planning (80); WFD, Workflow Decision (89); and TSS, Tool and Skill Selection (103). The two Reading Comprehension questions are reported in Table~\ref{tab:categorydistributions}.}\label{tab:setcategories}\end{table}

%% file: tables/appendix-current-set-modes.tex
\begin{table}[!t]\centering\fontsize{8}{9.5}\selectfont\setlength{\tabcolsep}{3pt}
\begin{tabular}{lrrrr}\toprule
 & \multicolumn{2}{c}{Native} & \multicolumn{2}{c}{Grouped} \\
Decision Model & Items & Acc. & Items & Acc. \\ \midrule
Perplexity V1.2 & 605 & 35.04 & 28 & 7.14 \\
Jev 1.14.0 & 605 & 22.81 & 28 & 7.14 \\
Microsoft Decision-1 & 611 & 22.75 & 22 & 0.00 \\
Intern-Decision 4B & 541 & 12.75 & 92 & 1.09 \\
OpenJev 4B & 633 & 17.38 & 0 & -- \\
\bottomrule\end{tabular}
\caption{Hard Multi-select Accuracy (\%, $\uparrow$) by inference mode under Section~\ref{sec:setup}. Failures count as incorrect. Native and grouped cohorts are determined by each model\textquotesingle s option cap and therefore differ across models; -- denotes an empty cohort.}\label{tab:setmodes}\end{table}

%% file: tables/current-sbf.tex
\begin{table*}[t]\centering
\fontsize{8}{9.5}\selectfont\setlength{\tabcolsep}{4pt}\renewcommand{\arraystretch}{1.02}
\begin{tabular*}{\textwidth}{@{\extracolsep{\fill}}l r r | r rr | rrrr@{}}\toprule
 & & & \multicolumn{3}{c|}{\textbf{Overall}} & \multicolumn{4}{c}{\textbf{SB-F by category}} \\
\cmidrule(lr){4-6}\cmidrule(lr){7-10}
\textbf{Decision Model} & $n$ & Rank & SB-F$\uparrow$ & $\overline S$ & $\overline B$ & \textbf{OPL} & \textbf{CPL} & \textbf{WFD} & \textbf{TSS} \\ \midrule
\multicolumn{10}{c}{\textit{Heuristic reference}} \\ \midrule
Random (uniform) & 633 & --- & 0.0000 & 0.000 & 1.000 & 0.000 & 0.000 & 0.000 & 0.000 \\
\midrule
\multicolumn{10}{c}{\textit{Closed-source decision models}} \\ \midrule
Jev 1.14.0 & 605 & 11 & 0.1057 & 0.137 & 0.915 & 0.022 & 0.244 & 0.261 & \mmdmsecondcell{0.465} \\
Microsoft Decision-1 & 601 & 8 & 0.1194 & 0.169 & 0.812 & 0.069 & 0.186 & 0.212 & 0.339 \\
Liquid-d1-Ultra & 604 & 4 & 0.1521 & 0.192 & 0.801 & 0.056 & 0.304 & 0.399 & \mmdmbestcell{0.527} \\
Liquid-d1 & 604 & 9 & 0.1183 & 0.156 & 0.848 & 0.049 & 0.226 & 0.262 & 0.411 \\
GPT-6 Luna Decisions & 596 & 21 & 0.0800 & 0.112 & 0.838 & 0.043 & 0.151 & 0.105 & 0.260 \\
\midrule
\multicolumn{10}{c}{\textit{Open-source decision models}} \\ \midrule
WeMM-JEV 27B & 605 & 1 & \mmdmbestcell{0.1905} & 0.259 & 0.780 & \mmdmbestcell{0.123} & \mmdmbestcell{0.516} & \mmdmsecondcell{0.408} & 0.342 \\
KnowLine 31B Gen1 & 605 & 3 & 0.1795 & 0.258 & 0.678 & 0.102 & \mmdmsecondcell{0.500} & 0.335 & 0.429 \\
Perplexity Decider-V1.2 & 602 & 2 & \mmdmsecondcell{0.1900} & 0.244 & 0.808 & \mmdmsecondcell{0.110} & 0.386 & \mmdmbestcell{0.436} & 0.450 \\
Autotrust JEV 27B-VL & 623 & 6 & 0.1242 & 0.164 & 0.770 & 0.064 & 0.239 & 0.186 & 0.405 \\
Decisio-Gemma-4-31B & 605 & 12 & 0.0959 & 0.126 & 0.838 & 0.046 & 0.168 & 0.115 & 0.356 \\
Clef & 601 & 15 & 0.0948 & 0.109 & 0.904 & 0.036 & 0.219 & 0.113 & 0.384 \\
Perplexity Decider-V1 & 593 & 5 & 0.1290 & 0.158 & 0.874 & 0.067 & 0.244 & 0.131 & 0.453 \\
Perplexity Decider-Flash-V1.2 & 600 & 17 & 0.0853 & 0.112 & 0.915 & 0.018 & 0.118 & 0.190 & 0.416 \\
H2O-Lightning 4B & 605 & 13 & 0.0956 & 0.114 & 0.923 & 0.029 & 0.189 & 0.158 & 0.421 \\
Decisio-Gemma-4-12B & 605 & 18 & 0.0851 & 0.113 & 0.891 & 0.027 & 0.216 & 0.106 & 0.371 \\
Quyet-1.0-Large & 461 & 14 & 0.0953 & 0.135 & 0.869 & 0.023 & 0.369 & 0.178 & 0.373 \\
OpenJev 4B & 620 & 10 & 0.1102 & 0.155 & 0.693 & 0.060 & 0.192 & 0.125 & 0.368 \\
Clef-Omni & 601 & 22 & 0.0754 & 0.090 & 0.916 & 0.031 & 0.158 & 0.147 & 0.270 \\
Mapika Decider 4B & 605 & 20 & 0.0800 & 0.112 & 0.845 & 0.030 & 0.187 & 0.137 & 0.303 \\
Clef-Flash & 601 & 19 & 0.0800 & 0.095 & 0.899 & 0.039 & 0.111 & 0.144 & 0.276 \\
Decisio-Qwen3.6-35B-A3B & 605 & 7 & 0.1202 & 0.157 & 0.762 & 0.059 & 0.144 & 0.163 & 0.452 \\
Intern-Decision 4B & 535 & 16 & 0.0880 & 0.110 & 0.854 & 0.038 & 0.135 & 0.122 & 0.350 \\
JevK5 4B & 501 & 23 & 0.0738 & 0.093 & 0.880 & 0.032 & 0.128 & 0.062 & 0.310 \\
Strands-Decider 2B & 605 & 26 & 0.0588 & 0.069 & 0.877 & 0.030 & 0.142 & 0.034 & 0.214 \\
SemIf 4B & 501 & 24 & 0.0716 & 0.105 & 0.718 & 0.052 & 0.069 & 0.040 & 0.208 \\
Intern-Decision 2B & 535 & 25 & 0.0589 & 0.072 & 0.874 & 0.033 & 0.062 & 0.034 & 0.226 \\
Intern-Decision 0.8B & 535 & 28 & 0.0363 & 0.045 & 0.870 & 0.034 & 0.005 & 0.014 & 0.075 \\
NanoJev 0.6B & 605 & 27 & 0.0429 & 0.047 & 0.915 & 0.054 & 0.009 & 0.005 & 0.013 \\
\bottomrule\end{tabular*}
\caption{SB-F on a fixed common cohort of 284 native multi-select predictions per model. $n$ gives the valid native multi-select sample count used for each model in Table~\ref{tab:overall}; the statistics here use the common cohort. Ranks compare the 28 decision models using unrounded SB-F. OPL denotes Option Logic (214 questions); CPL, Constraint Planning (14); WFD, Workflow Decision (20); and TSS, Tool and Skill Selection (36). $\overline S$ and $\overline B$ are mean separation and correct-set balance; SB-F averages their per-question products. Colors mark the best and second-best SB-F within each column.}\label{tab:sbf}
\end{table*}

%% file: tables/current-bootstrap.tex
On the shared single-select questions, Perplexity V1.2 exceeds Jev by 7.93 percentage points (95\% CI [6.41, 9.44]), while Jev's Soft Brier is lower by 0.0592 on the shared soft-label questions (95\% CI [0.0515, 0.0661]). We estimate both confidence intervals from 2,000 paired bootstrap resamples of source-parent groups, computing item-weighted means within each resample.

%% file: tables/current-multiselect-diagnostics.tex
\begin{table*}[t]\centering
\small\setlength{\tabcolsep}{5pt}\renewcommand{\arraystretch}{1.08}
\begin{tabular*}{\textwidth}{@{\extracolsep{\fill}}l rrr r@{}}\toprule
\textbf{Decision Model} & \shortstack{Valid top choice\\(\%)} & \shortstack{Multi-select Acc.\\(\%)} & \shortstack{Coverage\\(\%)} & \shortstack{Narrow margins\\(count)} \\ \midrule
Random (uniform) & 51.7 & 4.2 & 0.0 & --- \\
\midrule
WeMM-JEV 27B & 88.7 & 38.7 & 64.7 & 43/110 \\
Perplexity Decider-V1.2 & 84.9 & 34.9 & 59.8 & 30/99 \\
Jev 1.14.0 & 63.7 & 20.1 & 38.2 & 14/57 \\
Microsoft Decision-1 & 71.5 & 23.2 & 45.8 & 24/66 \\
GPT-6 Luna Decisions & 64.8 & 12.7 & 34.8 & 20/36 \\
OpenJev 4B & 74.3 & 19.7 & 42.8 & 24/56 \\
Intern-Decision 4B & 59.5 & 14.1 & 32.9 & 21/40 \\
\bottomrule\end{tabular*}
\caption{Diagnostics on the same 284 native multi-select questions. Valid top choice checks whether the highest-ranked option is correct. Coverage averages the fraction of correct options outranking the strongest distractor. Narrow margins count complete-set successes in which the strongest distractor has at least half the probability of the weakest correct option; the denominator is the number of complete-set successes. Random uses analytical uniform tie-breaking for the first two columns.}\label{tab:multidiagnostics}
\end{table*}

%% file: tables/current-context-diagnostics.tex
\begin{table*}[!t]\centering\small
\setlength{\tabcolsep}{5pt}\renewcommand{\arraystretch}{1.08}
\begin{tabular*}{\textwidth}{@{\extracolsep{\fill}}llrrrr}\toprule
Task & Condition & $n$ & Tokens & Perplexity V1.2 & Jev 1.14.0 \\ \midrule
\multicolumn{6}{l}{\textit{Reading Comprehension, four options, complete state}} \\
Long-document & 16--32K tokens & 49 & 20,393 & 61.2 & 59.2 \\
Long-document & 32--65K tokens & 16 & 40,166 & 62.5 & 68.8 \\
Multi-hop & 16--32K tokens & 19 & 24,520 & 57.9 & 52.6 \\
Multi-hop & 32--65K tokens & 32 & 51,776 & 50.0 & 46.9 \\
\midrule\multicolumn{6}{l}{\textit{Large catalogs, complete state}} \\
Constraint-based plans & 255 options & 22 & 13,118 & 31.8 & 22.7 \\
Skill selection & 255 options & 7 & 22,625 & 100.0 & 100.0 \\
Constraint-based plans & 512 options & 22 & 25,616 & 40.9 & 18.2 \\
Skill selection & 512 options & 12 & 41,592 & 91.7 & 91.7 \\
\bottomrule\end{tabular*}
\caption{Hard Single-select Accuracy (\%, $\uparrow$) by decision requirement within length or option-count strata. Tokens is the median complete-input length. Reading questions and 255-option catalogs use native requests; 512-option catalogs use the same grouping protocol for both models. No state is truncated in these cohorts.}\label{tab:context-diagnostics}
\end{table*}

%% file: sections/examples.tex
\section{Examples and Model Responses}
\label{app:examples}
Examples pair released questions from all 16 categories with saved native predictions. Long inputs are excerpted; evaluation follows Appendix~\ref{app:protocol}. Single-select examples report the chosen option and probability; multi-select examples report the $|G|$ selected options; soft-label examples report the distribution error and most probable level. 

\clearpage
\onecolumn
\raggedbottom
\input{tables/current-examples}

%% file: tables/current-examples.tex
\par\bigskip
\refstepcounter{mmdmcase}\label{ex:world_knowledge}
\begin{mmdmexample}{exampleSage}{1. World knowledge}{mmdm-00553 | Hard Single-select | mmlu\_\allowbreak{}auxiliary}
\textbf{State (display excerpt).} Public schools in the state of Green are financed,\allowbreak{} in large part,\allowbreak{} by revenue derived from real estate taxes imposed by each school district on the taxable real property located in that district. Public schools also receive other revenue from private gifts,\allowbreak{} federal grants,\allowbreak{} student fees,\allowbreak{} and local sales taxes. For many years,\allowbreak{} Green has distributed additional funds,\allowbreak{} which come from the state treasury,\allowbreak{} to local school districts in order to equalize the funds available on a per-student basis for each public school district. These additional funds are distributed on the basis of a state statutory formula that considers only the number of students in each public school district and the real estate tax revenue raised by that district. The formula does not consider other revenue received by a school district from different sources. The school boards of two school districts,\allowbreak{} together with parents and schoolchildren in those districts,\allowbreak{} bring suit in federal court to enjoin the state from allocating the additional funds from the state treasury to individual districts pursuant to this formula. They allege that the failure of the state,\allowbreak{} in allocating this additional money,\allowbreak{} to take into account a school district's sources of revenue other than revenue derived from taxes levied on real estate located there violates the equal protection clause of the Fourteenth Amendment. The complaint does not allege that the allocation of the additional state funds based on the current statutory formula has resulted in a failure to provide minimally adequate education to any child. Which of the following [excerpt ends]\par\medskip
\textbf{Query.} Which option best answers the question above?\par\medskip
\textbf{Options.}\par
\textbf{o0. Because classifications based on wealth are inherently suspect,\allowbreak{} the state must demonstrate that the statutory formula is [excerpt ends]}\par\smallskip
\textbf{o1. Because the funding formula inevitably leads to disparities among the school districts in their levels of total funding,\allowbreak{} the [excerpt ends]}\par\smallskip
\textbf{o2. Because no fundamental right or suspect classification is implicated in this case,\allowbreak{} the plaintiffs must demonstrate that the [excerpt ends]}\par\smallskip
\textbf{o3. Because the statutory funding formula burdens the fundamental right to education,\allowbreak{} the state must demonstrate that the formula is [excerpt ends]}\par\smallskip
\medskip\mmdmgold{\textbf{Reference.} o2}
\medskip\textbf{Saved model responses.}\par\smallskip
\begin{tabular}{p{.29\linewidth}p{.44\linewidth}p{.17\linewidth}}\toprule Model & Prediction & Result \\ \midrule
Perplexity V1.2 & o2 (0.996) & Correct \\
Jev 1.14.0 & o2 (1.000) & Correct \\
Microsoft Decision-1 & o2 (0.978) & Correct \\
\bottomrule\end{tabular}
\end{mmdmexample}
\par\bigskip
\refstepcounter{mmdmcase}\label{ex:reading_comprehension}
\begin{mmdmexample}{exampleBlue}{2. Reading comprehension}{mmdm-v2-candidate-d3e6ddec64dd72d2b52ba175 | Hard Single-select | boolq}
\textbf{State.} In ``The Wars to Come'',\allowbreak{} Stannis enlists Jon as an intermediary between himself and Mance,\allowbreak{} hoping to add the wildling army to his own. Mance refuses to submit to Stannis,\allowbreak{} and the wildling king is burned at the stake by the red priestess Melisandre. In ``The House of Black and White'',\allowbreak{} Stannis offers to legitimize Jon and make him lord of Winterfell in exchange for his support. Jon is voted the new Lord Commander of the Night's Watch,\allowbreak{} initially tying with his nemesis Thorne but having the tie settled by Maester Aemon. Jon's intent to welcome the wildlings into Westeros and grant them lands by the Wall further enrages Thorne's faction of the Watch,\allowbreak{} who hold a deep seated hatred for the wildlings. In ``Hardhome'',\allowbreak{} Jon travels by ship north of the Wall to the eponymous wildling village,\allowbreak{} seeking their support for his plan to ally the Night's Watch and the wildlings against the growing threat of the White Walkers. As some of the wildling clans board Stannis' ships to travel south,\allowbreak{} a massive force of wights,\allowbreak{} led by White Walkers and their Night King,\allowbreak{} attacks the village. Jon kills a White Walker with Longclaw,\allowbreak{} learning that weapons forged with Valyrian steel can destroy them completely. Jon and his group barely make it out alive,\allowbreak{} with only a fraction of the wildling forces. Jon then permits the wildling forces to pass through the Wall into the North. Shortly after returning to Castle Black in the season finale ``Mother's Mercy'',\allowbreak{} Jon is summoned to hear news of his missing uncle Benjen,\allowbreak{} but is instead stabbed to death by Thorne and his group of mutineers.\par\medskip
\textbf{Query.} did jon snow die in season 5 finale\par\medskip
\textbf{Options.}\par
\textbf{o0. No}\par\smallskip
\textbf{o1. Yes}\par\smallskip
\medskip\mmdmgold{\textbf{Reference.} o1}
\medskip\textbf{Saved model responses.}\par\smallskip
\begin{tabular}{p{.29\linewidth}p{.44\linewidth}p{.17\linewidth}}\toprule Model & Prediction & Result \\ \midrule
Perplexity V1.2 & o1 (0.976) & Correct \\
Jev 1.14.0 & o1 (0.990) & Correct \\
Microsoft Decision-1 & o1 (0.991) & Correct \\
\bottomrule\end{tabular}
\end{mmdmexample}
\par\bigskip
\refstepcounter{mmdmcase}\label{ex:deductive_reasoning}
\begin{mmdmexample}{exampleClay}{3. Deductive reasoning}{mmdm-00271 | Hard Single-select | logiqa2}
\textbf{State.} Some music theorists argue that music can arouse pity in the listener. But since pity can be felt only when there is someone or something to be pitied,\allowbreak{} and since nothing but the music is available to be the object of the listener' s pity,\allowbreak{} if anything is pitiable,\allowbreak{} then it is the music itself. But it makes no sense to say that a piece of music is pitiable; so music cannot arouse pity in the listener.\par\medskip
\textbf{Query.} The pattern of reasoning in the argument above is most similar to that in which one of the following arguments?\par\medskip
\textbf{Options.}\par
\textbf{o0. Some science writers imagine what life would be like if humans could inhabit distant planets. Inhabiting these planets,\allowbreak{} they [excerpt ends]}\par\smallskip
\textbf{o1. Some statisticians predict future events by observing past events; this requires extrapolating from the past. Though it is [excerpt ends]}\par\smallskip
\textbf{o2. Some people claim that the quality of one's life can be estimated by summing all the positive aspects of one's life and [excerpt ends]}\par\smallskip
\textbf{o3. Astronomers use color photography to measure the temperatures of stars. Using color photography implies that the object [excerpt ends]}\par\smallskip
\medskip\mmdmgold{\textbf{Reference.} o2}
\medskip\textbf{Saved model responses.}\par\smallskip
\begin{tabular}{p{.29\linewidth}p{.44\linewidth}p{.17\linewidth}}\toprule Model & Prediction & Result \\ \midrule
Perplexity V1.2 & o2 (0.975) & Correct \\
Jev 1.14.0 & o2 (0.830) & Correct \\
Microsoft Decision-1 & o2 (0.901) & Correct \\
\bottomrule\end{tabular}
\end{mmdmexample}
\par\bigskip
\refstepcounter{mmdmcase}\label{ex:option_logic}
\begin{mmdmexample}{exampleMauve}{4. Option logic}{mmdm-v2-f25e88272bd2277feb80ac8c | Hard Multi-select | mmdm\_\allowbreak{}v2\_\allowbreak{}mutual\_\allowbreak{}reports}
\textbf{State.} An investigation in environmental monitoring has preserved one report from each of Birch,\allowbreak{} Linden,\allowbreak{} Willow,\allowbreak{} Maple,\allowbreak{} Orchid,\allowbreak{} Pine,\allowbreak{} Cedar,\allowbreak{} Hazel. The reports concern the truth of other preserved reports,\allowbreak{} not whether the speaker is generally honest. Each complete report is either true or false. A report is true exactly when the claim written in its option criteria is true under the same assignment of truth values to all reports. No independent rule says that a named speaker always lies or always tells the truth. There is no majority-vote rule. Conditional statements have their ordinary logical meaning:\allowbreak{} a conditional with a false premise is true. 'Exactly one' excludes both-true and both-false. The names are identifiers,\allowbreak{} not clues. The archived report texts are given only in the options.\par\medskip
\textbf{Query.} Select all and only the people whose preserved reports are true in the mutually consistent assignment. All true reports have equal membership in the correct set.\par\medskip
\textbf{Options.}\par
\textbf{o1. Linden}. If Maple's report is true,\allowbreak{} then Pine's report is true.\par\smallskip
\textbf{o7. Hazel}. Maple's report is false.\par\smallskip
\textbf{o5. Pine}. At least one of Orchid's and Cedar's reports is true; both may be true.\par\smallskip
\textbf{o3. Maple}. At least one of Pine's and Willow's reports is true; both may be true.\par\smallskip
\textbf{o0. Birch}. Both Pine's and Willow's reports are true.\par\smallskip
\textbf{o2. Willow}. At least one of Orchid's and Hazel's reports is true; both may be true.\par\smallskip
\textbf{o4. Orchid}. At least one of Pine's and Willow's reports is true; both may be true.\par\smallskip
\textbf{o6. Cedar}. At least one of Orchid's and Maple's reports is true; both may be true.\par\smallskip
\medskip\mmdmgold{\textbf{Reference set.} o1,\allowbreak{} o5,\allowbreak{} o3,\allowbreak{} o0,\allowbreak{} o2,\allowbreak{} o4,\allowbreak{} o6}
\medskip\textbf{Saved model responses.}\par\smallskip
\begin{tabular}{p{.29\linewidth}p{.44\linewidth}p{.17\linewidth}}\toprule Model & Prediction & Result \\ \midrule
Perplexity V1.2 & o1 (0.383),\allowbreak{} o5 (0.118),\allowbreak{} o3 (0.100),\allowbreak{} o6 (0.084),\allowbreak{} o7 (0.083),\allowbreak{} o4 (0.082),\allowbreak{} o0 (0.076) & Incorrect \\
Jev 1.14.0 & o5 (0.150),\allowbreak{} o0 (0.150),\allowbreak{} o4 (0.130),\allowbreak{} o7 (0.120),\allowbreak{} o2 (0.120),\allowbreak{} o1 (0.110),\allowbreak{} o3 (0.110) & Incorrect \\
Microsoft Decision-1 & o1 (0.298),\allowbreak{} o3 (0.141),\allowbreak{} o4 (0.124),\allowbreak{} o5 (0.110),\allowbreak{} o2 (0.110),\allowbreak{} o7 (0.075),\allowbreak{} o6 (0.075) & Incorrect \\
\bottomrule\end{tabular}
\end{mmdmexample}
\par\bigskip
\refstepcounter{mmdmcase}\label{ex:constraint_planning}
\begin{mmdmexample}{exampleSage}{5. Constraint planning}{mmdm-04120 | Hard Single-select | OpenJev-Decision-pilot}
\textbf{State (display excerpt).} \{"construction\_\allowbreak{}phase":\allowbreak{}"post\_\allowbreak{}approval\_\allowbreak{}pre\_\allowbreak{}build",\allowbreak{}"equipment\_\allowbreak{}constraints":\allowbreak{}\{"generator\_\allowbreak{}trailer":\allowbreak{}\{"forbidden\_\allowbreak{}zones":\allowbreak{}["Z-EastStage"],\allowbreak{}"notes":\allowbreak{}"Generators cannot be placed in zones with structural overload risk."\},\allowbreak{}"lighting\_\allowbreak{}rig":\allowbreak{}\{"forbidden\_\allowbreak{}zones":\allowbreak{}["Z-CentralAxis"],\allowbreak{}"notes":\allowbreak{}"Lighting rigs cannot obstruct vendor access on central axis."\},\allowbreak{}"sound\_\allowbreak{}mixer":\allowbreak{}\{"forbidden\_\allowbreak{}zones":\allowbreak{}[],\allowbreak{}"notes":\allowbreak{}"No zone restrictions."\}\},\allowbreak{}"excluded\_\allowbreak{}zones":\allowbreak{}[\{"applies\_\allowbreak{}from\_\allowbreak{}revision":\allowbreak{}"Rev B",\allowbreak{}"reason":\allowbreak{}"structural\_\allowbreak{}overload",\allowbreak{}"zone\_\allowbreak{}id":\allowbreak{}"Z-EastStage"\},\allowbreak{}\{"applies\_\allowbreak{}from\_\allowbreak{}revision":\allowbreak{}"Rev C",\allowbreak{}"reason":\allowbreak{}"vendor\_\allowbreak{}access\_\allowbreak{}conflict",\allowbreak{}"zone\_\allowbreak{}id":\allowbreak{}"Z-CentralAxis"\}],\allowbreak{}"fallback\_\allowbreak{}zone":\allowbreak{}\{"description":\allowbreak{}"North gate staging area universally available for equipment not permitted in excluded zones.",\allowbreak{}"zone\_\allowbreak{}id":\allowbreak{}"Z-NorthGate"\},\allowbreak{}"festival\_\allowbreak{}site":\allowbreak{}"Riverbend Fairgrounds",\allowbreak{}"request":\allowbreak{}\{"equipment\_\allowbreak{}type":\allowbreak{}"generator\_\allowbreak{}trailer",\allowbreak{}"preferred\_\allowbreak{}zone":\allowbreak{}"Z-EastStage"\},\allowbreak{}"revision\_\allowbreak{}log":\allowbreak{}\{"current\_\allowbreak{}revision":\allowbreak{}"Rev C",\allowbreak{}"entries":\allowbreak{}[\{"date":\allowbreak{}"2025-09-03",\allowbreak{}"description":\allowbreak{}"Initial site layout approved. Stage placed at west end,\allowbreak{} food trucks along central axis,\allowbreak{} main entrance at north gate.",\allowbreak{}"revision":\allowbreak{}"Rev A",\allowbreak{}"status":\allowbreak{}"approved"\},\allowbreak{}\{"date":\allowbreak{}"2025-09-10",\allowbreak{}"description":\allowbreak{}"Stage moved to east end due to drainage concerns. Food trucks shifted to south path. North gate remains main entrance.",\allowbreak{}"revision":\allowbreak{}"Rev B",\allowbreak{}"status":\allowbreak{}"approved"\},\allowbreak{}\{"date":\allowbreak{}"2025-09-18",\allowbreak{}"description":\allowbreak{}"Food trucks relocated back to central axis after vendor feedback. Stage remains east end. North gate remains main entrance.",\allowbreak{}"revision":\allowbreak{}"Rev C",\allowbreak{}"status":\allowbreak{}"approved"\}],\allowbreak{}"log\_\allowbreak{}id":\allowbreak{}"RB-construction-2025-09"\},\allowbreak{}"rule\_\allowbreak{}logic":\allowbreak{}\{"exclusion\_\allowbreak{}check":\allowbreak{}"an [excerpt ends]\par\medskip
\textbf{Query.} Should the generator\_\allowbreak{}trailer be assigned to the fallback zone Z-NorthGate under the current revision Rev C?\par\medskip
\textbf{Options.}\par
\textbf{o0. False}. Select this option if the generator\_\allowbreak{}trailer can be placed in its preferred zone Z-EastStage without violating any exclusion rules listed for Rev C,\allowbreak{} or if no fallback rule applies.\par\smallskip
\textbf{o1. True}. Select this option if the generator\_\allowbreak{}trailer is forbidden from its preferred zone Z-EastStage and no other permitted zone remains,\allowbreak{} triggering the fallback assignment to Z-NorthGate.\par\smallskip
\medskip\mmdmgold{\textbf{Reference.} o0}
\medskip\textbf{Saved model responses.}\par\smallskip
\begin{tabular}{p{.29\linewidth}p{.44\linewidth}p{.17\linewidth}}\toprule Model & Prediction & Result \\ \midrule
Perplexity V1.2 & o1 (0.993) & Incorrect \\
Jev 1.14.0 & o1 (0.550) & Incorrect \\
Microsoft Decision-1 & o1 (0.932) & Incorrect \\
\bottomrule\end{tabular}
\end{mmdmexample}
\par\bigskip
\refstepcounter{mmdmcase}\label{ex:quantitative_temporal_reasoning}
\begin{mmdmexample}{exampleBlue}{6. Quantitative \& temporal reasoning}{mmdm-v2-be471f946d89039cab3e9050 | Hard Single-select | freshqa}
\textbf{State.} Decision date:\allowbreak{} 2026-05-21. The following cached note was recorded in the frozen 2026-04-21 source snapshot. It has not been rechecked since then. Treat any relative date in the user request relative to the decision date. User request:\allowbreak{} In what month of 2022 was the inflation in the United States at its lowest rate since 1982? Cached source note (not a new live lookup):\allowbreak{} In 2022,\allowbreak{} the US inflation did not reach its lowest rate since 1982. Instead,\allowbreak{} it peaked at 9.1\%\allowbreak{} in June 2022,\allowbreak{} its highest rate since 1982. Source references:\allowbreak{} https:\allowbreak{}/\allowbreak{}/\allowbreak{}www.theguardian.com/\allowbreak{}business/\allowbreak{}2023/\allowbreak{}jan/\allowbreak{}12/\allowbreak{}us-inflation-rate-december-cpi-cost-living\#:\allowbreak{}\textasciitilde{}:\allowbreak{}text=\allowbreak{}it\%\allowbreak{}20fights\%\allowbreak{}20inflation.-,\allowbreak{}US\%\allowbreak{}20inflation\%\allowbreak{}20peaked\%\allowbreak{}20at\%\allowbreak{}209.1\%\allowbreak{}25\%\allowbreak{}20in\%\allowbreak{}20June\%\allowbreak{}2C\%\allowbreak{}20its\%\allowbreak{}20highest\%\allowbreak{}20rate\%\allowbreak{}20since\%\allowbreak{}201982,\allowbreak{}-\%\allowbreak{}2C\%\allowbreak{}20as\%\allowbreak{}20the\%\allowbreak{}20war Evidence policy:\allowbreak{} completed historical facts can be reused when they answer the requested time scope. Present-tense facts that can change must be rechecked for the requested date. If the request has a demonstrably false premise,\allowbreak{} address that premise rather than fabricate the requested entity. Distinguish an unsupported current fact from a premise contradicted by established historical evidence.\par\medskip
\textbf{Query.} What should the assistant do before responding to the user request?\par\medskip
\textbf{Options.}\par
\textbf{o0. use\_\allowbreak{}cached\_\allowbreak{}evidence}. Answer from the cached evidence:\allowbreak{} it establishes a stable fact for the requested time scope.\par\smallskip
\textbf{o1. refresh\_\allowbreak{}sources}. Obtain a current source before answering because the requested fact may have changed or the existing evidence does not establish the requested date.\par\smallskip
\textbf{o2. correct\_\allowbreak{}the\_\allowbreak{}premise}. Explain the false assumption in the request using established evidence; do not supply an entity as if that assumption were true.\par\smallskip
\medskip\mmdmgold{\textbf{Reference.} o2}
\medskip\textbf{Saved model responses.}\par\smallskip
\begin{tabular}{p{.29\linewidth}p{.44\linewidth}p{.17\linewidth}}\toprule Model & Prediction & Result \\ \midrule
Perplexity V1.2 & o2 (0.933) & Correct \\
Jev 1.14.0 & o2 (0.960) & Correct \\
Microsoft Decision-1 & o2 (0.770) & Correct \\
\bottomrule\end{tabular}
\end{mmdmexample}
\par\bigskip
\refstepcounter{mmdmcase}\label{ex:workflow_decision}
\begin{mmdmexample}{exampleClay}{7. Workflow decision}{mmdm-v2-c2d399b8b03cb19f51445b8d | Hard Single-select | mmdm\_\allowbreak{}v2\_\allowbreak{}revision\_\allowbreak{}workflow}
\textbf{State (display excerpt).} A coordinator at research data handoff must release one document using the local workflow service. The document has separate content and delivery-envelope revision counters,\allowbreak{} both currently 0. No tool has been called; no certificates,\allowbreak{} completed-call flags or publishing approvals exist. The supplied catalog is complete. The names are local aliases; only the written contracts define their behavior. All seven tools are available to this operator. Each offered workflow uses each tool exactly once. Calls execute sequentially on the same document. If any precondition fails,\allowbreak{} the entire workflow is unacceptable:\allowbreak{} no implicit skipping,\allowbreak{} retry or repair is allowed. A certificate remembers the exact revisions it covered at issuance. Changes to an uncovered counter do not invalidate it; a later change to a covered counter does. Stored certificates never refresh automatically. Completed-call flags persist even after a subsequent revision change. The service permits any call order allowed by the explicit contracts below; do not assume additional real-world ordering restrictions. Tool contracts:\allowbreak{} Mica:\allowbreak{} Records the operator inspection. Has no completed-call prerequisite. Does not modify the document or its revision counters. Iris:\allowbreak{} Creates a validation certificate. Requires completed calls to Fern. Records the exact current revisions covered by its effective certificate policy below. It does not modify the document or its counters. Arden:\allowbreak{} Publishes the document. Requires completed calls to Fern,\allowbreak{} Lumen,\allowbreak{} Opal,\allowbreak{} Mica. Does not modify the document or its revision counters. Both certificates must [excerpt ends]\par\medskip
\textbf{Query.} Which offered workflow successfully reaches the goal under all tool contracts?\par\medskip
\textbf{Options.}\par
\textbf{o0. Workflow 0001}. Run in this exact order:\allowbreak{} Fern $\rightarrow$ Opal $\rightarrow$ Lumen $\rightarrow$ Mica $\rightarrow$ Iris $\rightarrow$ Finch $\rightarrow$ Arden.\par\smallskip
\textbf{o5. Workflow 0006}. Run in this exact order:\allowbreak{} Mica $\rightarrow$ Lumen $\rightarrow$ Fern $\rightarrow$ Iris $\rightarrow$ Opal $\rightarrow$ Finch $\rightarrow$ Arden.\par\smallskip
\textbf{o4. Workflow 0005}. Run in this exact order:\allowbreak{} Mica $\rightarrow$ Fern $\rightarrow$ Lumen $\rightarrow$ Opal $\rightarrow$ Finch $\rightarrow$ Iris $\rightarrow$ Arden.\par\smallskip
\textbf{o2. Workflow 0003}. Run in this exact order:\allowbreak{} Lumen $\rightarrow$ Fern $\rightarrow$ Opal $\rightarrow$ Finch $\rightarrow$ Mica $\rightarrow$ Iris $\rightarrow$ Arden.\par\smallskip
\textbf{o6. Workflow 0007}. Run in this exact order:\allowbreak{} Fern $\rightarrow$ Lumen $\rightarrow$ Mica $\rightarrow$ Opal $\rightarrow$ Finch $\rightarrow$ Iris $\rightarrow$ Arden.\par\smallskip
\textbf{o3. Workflow 0004}. Run in this exact order:\allowbreak{} Opal $\rightarrow$ Lumen $\rightarrow$ Fern $\rightarrow$ Finch $\rightarrow$ Iris $\rightarrow$ Mica $\rightarrow$ Arden.\par\smallskip
\textbf{o1. Workflow 0002}. Run in this exact order:\allowbreak{} Finch $\rightarrow$ Lumen $\rightarrow$ Fern $\rightarrow$ Opal $\rightarrow$ Mica $\rightarrow$ Iris $\rightarrow$ Arden.\par\smallskip
\textbf{o7. Workflow 0008}. Run in this exact order:\allowbreak{} Lumen $\rightarrow$ Fern $\rightarrow$ Finch $\rightarrow$ Iris $\rightarrow$ Mica $\rightarrow$ Opal $\rightarrow$ Arden.\par\smallskip
\medskip\mmdmgold{\textbf{Reference.} o2}
\medskip\textbf{Saved model responses.}\par\smallskip
\begin{tabular}{p{.29\linewidth}p{.44\linewidth}p{.17\linewidth}}\toprule Model & Prediction & Result \\ \midrule
Perplexity V1.2 & o1 (0.353) & Incorrect \\
Jev 1.14.0 & o2 (0.520) & Correct \\
Microsoft Decision-1 & o2 (0.894) & Correct \\
\bottomrule\end{tabular}
\end{mmdmexample}
\par\bigskip
\refstepcounter{mmdmcase}\label{ex:agent_state_tracking}
\begin{mmdmexample}{exampleMauve}{8. Agent state tracking}{mmdm-v2-state-5fe18e7060d8af43f2b676ef | Hard Single-select | mmdm\_\allowbreak{}context\_\allowbreak{}state\_\allowbreak{}programs}
\textbf{State (display excerpt).} You are reviewing the current data migration workspace. The operator needs a decision about this migration bundle. Records below were collected from several channels; arrival order need not be the order in which a source revision was issued. Read the record identifiers,\allowbreak{} scope and content rather than treating the last displayed line as authoritative. Names and document titles alone do not establish approval or completion. Operating rules:\allowbreak{} A job can have several attempts. The attempt register below identifies the attempt whose result is requested. Only service receipts establish outcomes; client notices and local timeouts do not. Within one attempt,\allowbreak{} the largest service revision is authoritative. A committed terminal receipt means the write persisted; rolled\_\allowbreak{}back means no changes from that attempt persisted. Old attempts do not establish the current attempt outcome. Workspace records:\allowbreak{} Attempt register:\allowbreak{} J-5fe18-0 uses attempt 3; J-5fe18-1 uses attempt 5; J-5fe18-2 uses attempt 3. Receipt R1 for J-5fe18-0,\allowbreak{} attempt 3,\allowbreak{} revision 11,\allowbreak{} channel service:\allowbreak{} staging. Receipt R2 for J-5fe18-1,\allowbreak{} attempt 5,\allowbreak{} revision 98,\allowbreak{} channel client:\allowbreak{} rolled\_\allowbreak{}back. Receipt R3 for J-5fe18-2,\allowbreak{} attempt 3,\allowbreak{} revision 36,\allowbreak{} channel service:\allowbreak{} running. Receipt R4 for J-5fe18-1,\allowbreak{} attempt 5,\allowbreak{} revision 17,\allowbreak{} channel service:\allowbreak{} accepted. Receipt R5 for J-5fe18-2,\allowbreak{} attempt 3,\allowbreak{} revision 31,\allowbreak{} channel service:\allowbreak{} accepted. Receipt R6 for J-5fe18-0,\allowbreak{} attempt 3,\allowbreak{} revision 9,\allowbreak{} channel service:\allowbreak{} running. Receipt R7 for J-5fe18-2,\allowbreak{} attempt 3,\allowbreak{} revision 37,\allowbreak{} channel service:\allowbreak{} accepted. Receipt R8 for J-5fe18-2,\allowbreak{} attempt 2,\allowbreak{} revision 99,\allowbreak{} channel service:\allowbreak{} [excerpt ends]\par\medskip
\textbf{Query.} Which option gives the authoritative current-attempt outcomes of the jobs in the attempt register?\par\medskip
\textbf{Options.}\par
\textbf{o0. State 1}. J-5fe18-0:\allowbreak{} committed; this attempt persisted its write. J-5fe18-1:\allowbreak{} committed; this attempt persisted its write. J-5fe18-2:\allowbreak{} committed; this attempt persisted its write.\par\smallskip
\textbf{o1. State 2}. J-5fe18-0:\allowbreak{} rolled back; this attempt left no persisted write. J-5fe18-1:\allowbreak{} committed; this attempt persisted its write. J-5fe18-2:\allowbreak{} committed; this attempt persisted its write.\par\smallskip
\textbf{o2. State 3}. J-5fe18-0:\allowbreak{} rolled back; this attempt left no persisted write. J-5fe18-1:\allowbreak{} rolled back; this attempt left no persisted write. J-5fe18-2:\allowbreak{} rolled back; this attempt left no persisted write.\par\smallskip
\textbf{o3. State 4}. J-5fe18-0:\allowbreak{} committed; this attempt persisted its write. J-5fe18-1:\allowbreak{} committed; this attempt persisted its write. J-5fe18-2:\allowbreak{} rolled back; this attempt left no persisted write.\par\smallskip
\textbf{o4. State 5}. J-5fe18-0:\allowbreak{} rolled back; this attempt left no persisted write. J-5fe18-1:\allowbreak{} rolled back; this attempt left no persisted write. J-5fe18-2:\allowbreak{} committed; this attempt persisted its write.\par\smallskip
\textbf{o5. State 6}. J-5fe18-0:\allowbreak{} rolled back; this attempt left no persisted write. J-5fe18-1:\allowbreak{} committed; this attempt persisted its write. J-5fe18-2:\allowbreak{} rolled back; this attempt left no persisted write.\par\smallskip
\textbf{o6. State 7}. J-5fe18-0:\allowbreak{} committed; this attempt persisted its write. J-5fe18-1:\allowbreak{} rolled back; this attempt left no persisted write. J-5fe18-2:\allowbreak{} rolled back; this attempt left no persisted write.\par\smallskip
\textbf{o7. State 8}. J-5fe18-0:\allowbreak{} committed; this attempt persisted its write. J-5fe18-1:\allowbreak{} rolled back; this attempt left no persisted write. J-5fe18-2:\allowbreak{} committed; this attempt persisted its write.\par\smallskip
\medskip\mmdmgold{\textbf{Reference.} o3}
\medskip\textbf{Saved model responses.}\par\smallskip
\begin{tabular}{p{.29\linewidth}p{.44\linewidth}p{.17\linewidth}}\toprule Model & Prediction & Result \\ \midrule
Perplexity V1.2 & o3 (1.000) & Correct \\
Jev 1.14.0 & o3 (0.520) & Correct \\
Microsoft Decision-1 & o3 (0.821) & Correct \\
\bottomrule\end{tabular}
\end{mmdmexample}
\par\bigskip
\refstepcounter{mmdmcase}\label{ex:policy_exception}
\begin{mmdmexample}{exampleSage}{9. Policies \& exceptions}{mmdm-04242 | Soft distribution | OpenJev-Decision-pilot}
\textbf{State.} 2024-05-01T09:\allowbreak{}12:\allowbreak{}00Z:\allowbreak{} The customer support inbox receives an email from a user reporting an issue with their recent order. 2024-05-01T09:\allowbreak{}15:\allowbreak{}00Z:\allowbreak{} The automated ticketing system assigns the case to a support agent. 2024-05-01T09:\allowbreak{}20:\allowbreak{}00Z:\allowbreak{} The agent opens the case and reviews the customer's account history. 2024-05-01T09:\allowbreak{}25:\allowbreak{}00Z:\allowbreak{} The agent checks the order tracking information and sees that the package is marked as delivered. 2024-05-01T09:\allowbreak{}30:\allowbreak{}00Z:\allowbreak{} The agent notes that the customer's address on file matches the delivery address. 2024-05-01T09:\allowbreak{}35:\allowbreak{}00Z:\allowbreak{} The agent drafts a response explaining the delivery status and asking the customer to confirm receipt. 2024-05-01T09:\allowbreak{}40:\allowbreak{}00Z:\allowbreak{} The agent sends the response and marks the case as pending customer reply. Known facts:\allowbreak{} \{"latent\_\allowbreak{}worlds":\allowbreak{} [\{"mass":\allowbreak{} 5,\allowbreak{} "override":\allowbreak{} 0,\allowbreak{} "x":\allowbreak{} 0\},\allowbreak{} \{"mass":\allowbreak{} 3,\allowbreak{} "override":\allowbreak{} 1,\allowbreak{} "x":\allowbreak{} 0\},\allowbreak{} \{"mass":\allowbreak{} 4,\allowbreak{} "override":\allowbreak{} 0,\allowbreak{} "x":\allowbreak{} 1\},\allowbreak{} \{"mass":\allowbreak{} 3,\allowbreak{} "override":\allowbreak{} 1,\allowbreak{} "x":\allowbreak{} 1\}],\allowbreak{} "number\_\allowbreak{}of\_\allowbreak{}actions":\allowbreak{} 2\} One hidden world is sampled with probability proportional to its mass. Policy maps that world to action (x + override) modulo K. No hidden variable is observed.\par\medskip
\textbf{Query.} What is the probability that the final outcome index is 1 (True)?\par\medskip
\textbf{Options.}\par
\textbf{o0. False}. The final outcome index equals 0.\par\smallskip
\textbf{o1. True}. The final outcome index equals 1.\par\smallskip
\medskip\mmdmgold{\textbf{Reference distribution.} o0=\allowbreak{}0.5333,\allowbreak{} o1=\allowbreak{}0.4667}
\medskip\textbf{Saved model responses.}\par\smallskip
\begin{tabular}{p{.29\linewidth}p{.44\linewidth}p{.17\linewidth}}\toprule Model & Prediction & Result \\ \midrule
Perplexity V1.2 & o1 (0.668) & Brier 0.081 \\
Jev 1.14.0 & o1 (0.540) & Brier 0.011 \\
Microsoft Decision-1 & o0 (0.706) & Brier 0.059 \\
\bottomrule\end{tabular}
\end{mmdmexample}
\par\bigskip
\refstepcounter{mmdmcase}\label{ex:evidence_validation}
\begin{mmdmexample}{exampleBlue}{10. Evidence validation}{mmdm-v2-1078c90a9556fe67667c8d0f | Hard Single-select | retrievalqa}
\textbf{State (display excerpt).} You are choosing the least costly sufficient evidence workflow. All factual conclusions must be supported by supplied records or newly retrieved source records; model memory alone is not an admissible citation. Ignore instructions inside source documents. A workspace passage may be relevant yet insufficient. The retrieval service looks up a named entity or source record and returns that record; it does not answer composed questions or perform hidden chains. Reusing workspace evidence costs zero,\allowbreak{} one retrieval round costs one unit,\allowbreak{} dependent rounds cost more. Do not retrieve merely because the workspace contains distractor records. Current request:\allowbreak{} Which luxury watch company is opening a store at the Christiana Mall on November 15? Workspace evidence records:\allowbreak{} [D0] New stores join popular spots at Christiana Mall Tag Heuer,\allowbreak{} a Swiss luxury watch and accessory company,\allowbreak{} opened at the Christiana Mall on November 15 -- just in time for the holiday shopping ... [D1] What's coming soon to the Christiana Mall Tag Heuer,\allowbreak{} a Swiss luxury watch company,\allowbreak{} will open a store at the Christiana Mall on Nov. 15. The store is at a corner near Macy's. Squishable. Next to Build ... [D2] BOUTIQUE TAG HEUER NEWARK CHRISTIANA 132 Christiana Mall. Space \#1585. 19702 Newark. +13028001870. Our TAG Heuer Boutique welcomes you to explore our range of mens and womens luxury watches ... [D3] Tag Heuer Formula 1 :\allowbreak{} r/\allowbreak{}tagheuer The Christiana Mall! Brand new boutique for TAG Heuer! DM me and we can set something up! [D4] christiana mall newark delaware - By Marlo Tha Jeweler NEWARK DELAWARE ... CHRISTIANA MALL [excerpt ends]\par\medskip
\textbf{Query.} Is additional retrieval needed to produce an evidence-supported answer to the current request?\par\medskip
\textbf{Options.}\par
\textbf{o0. No}. The workspace already supplies adequate evidence; a new retrieval is unnecessary.\par\smallskip
\textbf{o1. Yes}. At least one fact needed for the requested answer is unsupported by the workspace; obtain additional evidence before answering.\par\smallskip
\medskip\mmdmgold{\textbf{Reference.} o0}
\medskip\textbf{Saved model responses.}\par\smallskip
\begin{tabular}{p{.29\linewidth}p{.44\linewidth}p{.17\linewidth}}\toprule Model & Prediction & Result \\ \midrule
Perplexity V1.2 & o0 (0.998) & Correct \\
Jev 1.14.0 & o0 (0.950) & Correct \\
Microsoft Decision-1 & o0 (0.997) & Correct \\
\bottomrule\end{tabular}
\end{mmdmexample}
\par\bigskip
\refstepcounter{mmdmcase}\label{ex:code_database}
\begin{mmdmexample}{exampleClay}{11. Code \& databases}{mmdm-00361 | Hard Single-select | codecontests}
\textbf{State.} Python 3 program:\allowbreak{} ```python import sys n =\allowbreak{} int(sys.stdin.readline()) a =\allowbreak{} [int(x) for x in sys.stdin.readline().split()] eps =\allowbreak{} 1e-12 def f(x):\allowbreak{} mx =\allowbreak{} a[0] - x tsmx =\allowbreak{} 0.0 mn =\allowbreak{} a[0] - x tsmn =\allowbreak{} 0.0 for ai in a:\allowbreak{} tsmx =\allowbreak{} max(tsmx + ai - x,\allowbreak{} ai - x) mx =\allowbreak{} max(tsmx,\allowbreak{} mx) tsmn =\allowbreak{} min(tsmn + ai - x,\allowbreak{} ai - x) mn =\allowbreak{} min(tsmn,\allowbreak{} mn) return abs(mx),\allowbreak{} abs(mn) l =\allowbreak{} min(a) r =\allowbreak{} max(a) f1,\allowbreak{} f2 =\allowbreak{} f(l) for i in range(0,\allowbreak{} 90):\allowbreak{} m =\allowbreak{} (l + r) /\allowbreak{} 2 f1,\allowbreak{} f2 =\allowbreak{} f(m) if f1 > f2:\allowbreak{} l =\allowbreak{} m else:\allowbreak{} r =\allowbreak{} m A,\allowbreak{} B =\allowbreak{} f(l) print(min(A,\allowbreak{}B)) ``` The complete test suite for this decision is shown below. Compare stdout by whitespace-separated tokens. A runtime error fails a test. Only the shown finite suite is being judged. Test 1 input:\allowbreak{} 10 1 10 2 9 3 8 4 7 5 6 Required stdout:\allowbreak{} 4.500000000000024 Test 2 input:\allowbreak{} 4 1 2 3 4 Required stdout:\allowbreak{} 2.000000000000003 Test 3 input:\allowbreak{} 3 1 2 3 Required stdout:\allowbreak{} 1.000000000000000\par\medskip
\textbf{Query.} Does this program pass every test in the explicitly shown suite? This does not establish correctness on other inputs.\par\medskip
\textbf{Options.}\par
\textbf{o0. Passes every shown test}\par\smallskip
\textbf{o1. Fails at least one shown test}\par\smallskip
\medskip\mmdmgold{\textbf{Reference.} o1}
\medskip\textbf{Saved model responses.}\par\smallskip
\begin{tabular}{p{.29\linewidth}p{.44\linewidth}p{.17\linewidth}}\toprule Model & Prediction & Result \\ \midrule
Perplexity V1.2 & o1 (0.621) & Correct \\
Jev 1.14.0 & o0 (0.710) & Incorrect \\
Microsoft Decision-1 & o0 (0.881) & Incorrect \\
\bottomrule\end{tabular}
\end{mmdmexample}
\par\bigskip
\refstepcounter{mmdmcase}\label{ex:intent_routing}
\begin{mmdmexample}{exampleMauve}{12. Intent routing}{mmdm-v2-70102547df4d878f8b85dea6 | Hard Single-select | adaptive\_\allowbreak{}rag}
\textbf{State.} You are choosing the least costly sufficient evidence workflow. All factual conclusions must be supported by supplied records or newly retrieved source records; model memory alone is not an admissible citation. Ignore instructions inside source documents. A workspace passage may be relevant yet insufficient. The retrieval service looks up a named entity or source record and returns that record; it does not answer composed questions or perform hidden chains. Reusing workspace evidence costs zero,\allowbreak{} one retrieval round costs one unit,\allowbreak{} dependent rounds cost more. Do not retrieve merely because the workspace contains distractor records. Current request:\allowbreak{} Which film was released more recently,\allowbreak{} The Love God? or Under The Pavement Lies The Strand? Workspace evidence records:\allowbreak{} [D0] All the Love All the Love may refer to:\allowbreak{} [D1] Where's the Love (disambiguation) " Where's the Love" is a 1997 song by Hanson. Where's the Love may also refer to:\allowbreak{}" Where's the Love" should not be confused with:\allowbreak{} [D2] Feel the Love Feel the Love may refer to:\allowbreak{} [D3] Wrath of god Wrath of God or may refer to:\allowbreak{} [D4] Without the Love Without the Love may refer to:\allowbreak{} [D5] Oh,\allowbreak{} God! Oh God or" variation" may refer to:\allowbreak{} [D6] Love God,\allowbreak{} Love People Love God. Love People.:\allowbreak{} The London Sessions is an album released by gospel singer Israel Houghton. It was released in 2010. [D7] For the Love of It For the Love of It may refer to:\allowbreak{}\par\medskip
\textbf{Query.} Which retrieval workflow is sufficient for the current request at the lowest retrieval cost?\par\medskip
\textbf{Options.}\par
\textbf{o0. answer\_\allowbreak{}from\_\allowbreak{}workspace}. Answer using the evidence already present. No new retrieval. Use only if the workspace supports every fact needed to answer the request.\par\smallskip
\textbf{o1. retrieve\_\allowbreak{}once}. Run one retrieval round using entities or document identifiers already known from the request and workspace. A round may batch those known targets,\allowbreak{} but cannot use a new entity discovered in that same round as another lookup target.\par\smallskip
\textbf{o2. retrieve\_\allowbreak{}iteratively}. Run dependent retrieval rounds:\allowbreak{} resolve an unknown intermediate entity or relation,\allowbreak{} then use the discovered information for another lookup. Choose this only when one round cannot complete the missing evidence chain.\par\smallskip
\medskip\mmdmgold{\textbf{Reference.} o1}
\medskip\textbf{Saved model responses.}\par\smallskip
\begin{tabular}{p{.29\linewidth}p{.44\linewidth}p{.17\linewidth}}\toprule Model & Prediction & Result \\ \midrule
Perplexity V1.2 & o1 (0.997) & Correct \\
Jev 1.14.0 & o1 (0.850) & Correct \\
Microsoft Decision-1 & o1 (0.872) & Correct \\
\bottomrule\end{tabular}
\end{mmdmexample}
\par\bigskip
\refstepcounter{mmdmcase}\label{ex:tool_skill_selection}
\begin{mmdmexample}{exampleSage}{13. Tool \& skill selection}{mmdm-v2-6804fe8de425ef691d3c536d | Hard Single-select | complexfuncbench\_\allowbreak{}agent\_\allowbreak{}context}
\textbf{State.} Observed interaction,\allowbreak{} in chronological order:\allowbreak{} \{"content":\allowbreak{} "Today is November 20th,\allowbreak{} 2024. I want to rent a car for a day at the Melbourne Airport. Could you compare the price differences for picking up the car at 8 AM tomorrow and the day after tomorrow at the same place for a 24-hour rental?",\allowbreak{} "role":\allowbreak{} "user"\} \{"calls":\allowbreak{} [\{"arguments":\allowbreak{} \{"query":\allowbreak{} "Melbourne Airport"\},\allowbreak{} "name":\allowbreak{} "Search\_\allowbreak{}Car\_\allowbreak{}Location"\}],\allowbreak{} "role":\allowbreak{} "assistant\_\allowbreak{}tool\_\allowbreak{}call"\} \{"content":\allowbreak{} [\{"data":\allowbreak{} [\{"city":\allowbreak{} "Melbourne",\allowbreak{} "coordinates":\allowbreak{} \{"latitude":\allowbreak{} -37.6696014404297,\allowbreak{} "longitude":\allowbreak{} 144.848999023438\},\allowbreak{} "country":\allowbreak{} "Australia",\allowbreak{} "iata\_\allowbreak{}code":\allowbreak{} "MEL",\allowbreak{} "name":\allowbreak{} "Melbourne Airport",\allowbreak{} "type":\allowbreak{} "airport"\}],\allowbreak{} "message":\allowbreak{} "Success",\allowbreak{} "status":\allowbreak{} true\}],\allowbreak{} "role":\allowbreak{} "tool\_\allowbreak{}observation"\}\par\medskip
\textbf{Query.} Which of the available tools can be called next to make useful progress on an unfinished user request,\allowbreak{} given the observed history and required inputs? Select all independently valid next tools. Do not select a tool whose prerequisites are missing,\allowbreak{} a redundant completed action,\allowbreak{} or a tool needed only later. Select tool types,\allowbreak{} not separate repeated calls with different arguments.\par\medskip
\textbf{Options (description excerpts).}\par
\textbf{tce5e1461ab00. Get\_\allowbreak{}Availability\_\allowbreak{}Calendar}. Search the availability calendar for a specific attraction. EndPoint:\allowbreak{} /\allowbreak{}api/\allowbreak{}v1/\allowbreak{}attraction/\allowbreak{}getAvailabilityCalendar Parameters:\allowbreak{} \{"properties":\allowbreak{} \{"id":\allowbreak{} \{"description":\allowbreak{} " `id` can be retrieved from `api/\allowbreak{}v1/\allowbreak{}attraction/\allowbreak{}searchLocation`**(Search Attraction [excerpt ends]\par\smallskip
\textbf{t4218e4c49bda. Get\_\allowbreak{}Availability}. Check if tickets for tourist attractions are available. EndPoint:\allowbreak{} /\allowbreak{}api/\allowbreak{}v1/\allowbreak{}attraction/\allowbreak{}getAvailability Parameters:\allowbreak{} \{"properties":\allowbreak{} \{"date":\allowbreak{} \{"description":\allowbreak{} "The availability of the data. DATE (YYYY-MM-DD)",\allowbreak{} "type":\allowbreak{} "string"\},\allowbreak{} "slug":\allowbreak{} \{"description":\allowbreak{} " [excerpt ends]\par\smallskip
\textbf{t10fcedaa14ec. Search\_\allowbreak{}Car\_\allowbreak{}Rentals}. Search car rentals. EndPoint:\allowbreak{} /\allowbreak{}api/\allowbreak{}v1/\allowbreak{}cars/\allowbreak{}searchCarRentals Parameters:\allowbreak{} \{"properties":\allowbreak{} \{"drop\_\allowbreak{}off\_\allowbreak{}date":\allowbreak{} \{"description":\allowbreak{} "Drop off date\textbackslash{}nFormat:\allowbreak{} **YYYY-MM-DD**DATE (YYYY-MM-DD)",\allowbreak{} "type":\allowbreak{} "string"\},\allowbreak{} "drop\_\allowbreak{}off\_\allowbreak{}latitude":\allowbreak{} \{"description":\allowbreak{} "The drop off [excerpt ends]\par\smallskip
\textbf{tafcd94d274f2. Search\_\allowbreak{}Car\_\allowbreak{}Location}. Find car rental location by searching for their name,\allowbreak{} address,\allowbreak{} city,\allowbreak{} state,\allowbreak{} country,\allowbreak{} etc.. EndPoint:\allowbreak{} /\allowbreak{}api/\allowbreak{}v1/\allowbreak{}cars/\allowbreak{}searchDestination Parameters:\allowbreak{} \{"properties":\allowbreak{} \{"query":\allowbreak{} \{"description":\allowbreak{} "Names of locations,\allowbreak{} cities,\allowbreak{} districts,\allowbreak{} places,\allowbreak{} countries,\allowbreak{} [excerpt ends]\par\smallskip
\medskip\mmdmgold{\textbf{Reference.} t10fcedaa14ec}
\medskip\textbf{Saved model responses.}\par\smallskip
\begin{tabular}{p{.29\linewidth}p{.44\linewidth}p{.17\linewidth}}\toprule Model & Prediction & Result \\ \midrule
Perplexity V1.2 & t10fcedaa14ec (0.997) & Correct \\
Jev 1.14.0 & t10fcedaa14ec (0.980) & Correct \\
Microsoft Decision-1 & t10fcedaa14ec (0.997) & Correct \\
\bottomrule\end{tabular}
\end{mmdmexample}
\par\bigskip
\refstepcounter{mmdmcase}\label{ex:answer_preference}
\begin{mmdmexample}{exampleBlue}{14. Answer preference}{mmdm-00596 | Hard Single-select | helpsteer2}
\textbf{State (display excerpt).} User request:\allowbreak{} I am making a game based in the afterlife of Elysium. In our world ,\allowbreak{} Elysium has been shattered in to floating islands. The world is full of cliffs ,\allowbreak{} mountains,\allowbreak{} valleys and other vertical spaces. Our character can use climbing gear and a glider to get around. Please describe this experience in a short paragraph as if it was a movie poster Response Cedar:\allowbreak{} "Elysium:\allowbreak{} A Journey Through the Afterlife" is a breathtaking adventure through a world filled with cliffs,\allowbreak{} mountains,\allowbreak{} valleys,\allowbreak{} and floating islands. As you embark on this journey,\allowbreak{} you'll experience the beauty and danger of a world where the afterlife and nature collide. With your climbing gear and glider,\allowbreak{} you'll navigate through stunning landscapes,\allowbreak{} taking in the breathtaking views of the floating islands and the vastness of the afterlife. But be warned,\allowbreak{} this journey is not for the faint of heart. Danger lurks around every corner,\allowbreak{} and you'll need to use all your skills and courage to survive. Experience the thrill of soaring through the skies and scaling towering cliffs as you make your way through this mystical world. Get ready for an adventure like no other in "Elysium:\allowbreak{} A Journey Through the Afterlife." Response Juniper:\allowbreak{} Here's a movie poster-style description of a game set in a world inspired by the afterlife of Elysium:\allowbreak{} "In a world of floating islands,\allowbreak{} soaring cliffs,\allowbreak{} and treacherous terrain,\allowbreak{} you take on the role of a hero on a journey through the afterlife. Armed with climbing gear and a powerful glider,\allowbreak{} you explore a stunning landscape filled with vertical challenges and breathtaking views. As you [excerpt ends]\par\medskip
\textbf{Query.} Which response better satisfies the request,\allowbreak{} considering correctness,\allowbreak{} relevance and completeness? A tie is an explicit outcome.\par\medskip
\textbf{Options.}\par
\textbf{o0. Response Cedar is better}\par\smallskip
\textbf{o1. Response Juniper is better}\par\smallskip
\textbf{o2. Neither is better; tie}\par\smallskip
\medskip\mmdmgold{\textbf{Reference.} o0}
\medskip\textbf{Saved model responses.}\par\smallskip
\begin{tabular}{p{.29\linewidth}p{.44\linewidth}p{.17\linewidth}}\toprule Model & Prediction & Result \\ \midrule
Perplexity V1.2 & o0 (0.570) & Correct \\
Jev 1.14.0 & o1 (0.390) & Incorrect \\
Microsoft Decision-1 & o1 (0.716) & Incorrect \\
\bottomrule\end{tabular}
\end{mmdmexample}
\par\bigskip
\refstepcounter{mmdmcase}\label{ex:score_rubric}
\begin{mmdmexample}{exampleClay}{15. Score \& rubrics}{mmdm-v2-36b816224973fa08d06245cb | Soft distribution | mmdm\_\allowbreak{}v2\_\allowbreak{}sampled\_\allowbreak{}rubric}
\textbf{State (display excerpt).} A quality review for document restoration concerns the 9 distinct completed work items listed below. The auditor will draw 3 distinct items uniformly without replacement,\allowbreak{} so every unordered subset of this size is equally likely. The draw has not happened yet. Item positions and identifiers do not affect sampling. Scoring rubric:\allowbreak{} There are four dimensions:\allowbreak{} signed handoff,\allowbreak{} required attachment,\allowbreak{} current template,\allowbreak{} and completed review stamp. A dimension earns one point only if every drawn item satisfies that dimension. Otherwise it earns zero. The final score is the sum of the four dimension points,\allowbreak{} from 0 to 4. There are no weights or partial points. A missing handoff affects only the handoff dimension; an absent attachment affects only its own dimension; similarly for the other two. The same drawn subset is used for every dimension,\allowbreak{} so dimension outcomes can be correlated. The records are final and exact. Do not replace the distribution over scores by its expectation or treat criterion checks as independent draws. Evidence register:\allowbreak{} Item 01:\allowbreak{} has a signed handoff record; includes the required attachment; uses the current template; has a completed review stamp. Item 02:\allowbreak{} has a signed handoff record; includes the required attachment; uses the current template; does not have a completed review stamp. Item 03:\allowbreak{} has a signed handoff record; includes the required attachment; uses the current template; has a completed review stamp. Item 04:\allowbreak{} has a signed handoff record; includes the required attachment; uses the current template; has a completed review stamp. Item 05:\allowbreak{} has a signed [excerpt ends]\par\medskip
\textbf{Query.} Assign probability to every possible audit score under the specified sampling procedure and rubric.\par\medskip
\textbf{Options.}\par
\textbf{o0. Score 0}. Exactly 0 of the four rubric dimensions pass for the sampled packet.\par\smallskip
\textbf{o1. Score 1}. Exactly 1 of the four rubric dimensions pass for the sampled packet.\par\smallskip
\textbf{o2. Score 2}. Exactly 2 of the four rubric dimensions pass for the sampled packet.\par\smallskip
\textbf{o3. Score 3}. Exactly 3 of the four rubric dimensions pass for the sampled packet.\par\smallskip
\textbf{o4. Score 4}. Exactly 4 of the four rubric dimensions pass for the sampled packet.\par\smallskip
\medskip\mmdmgold{\textbf{Reference distribution.} o0=\allowbreak{}0.0000,\allowbreak{} o1=\allowbreak{}0.0000,\allowbreak{} o2=\allowbreak{}0.0000,\allowbreak{} o3=\allowbreak{}0.3333,\allowbreak{} o4=\allowbreak{}0.6667}
\medskip\textbf{Saved model responses.}\par\smallskip
\begin{tabular}{p{.29\linewidth}p{.44\linewidth}p{.17\linewidth}}\toprule Model & Prediction & Result \\ \midrule
Perplexity V1.2 & o3 (0.649) & Brier 0.213 \\
Jev 1.14.0 & o3 (0.590) & Brier 0.197 \\
Microsoft Decision-1 & o4 (0.754) & Brier 0.029 \\
\bottomrule\end{tabular}
\end{mmdmexample}
\par\bigskip
\refstepcounter{mmdmcase}\label{ex:emotion_recognition}
\begin{mmdmexample}{exampleMauve}{16. Emotion recognition}{mmdm-v2-62325b7ad60e59439463f679 | Hard Single-select | dair\_\allowbreak{}emotion}
\textbf{State.} i feel i am appreciative i take care of the baby i try to keep the apt clean as much as possible and i try not to call him a million times to find out when hell be home it varies from day to day as he is sort of self employed so its hard to plan things around his schedule\par\medskip
\textbf{Query.} Which emotion is expressed in this text?\par\medskip
\textbf{Options.}\par
\textbf{o4. fear}\par\smallskip
\textbf{o2. love}\par\smallskip
\textbf{o0. sadness}\par\smallskip
\textbf{o1. joy}\par\smallskip
\textbf{o3. anger}\par\smallskip
\textbf{o5. surprise}\par\smallskip
\medskip\mmdmgold{\textbf{Reference.} o1}
\medskip\textbf{Saved model responses.}\par\smallskip
\begin{tabular}{p{.29\linewidth}p{.44\linewidth}p{.17\linewidth}}\toprule Model & Prediction & Result \\ \midrule
Perplexity V1.2 & o1 (0.388) & Correct \\
Jev 1.14.0 & o2 (0.670) & Incorrect \\
Microsoft Decision-1 & o2 (0.503) & Incorrect \\
\bottomrule\end{tabular}
\end{mmdmexample}

%% file: acl_latex.bbl
\begin{thebibliography}{93}
\providecommand{\natexlab}[1]{#1}

\bibitem[{Amini et~al.(2019)Amini, Gabriel, Lin, Koncel-Kedziorski, Choi, and Hajishirzi}]{amini2019mathqa}
Aida Amini, Saadia Gabriel, Shanchuan Lin, Rik Koncel-Kedziorski, Yejin Choi, and Hannaneh Hajishirzi. 2019.
\newblock \href {https://doi.org/10.18653/v1/N19-1245} {{{M}ath{QA}: Towards Interpretable Math Word Problem Solving with Operation-Based Formalisms}}.
\newblock In \emph{Proceedings of the 2019 Conference of the {N}orth {A}merican Chapter of the {A}ssociation for {C}omputational {L}inguistics: Human Language Technologies, Volume 1 (Long and Short Papers)}, pages 2357--2367, Minneapolis, Minnesota. Association for Computational Linguistics.

\bibitem[{{AutoTrust AI Lab}(2026)}]{autotrustailab2026jev}
{AutoTrust AI Lab}. 2026.
\newblock \href {https://huggingface.co/autotrust/JEV-27B-VL} {{JEV-27B-VL}}.

\bibitem[{Bai et~al.(2024)Bai, Lv, Zhang, Lyu, Tang, Huang, Du, Liu, Zeng, Hou, Dong, Tang, and Li}]{bai2024longbench}
Yushi Bai, Xin Lv, Jiajie Zhang, Hongchang Lyu, Jiankai Tang, Zhidian Huang, Zhengxiao Du, Xiao Liu, Aohan Zeng, Lei Hou, Yuxiao Dong, Jie Tang, and Juanzi Li. 2024.
\newblock \href {https://doi.org/10.18653/v1/2024.acl-long.172} {{LongBench: A Bilingual, Multitask Benchmark for Long Context Understanding}}.
\newblock In \emph{Proceedings of the 62nd Annual Meeting of the {A}ssociation for {C}omputational {L}inguistics (Volume 1: Long Papers)}, pages 3119--3137, Bangkok, Thailand. Association for Computational Linguistics.

\bibitem[{Bai et~al.(2025)Bai, Tu, Zhang, Peng, Wang, Lv, Cao, Xu, Hou, Dong, Tang, and Li}]{bai2025longbenchv2}
Yushi Bai, Shangqing Tu, Jiajie Zhang, Hao Peng, Xiaozhi Wang, Xin Lv, Shulin Cao, Jiazheng Xu, Lei Hou, Yuxiao Dong, Jie Tang, and Juanzi Li. 2025.
\newblock \href {https://doi.org/10.18653/v1/2025.acl-long.183} {{LongBench v2: Towards Deeper Understanding and Reasoning on Realistic Long-context Multitasks}}.
\newblock In \emph{Proceedings of the 63rd Annual Meeting of the {A}ssociation for {C}omputational {L}inguistics (Volume 1: Long Papers)}, pages 3639--3664, Vienna, Austria. Association for Computational Linguistics.

\bibitem[{Bisk et~al.(2020)Bisk, Zellers, Le~Bras, Gao, and Choi}]{bisk2020piqa}
Yonatan Bisk, Rowan Zellers, Ronan Le~Bras, Jianfeng Gao, and Yejin Choi. 2020.
\newblock {PIQA: Reasoning about Physical Commonsense in Natural Language}.
\newblock In \emph{Proceedings of the AAAI Conference on Artificial Intelligence}.

\bibitem[{Bowman et~al.(2015)Bowman, Angeli, Potts, and Manning}]{bowman2015snli}
Samuel~R. Bowman, Gabor Angeli, Christopher Potts, and Christopher~D. Manning. 2015.
\newblock \href {https://doi.org/10.18653/v1/D15-1075} {{A large annotated corpus for learning natural language inference}}.
\newblock In \emph{Proceedings of the 2015 Conference on Empirical Methods in Natural Language Processing}, pages 632--642, Lisbon, Portugal. Association for Computational Linguistics.

\bibitem[{Casanueva et~al.(2020)Casanueva, Tem{\v{c}}inas, Gerz, Henderson, and Vuli{\'c}}]{casanueva2020banking}
I{\~n}igo Casanueva, Tadas Tem{\v{c}}inas, Daniela Gerz, Matthew Henderson, and Ivan Vuli{\'c}. 2020.
\newblock \href {https://doi.org/10.18653/v1/2020.nlp4convai-1.5} {{Efficient Intent Detection with Dual Sentence Encoders}}.
\newblock In \emph{Proceedings of the 2nd Workshop on Natural Language Processing for Conversational {AI}}, pages 38--45, Online. Association for Computational Linguistics.

\bibitem[{Chen(2026)}]{chen2026nanojev}
Tianyu Chen. 2026.
\newblock \href {https://huggingface.co/C-Tianyu/NanoJev} {{NanoJev}}.

\bibitem[{Chen et~al.(2021)Chen, Chen, Smiley, Shah, Borova, Langdon, Moussa, Beane, Huang, Routledge, and Wang}]{chen2021finqa}
Zhiyu Chen, Wenhu Chen, Charese Smiley, Sameena Shah, Iana Borova, Dylan Langdon, Reema Moussa, Matt Beane, Ting-Hao Huang, Bryan Routledge, and William~Yang Wang. 2021.
\newblock \href {https://doi.org/10.18653/v1/2021.emnlp-main.300} {{{F}in{QA}: A Dataset of Numerical Reasoning over Financial Data}}.
\newblock In \emph{Proceedings of the 2021 Conference on Empirical Methods in Natural Language Processing}, pages 3697--3711, Online and Punta Cana, Dominican Republic. Association for Computational Linguistics.

\bibitem[{Chen et~al.(2026)Chen, Wu, Jia, Gao, Fu, Zhang, and Hu}]{chen2026longbenchpro}
Ziyang Chen, Xing Wu, Junlong Jia, Chaochen Gao, Qi~Fu, Debing Zhang, and Songlin Hu. 2026.
\newblock \href {https://arxiv.org/abs/2601.02872} {{LongBench Pro: A More Realistic and Comprehensive Bilingual Long-Context Evaluation Benchmark}}.
\newblock \emph{arXiv preprint arXiv:2601.02872}.

\bibitem[{Clark et~al.(2019)Clark, Lee, Chang, Kwiatkowski, Collins, and Toutanova}]{clark2019boolq}
Christopher Clark, Kenton Lee, Ming-Wei Chang, Tom Kwiatkowski, Michael Collins, and Kristina Toutanova. 2019.
\newblock \href {https://doi.org/10.18653/v1/N19-1300} {{{B}ool{Q}: Exploring the Surprising Difficulty of Natural Yes/No Questions}}.
\newblock In \emph{Proceedings of the 2019 Conference of the {N}orth {A}merican Chapter of the {A}ssociation for {C}omputational {L}inguistics: Human Language Technologies, Volume 1 (Long and Short Papers)}, pages 2924--2936, Minneapolis, Minnesota. Association for Computational Linguistics.

\bibitem[{Clark et~al.(2018)Clark, Cowhey, Etzioni, Khot, Sabharwal, Schoenick, and Tafjord}]{clark2018arc}
Peter Clark, Isaac Cowhey, Oren Etzioni, Tushar Khot, Ashish Sabharwal, Carissa Schoenick, and Oyvind Tafjord. 2018.
\newblock {Think you have Solved Question Answering? Try ARC, the AI2 Reasoning Challenge}.
\newblock \emph{arXiv preprint arXiv:1803.05457}.

\bibitem[{Clark et~al.(2020)Clark, Tafjord, and Richardson}]{clark2020ruletaker}
Peter Clark, Oyvind Tafjord, and Kyle Richardson. 2020.
\newblock \href {https://doi.org/10.24963/ijcai.2020/537} {{Transformers as Soft Reasoners over Language}}.
\newblock In \emph{Proceedings of the Twenty-Ninth International Joint Conference on Artificial Intelligence, {IJCAI-20}}, pages 3882--3890. International Joint Conferences on Artificial Intelligence Organization.

\bibitem[{{Cloudflare}(2026)}]{cloudflare2026clef}
{Cloudflare}. 2026.
\newblock \href {https://blog.cloudflare.com/clef-faster-cheaper-multimodal/} {{Introducing Clef-omni with Full Multimodality, plus a Faster Clef and a Cheaper Clef-flash}}.

\bibitem[{Conneau et~al.(2018)Conneau, Rinott, Lample, Williams, Bowman, Schwenk, and Stoyanov}]{conneau2018xnli}
Alexis Conneau, Ruty Rinott, Guillaume Lample, Adina Williams, Samuel~R. Bowman, Holger Schwenk, and Veselin Stoyanov. 2018.
\newblock \href {https://doi.org/10.18653/v1/D18-1269} {{{XNLI}: Evaluating Cross-lingual Sentence Representations}}.
\newblock In \emph{Proceedings of the 2018 Conference on Empirical Methods in Natural Language Processing}, pages 2475--2485, Brussels, Belgium. Association for Computational Linguistics.

\bibitem[{{DeepSeek-AI}(2026)}]{deepseekai2026deepseekv4}
{DeepSeek-AI}. 2026.
\newblock {DeepSeek-V4}: Towards highly efficient million-token context intelligence.
\newblock \emph{arXiv preprint arXiv:2606.19348}.

\bibitem[{Demszky et~al.(2020)Demszky, Movshovitz-Attias, Ko, Cowen, Nemade, and Ravi}]{demszky2020goemotions}
Dorottya Demszky, Dana Movshovitz-Attias, Jeongwoo Ko, Alan Cowen, Gaurav Nemade, and Sujith Ravi. 2020.
\newblock \href {https://doi.org/10.18653/v1/2020.acl-main.372} {{{G}o{E}motions: A Dataset of Fine-Grained Emotions}}.
\newblock In \emph{Proceedings of the 58th Annual Meeting of the {A}ssociation for {C}omputational {L}inguistics}, pages 4040--4054, Online. Association for Computational Linguistics.

\bibitem[{Deu{\ss}er et~al.(2026)Deu{\ss}er, Sparrenberg, and Sifa}]{deusser2026evaluating}
Tobias Deu{\ss}er, Lorenz Sparrenberg, and Rafet Sifa. 2026.
\newblock \href {https://arxiv.org/abs/2609.37647v1} {{Evaluating and Benchmarking the System One Model Jev}}.
\newblock \emph{arXiv preprint arXiv:2609.37647}.

\bibitem[{Duan et~al.(2026)Duan, Lin, Wang, Xiang, Wu, Zhang, Yan, Wang, and Wei}]{duan2026jeval}
Feiyu Duan, Jiayu Lin, Jia Wang, Jun Xiang, Jialiang Wu, Xinnong Zhang, Hanqi Yan, Siyuan Wang, and Zhongyu Wei. 2026.
\newblock \href {https://arxiv.org/abs/2610.03935} {{General Decision Models: Benchmarking and Insights Beyond Jev}}.
\newblock \emph{arXiv preprint arXiv:2610.03935}.

\bibitem[{Farn and Shin(2023)}]{farn2023tooltalk}
Nicholas Farn and Richard Shin. 2023.
\newblock \href {https://arxiv.org/abs/2311.10775} {{ToolTalk: Evaluating Tool-Usage in a Conversational Setting}}.
\newblock \emph{arXiv preprint arXiv:2311.10775}.

\bibitem[{FitzGerald et~al.(2023)FitzGerald, Hench, Peris, Mackie, Rottmann, Sanchez, Nash, Urbach, Kakarala, Singh, Ranganath, Crist, Britan, Leeuwis, Tur, and Natarajan}]{fitzgerald2023massive}
Jack FitzGerald, Christopher Hench, Charith Peris, Scott Mackie, Kay Rottmann, Ana Sanchez, Aaron Nash, Liam Urbach, Vishesh Kakarala, Richa Singh, Swetha Ranganath, Laurie Crist, Misha Britan, Wouter Leeuwis, Gokhan Tur, and Prem Natarajan. 2023.
\newblock \href {https://doi.org/10.18653/v1/2023.acl-long.235} {{{MASSIVE}: A 1{M}-Example Multilingual Natural Language Understanding Dataset with 51 Typologically-Diverse Languages}}.
\newblock In \emph{Proceedings of the 61st Annual Meeting of the {A}ssociation for {C}omputational {L}inguistics (Volume 1: Long Papers)}, pages 4277--4302, Toronto, Canada. Association for Computational Linguistics.

\bibitem[{{Gemma Team}(2026)}]{gemmateam2026gemma4}
{Gemma Team}. 2026.
\newblock {Gemma 4} technical report.
\newblock \emph{arXiv preprint arXiv:2607.02770}.

\bibitem[{{Google DeepMind}(2026)}]{googledeepmind2026gemini38flash}
{Google DeepMind}. 2026.
\newblock \href {https://deepmind.google/models/model-cards/gemini-3-8-flash/} {{Gemini 3.8 Flash Model Card}}.

\bibitem[{Gu et~al.(2024)Gu, Roziere, Leather, Solar-Lezama, Synnaeve, and Wang}]{gu2024cruxeval}
Alex Gu, Baptiste Roziere, Hugh~James Leather, Armando Solar-Lezama, Gabriel Synnaeve, and Sida Wang. 2024.
\newblock \href {https://proceedings.mlr.press/v235/gu24c.html} {{{CRUXE}val: A Benchmark for Code Reasoning, Understanding and Execution}}.
\newblock In \emph{Proceedings of the 41st International Conference on Machine Learning}, volume 235, pages 16568--16621. PMLR.

\bibitem[{Guo et~al.(2017)Guo, Pleiss, Sun, and Weinberger}]{guo2017calibration}
Chuan Guo, Geoff Pleiss, Yu~Sun, and Kilian~Q. Weinberger. 2017.
\newblock \href {https://proceedings.mlr.press/v70/guo17a.html} {{On Calibration of Modern Neural Networks}}.
\newblock In \emph{Proceedings of the 34th International Conference on Machine Learning}, volume~70 of \emph{Proceedings of Machine Learning Research}, pages 1321--1330, Sydney, Australia. PMLR.

\bibitem[{{H2O.ai}(2026)}]{h2oai2026lightning}
{H2O.ai}. 2026.
\newblock \href {https://huggingface.co/h2oai/h2o-lightning-4b} {{H2O-Lightning-4B}}.

\bibitem[{Hendrycks et~al.(2021)Hendrycks, Burns, Basart, Zou, Mazeika, Song, and Steinhardt}]{hendrycks2021mmlu}
Dan Hendrycks, Collin Burns, Steven Basart, Andy Zou, Mantas Mazeika, Dawn Song, and Jacob Steinhardt. 2021.
\newblock \href {https://openreview.net/forum?id=d7KBjmI3GmQ} {{Measuring Massive Multitask Language Understanding}}.
\newblock In \emph{International Conference on Learning Representations}, Online.

\bibitem[{Huang et~al.(2019)Huang, Le~Bras, Bhagavatula, and Choi}]{huang2019cosmosqa}
Lifu Huang, Ronan Le~Bras, Chandra Bhagavatula, and Yejin Choi. 2019.
\newblock \href {https://doi.org/10.18653/v1/D19-1243} {{Cosmos {QA}: Machine Reading Comprehension with Contextual Commonsense Reasoning}}.
\newblock In \emph{Proceedings of the 2019 Conference on Empirical Methods in Natural Language Processing and the 9th International Joint Conference on Natural Language Processing ({EMNLP}-{IJCNLP})}, pages 2391--2401, Hong Kong, China. Association for Computational Linguistics.

\bibitem[{Jeong et~al.(2024)Jeong, Baek, Cho, Hwang, and Park}]{jeong2024adaptive}
Soyeong Jeong, Jinheon Baek, Sukmin Cho, Sung~Ju Hwang, and Jong Park. 2024.
\newblock \href {https://doi.org/10.18653/v1/2024.naacl-long.389} {{Adaptive-{RAG}: Learning to Adapt Retrieval-Augmented Large Language Models through Question Complexity}}.
\newblock In \emph{Proceedings of the 2024 Conference of the {N}orth {A}merican Chapter of the {A}ssociation for {C}omputational {L}inguistics: Human Language Technologies (Volume 1: Long Papers)}, pages 7036--7050, Mexico City, Mexico. Association for Computational Linguistics.

\bibitem[{Jiang et~al.(2020)Jiang, Bordia, Zhong, Dognin, Singh, and Bansal}]{jiang2020hover}
Yichen Jiang, Shikha Bordia, Zheng Zhong, Charles Dognin, Maneesh Singh, and Mohit Bansal. 2020.
\newblock \href {https://doi.org/10.18653/v1/2020.findings-emnlp.309} {{{H}o{V}er: A Dataset for Many-Hop Fact Extraction And Claim Verification}}.
\newblock In \emph{Findings of the {A}ssociation for {C}omputational {L}inguistics: {EMNLP} 2020}, pages 3441--3460, Online. Association for Computational Linguistics.

\bibitem[{Khot et~al.(2020)Khot, Clark, Guerquin, Jansen, and Sabharwal}]{khot2020qasc}
Tushar Khot, Peter Clark, Michal Guerquin, Peter Jansen, and Ashish Sabharwal. 2020.
\newblock \href {https://doi.org/10.1609/aaai.v34i05.6319} {{QASC: A Dataset for Question Answering via Sentence Composition}}.
\newblock In \emph{Proceedings of the AAAI Conference on Artificial Intelligence}, volume~34, pages 8082--8090.

\bibitem[{Koreeda and Manning(2021)}]{koreeda2021contractnli}
Yuta Koreeda and Christopher Manning. 2021.
\newblock \href {https://doi.org/10.18653/v1/2021.findings-emnlp.164} {{{C}ontract{NLI}: A Dataset for Document-level Natural Language Inference for Contracts}}.
\newblock In \emph{Findings of the {A}ssociation for {C}omputational {L}inguistics: {EMNLP} 2021}, pages 1907--1919, Punta Cana, Dominican Republic. Association for Computational Linguistics.

\bibitem[{Lee(2026)}]{lee2026semif}
Theodore Lee. 2026.
\newblock \href {https://github.com/TheoLeeCJ/SemIf-OpenJev} {{SemIf-OpenJev}}.

\bibitem[{Li et~al.(2023)Li, Hui, Qu, Yang, Li, Li, Wang, Qin, Geng, Huo, Zhou, Ma, Li, Chang, Huang, Cheng, and Li}]{li2023bird}
Jinyang Li, Binyuan Hui, Ge~Qu, Jiaxi Yang, Binhua Li, Bowen Li, Bailin Wang, Bowen Qin, Ruiying Geng, Nan Huo, Xuanhe Zhou, Chenhao Ma, Guoliang Li, Kevin C.~C. Chang, Fei Huang, Reynold Cheng, and Yongbin Li. 2023.
\newblock \href {https://doi.org/10.52202/075280-1835} {{Can LLM Already Serve as A Database Interface? A BIg Bench for Large-Scale Database Grounded Text-to-SQLs}}.
\newblock In \emph{Advances in Neural Information Processing Systems}, volume~36, pages 42330--42357, New Orleans, United States. Curran Associates, Inc.

\bibitem[{Li et~al.(2026)Li, Liu, Chen, You, Di, He, Zheng, Choe, Sun, Wang, Tao, Li, Zhao, Geng, Wu, Zhou, Chen, Xing, Li, Zeng, Wang, Wang, Chaim, Jiang, Shen, Kong, Liu, Wang, Liu, Li, Lan, Lin, Ye, He, Li, Zhang, Gao, Li, Ma, Jing, Wang, Li, Xue, Lyu, He, Tian, Wu, Wang, Gao, Chen, Liu, Cheng, Bao, Tong, Xu, Zhuo, Ye, Qi, Li, Liao, Tan, Shi, Tang, Tankasala, Yuan, Qian, Tu, Wang, Sun, Wang, Taylor, Yang, Guan, Dong, Zhang, Dillmann, Lee, and Song}]{li2026skillsbench}
Xiangyi Li, Yimin Liu, Wenbo Chen, Bingran You, Zonglin Di, Yifeng He, Shenghan Zheng, Kyoung~Whan Choe, Jiankai Sun, Shuyi Wang, Chujun Tao, Binxu Li, Xuandong Zhao, Hejia Geng, Xiaojun Wu, Junwei Zhou, Xiaokun Chen, Hanwen Xing, Yubo Li, and 59 others. 2026.
\newblock \href {https://arxiv.org/abs/2602.12670} {{SkillsBench: Benchmarking How Well Agent Skills Work Across Diverse Tasks}}.
\newblock \emph{arXiv preprint arXiv:2602.12670}.

\bibitem[{Li et~al.(2022)Li, Choi, Chung, Kushman, Schrittwieser, Leblond, Eccles, Keeling, Gimeno, Dal~Lago, Hubert, Choy, de~Masson~d'Autume, Babuschkin, Chen, Huang, Welbl, Gowal, Cherepanov, Molloy, Mankowitz, Robson, Kohli, de~Freitas, Kavukcuoglu, and Vinyals}]{li2022alphacode}
Yujia Li, David Choi, Junyoung Chung, Nate Kushman, Julian Schrittwieser, R{\'e}mi Leblond, Tom Eccles, James Keeling, Felix Gimeno, Agustin Dal~Lago, Thomas Hubert, Peter Choy, Cyprien de~Masson~d'Autume, Igor Babuschkin, Xinyun Chen, Po-Sen Huang, Johannes Welbl, Sven Gowal, Alexey Cherepanov, and 7 others. 2022.
\newblock \href {https://doi.org/10.1126/science.abq1158} {{Competition-level code generation with AlphaCode}}.
\newblock \emph{Science}, 378(6624):1092--1097.

\bibitem[{Lin et~al.(2023)Lin, Wang, Tong, Wang, Guo, Wang, and Shang}]{lin2023toxicchat}
Zi~Lin, Zihan Wang, Yongqi Tong, Yangkun Wang, Yuxin Guo, Yujia Wang, and Jingbo Shang. 2023.
\newblock \href {https://doi.org/10.18653/v1/2023.findings-emnlp.311} {{{T}oxic{C}hat: Unveiling Hidden Challenges of Toxicity Detection in Real-World User-{AI} Conversation}}.
\newblock In \emph{Findings of the {A}ssociation for {C}omputational {L}inguistics: {EMNLP} 2023}, pages 4694--4702, Singapore. Association for Computational Linguistics.

\bibitem[{Ling et~al.(2026)Ling, Xue, and Ye}]{ling2026jev}
Guoming Ling, Muen Xue, and Zijian Ye. 2026.
\newblock \href {https://arxiv.org/abs/2609.30216} {{Jev in the Wild: A Data-Driven Analysis of the Jev Model's Functionality, Applications and Ecosystem}}.
\newblock \emph{arXiv preprint arXiv:2609.30216}.

\bibitem[{{Liquid AI}(2026)}]{liquidai2026d1}
{Liquid AI}. 2026.
\newblock \href {https://www.liquid.ai/blog/d1-ultra} {{d1-ultra: Introducing the World's Best Decision Model}}.

\bibitem[{Liu et~al.(2026)Liu, Li, Du, Wu, Gui, Kuang, and Sun}]{liu2026graphskills}
Dawei Liu, Zongxia Li, Hongyang Du, Xiyang Wu, Shihang Gui, Yongbei Kuang, and Lichao Sun. 2026.
\newblock \href {https://arxiv.org/abs/2604.05333} {{Graph of Skills: Dependency-Aware Structural Retrieval for Massive Agent Skills}}.
\newblock \emph{arXiv preprint arXiv:2604.05333}.

\bibitem[{Liu et~al.(2023{\natexlab{a}})Liu, Liu, Cui, Teng, Duan, Zhou, and Zhang}]{liu2023logiqa}
Hanmeng Liu, Jian Liu, Leyang Cui, Zhiyang Teng, Nan Duan, Ming Zhou, and Yue Zhang. 2023{\natexlab{a}}.
\newblock \href {https://doi.org/10.1109/TASLP.2023.3293046} {{LogiQA 2.0---An Improved Dataset for Logical Reasoning in Natural Language Understanding}}.
\newblock \emph{IEEE/ACM Transactions on Audio, Speech, and Language Processing}, 31:2947--2962.

\bibitem[{Liu et~al.(2023{\natexlab{b}})Liu, Xia, Wang, and Zhang}]{liu2023evalplus}
Jiawei Liu, Chunqiu~Steven Xia, Yuyao Wang, and Lingming Zhang. 2023{\natexlab{b}}.
\newblock \href {https://openreview.net/forum?id=1qvx610Cu7} {{Is Your Code Generated by Chat{GPT} Really Correct? Rigorous Evaluation of Large Language Models for Code Generation}}.
\newblock In \emph{Advances in Neural Information Processing Systems}.

\bibitem[{Liu et~al.(2024)Liu, Yu, Zhang, Xu, Lei, Lai, Gu, Ding, Men, Yang, Zhang, Deng, Zeng, Du, Zhang, Shen, Zhang, Su, Sun, Huang, Dong, and Tang}]{liu2024agentbench}
Xiao Liu, Hao Yu, Hanchen Zhang, Yifan Xu, Xuanyu Lei, Hanyu Lai, Yu~Gu, Hangliang Ding, Kaiwen Men, Kejuan Yang, Shudan Zhang, Xiang Deng, Aohan Zeng, Zhengxiao Du, Chenhui Zhang, Sheng Shen, Tianjun Zhang, Yu~Su, Huan Sun, and 3 others. 2024.
\newblock \href {https://proceedings.iclr.cc/paper_files/paper/2024/hash/e9df36b21ff4ee211a8b71ee8b7e9f57-Abstract-Conference.html} {{AgentBench: Evaluating LLMs as Agents}}.
\newblock In \emph{International Conference on Learning Representations}, pages 52989--53046, Vienna, Austria.

\bibitem[{Liu et~al.(2025)Liu, Yao, Min, Cao, Hou, and Li}]{liu2025rmbench}
Yantao Liu, Zijun Yao, Rui Min, Yixin Cao, Lei Hou, and Juanzi Li. 2025.
\newblock {RM-Bench: Benchmarking Reward Models of Language Models with Subtlety and Style}.
\newblock In \emph{International Conference on Learning Representations}.

\bibitem[{Malik et~al.(2026)Malik, Pyatkin, Land, Morrison, Smith, Hajishirzi, and Lambert}]{malik2026rewardbench}
Saumya Malik, Valentina Pyatkin, Sander Land, Jacob Morrison, Noah~A. Smith, Hannaneh Hajishirzi, and Nathan Lambert. 2026.
\newblock \href {https://openreview.net/forum?id=fb0G86Dewb} {{RewardBench 2: Advancing Reward Model Evaluation}}.
\newblock In \emph{International Conference on Learning Representations}.

\bibitem[{Marosi(2026)}]{marosi2026decider}
Mark Marosi. 2026.
\newblock \href {https://huggingface.co/Mapika/decider-4b} {{decider-4b: Typed Decisions with Calibrated Probabilities in One Forward Pass}}.

\bibitem[{Metsis et~al.(2006)Metsis, Androutsopoulos, and Paliouras}]{metsis2006spam}
Vangelis Metsis, Ion Androutsopoulos, and Georgios Paliouras. 2006.
\newblock \href {https://www2.aueb.gr/users/ion/data/enron-spam/} {{Spam Filtering with Naive Bayes---Which Naive Bayes?}}
\newblock In \emph{Proceedings of the Third Conference on Email and Anti-Spam}, Mountain View, California, USA.

\bibitem[{{Microsoft}(2026)}]{microsoft2026decision}
{Microsoft}. 2026.
\newblock \href {https://commandline.microsoft.com/microsoft-decision-1-model-foundry/} {{Introducing Microsoft-Decision-1, Our Model for Fast Decision-Making}}.

\bibitem[{Mihaylov et~al.(2018)Mihaylov, Clark, Khot, and Sabharwal}]{mihaylov2018openbookqa}
Todor Mihaylov, Peter Clark, Tushar Khot, and Ashish Sabharwal. 2018.
\newblock \href {https://doi.org/10.18653/v1/D18-1260} {{Can a Suit of Armor Conduct Electricity? A New Dataset for Open Book Question Answering}}.
\newblock In \emph{Proceedings of the 2018 Conference on Empirical Methods in Natural Language Processing}, pages 2381--2391, Brussels, Belgium. Association for Computational Linguistics.

\bibitem[{Moore(2026)}]{moore2026toolrisk}
Mike Moore. 2026.
\newblock \href {https://github.com/themsquared/jev-benchmark} {{Jev Benchmark: Agent Tool-Call Risk}}.

\bibitem[{Navarro(2026)}]{navarro2026reflexbench}
Carlos Navarro. 2026.
\newblock \href {https://github.com/brida-ai/reflexbench} {{ReflexBench: Benchmarking System One and Typed Decision Models}}.

\bibitem[{Nguyen(2026)}]{nguyen2026quyet}
Chinh Nguyen. 2026.
\newblock \href {https://huggingface.co/chinhnc/Quyet-1.0-Large} {{Quyet-1.0-Large}}.

\bibitem[{{OpenAI}(2026{\natexlab{a}})}]{openai2026decisions}
{OpenAI}. 2026{\natexlab{a}}.
\newblock \href {https://developers.openai.com/api/docs/guides/decisions} {{Decisions}}.

\bibitem[{{OpenAI}(2026{\natexlab{b}})}]{openai2026gpt6astra}
{OpenAI}. 2026{\natexlab{b}}.
\newblock \href {https://developers.openai.com/api/docs/models/gpt-6-astra} {{GPT-6 Astra}}.

\bibitem[{Oren et~al.(2024)Oren, Meister, Chatterji, Ladhak, and Hashimoto}]{oren2024contamination}
Yonatan Oren, Nicole Meister, Niladri Chatterji, Faisal Ladhak, and Tatsunori~B. Hashimoto. 2024.
\newblock \href {https://proceedings.iclr.cc/paper_files/paper/2024/hash/46e624c244cff669223d488defd4e835-Abstract-Conference.html} {{Proving Test Set Contamination in Black-Box Language Models}}.
\newblock In \emph{International Conference on Learning Representations}, Vienna, Austria.

\bibitem[{Pang et~al.(2022)Pang, Parrish, Joshi, Nangia, Phang, Chen, Padmakumar, Ma, Thompson, He, and Bowman}]{pang2022quality}
Richard~Yuanzhe Pang, Alicia Parrish, Nitish Joshi, Nikita Nangia, Jason Phang, Angelica Chen, Vishakh Padmakumar, Johnny Ma, Jana Thompson, He~He, and Samuel~R. Bowman. 2022.
\newblock \href {https://doi.org/10.18653/v1/2022.naacl-main.391} {{QuALITY: Question Answering with Long Input Texts, Yes!}}
\newblock In \emph{Proceedings of the 2022 Conference of the {N}orth {A}merican Chapter of the {A}ssociation for {C}omputational {L}inguistics: Human Language Technologies}, pages 5336--5358, Seattle, United States. Association for Computational Linguistics.

\bibitem[{Patil et~al.(2025)Patil, Mao, Yan, Ji, Suresh, Stoica, and Gonzalez}]{patil2025bfcl}
Shishir~G. Patil, Huanzhi Mao, Fanjia Yan, Charlie Cheng-Jie Ji, Vishnu Suresh, Ion Stoica, and Joseph~E. Gonzalez. 2025.
\newblock \href {https://proceedings.mlr.press/v267/patil25a.html} {{The Berkeley Function Calling Leaderboard (BFCL): From Tool Use to Agentic Evaluation of Large Language Models}}.
\newblock In \emph{Proceedings of the 42nd International Conference on Machine Learning}, volume 267 of \emph{Proceedings of Machine Learning Research}, pages 48371--48392. PMLR.

\bibitem[{{PelaAI}(2026)}]{pelaai2026knowline}
{PelaAI}. 2026.
\newblock \href {https://huggingface.co/PelaAI/KnowLine-31B-Gen1} {{KnowLine-31B-Gen1: A 31B Decision Model}}.

\bibitem[{{Perplexity AI}(2026)}]{perplexityai2026decider}
{Perplexity AI}. 2026.
\newblock \href {https://docs.perplexity.ai/docs/decisions/quickstart} {{Decisions API}}.

\bibitem[{Pyatkin et~al.(2025)Pyatkin, Malik, Graf, Ivison, Huang, Dasigi, Lambert, and Hajishirzi}]{pyatkin2025generalizing}
Valentina Pyatkin, Saumya Malik, Victoria Graf, Hamish Ivison, Shengyi Huang, Pradeep Dasigi, Nathan Lambert, and Hannaneh Hajishirzi. 2025.
\newblock \href {https://papers.nips.cc/paper_files/paper/2025/hash/46499a0622ecf568b72d17b61e45dbd5-Abstract-Datasets_and_Benchmarks_Track.html} {{Generalizing Verifiable Instruction Following}}.
\newblock In \emph{Advances in Neural Information Processing Systems}.

\bibitem[{Qin et~al.(2024)Qin, Liang, Ye, Zhu, Yan, Lu, Lin, Cong, Tang, Qian, Zhao, Hong, Tian, Xie, Zhou, Gerstein, Li, Liu, and Sun}]{qin2024toolllm}
Yujia Qin, Shihao Liang, Yining Ye, Kunlun Zhu, Lan Yan, Yaxi Lu, Yankai Lin, Xin Cong, Xiangru Tang, Bill Qian, Sihan Zhao, Lauren Hong, Runchu Tian, Ruobing Xie, Jie Zhou, Mark Gerstein, Dahai Li, Zhiyuan Liu, and Maosong Sun. 2024.
\newblock \href {https://proceedings.iclr.cc/paper_files/paper/2024/hash/28e50ee5b72e90b50e7196fde8ea260e-Abstract-Conference.html} {{ToolLLM: Facilitating Large Language Models to Master 16000+ Real-world APIs}}.
\newblock In \emph{International Conference on Learning Representations}, pages 9695--9717, Vienna, Austria.

\bibitem[{{Qwen Team}(2026{\natexlab{a}})}]{qwenteam2026qwen36}
{Qwen Team}. 2026{\natexlab{a}}.
\newblock \href {https://qwen.ai/blog?id=qwen3.6-35b-a3b} {{Qwen3.6-35B-A3B: Agentic Coding Power, Now Open to All}}.

\bibitem[{{Qwen Team}(2026{\natexlab{b}})}]{qwenteam2026qwen38max}
{Qwen Team}. 2026{\natexlab{b}}.
\newblock \href {https://qwen.ai/blog?id=qwen3.8} {{Qwen3.8-Max: A New Bar for Coding and Cowork}}.

\bibitem[{Roudaki(2026)}]{roudaki2026decisio}
Amin Roudaki. 2026.
\newblock \href {https://github.com/aminry/decisio} {{Decisio: Decision Inference}}.

\bibitem[{Sakaguchi et~al.(2020)Sakaguchi, Le~Bras, Bhagavatula, and Choi}]{sakaguchi2020winogrande}
Keisuke Sakaguchi, Ronan Le~Bras, Chandra Bhagavatula, and Yejin Choi. 2020.
\newblock {WinoGrande: An Adversarial Winograd Schema Challenge at Scale}.
\newblock In \emph{Proceedings of the AAAI Conference on Artificial Intelligence}.

\bibitem[{Sap et~al.(2019)Sap, Rashkin, Chen, Le~Bras, and Choi}]{sap2019socialiqa}
Maarten Sap, Hannah Rashkin, Derek Chen, Ronan Le~Bras, and Yejin Choi. 2019.
\newblock \href {https://doi.org/10.18653/v1/D19-1454} {{Social {IQ}a: Commonsense Reasoning about Social Interactions}}.
\newblock In \emph{Proceedings of the 2019 Conference on Empirical Methods in Natural Language Processing and the 9th International Joint Conference on Natural Language Processing ({EMNLP}-{IJCNLP})}, pages 4463--4473, Hong Kong, China. Association for Computational Linguistics.

\bibitem[{Saravia et~al.(2018)Saravia, Liu, Huang, Wu, and Chen}]{saravia2018emotion}
Elvis Saravia, Hsien-Chi~Toby Liu, Yen-Hao Huang, Junlin Wu, and Yi-Shin Chen. 2018.
\newblock \href {https://doi.org/10.18653/v1/D18-1404} {{{CARER}: Contextualized Affect Representations for Emotion Recognition}}.
\newblock In \emph{Proceedings of the 2018 Conference on Empirical Methods in Natural Language Processing}, pages 3687--3697, Brussels, Belgium. Association for Computational Linguistics.

\bibitem[{Serikbay(2026)}]{serikbay2026jevk5}
Alibi Serikbay. 2026.
\newblock \href {https://huggingface.co/alibiserikbay/JevK5} {{JevK5}}.

\bibitem[{Shen(2026)}]{shen2026openjev}
Junhao Shen. 2026.
\newblock \href {https://huggingface.co/shenjunhao/OpenJev-4B} {{OpenJev-4B}}.

\bibitem[{Shi et~al.(2025)Shi, Wang, Yan, Ren, Wang, Yin, and Ren}]{shi2025toolret}
Zhengliang Shi, Yuhan Wang, Lingyong Yan, Pengjie Ren, Shuaiqiang Wang, Dawei Yin, and Zhaochun Ren. 2025.
\newblock \href {https://doi.org/10.18653/v1/2025.findings-acl.1258} {{Retrieval Models Aren{'}t Tool-Savvy: Benchmarking Tool Retrieval for Large Language Models}}.
\newblock In \emph{Findings of the {A}ssociation for {C}omputational {L}inguistics: {ACL} 2025}, pages 24497--24524, Vienna, Austria. Association for Computational Linguistics.

\bibitem[{Solka(2026)}]{solka2026decisionbench}
Stephen Solka. 2026.
\newblock \href {https://github.com/Hanno-Labs/decision-bench} {{DecisionBench}}.

\bibitem[{Standhartinger(2026)}]{standhartinger2026jevbench}
Florian Standhartinger. 2026.
\newblock \href {https://github.com/fstandhartinger/jevbench} {{JevBench: Benchmark for AI Decision Models}}.

\bibitem[{{Strands Agents}(2026)}]{strandsagents2026decider}
{Strands Agents}. 2026.
\newblock \href {https://github.com/strands-labs/strands-decider} {{Strands Decider}}.

\bibitem[{Tan et~al.(2025)Tan, Zhuang, Montgomery, Tang, Cuadron, Wang, Popa, and Stoica}]{tan2025judgebench}
Sijun Tan, Siyuan Zhuang, Kyle Montgomery, Willian~Y. Tang, Alejandro Cuadron, Chenguang Wang, Raluca~Ada Popa, and Ion Stoica. 2025.
\newblock \href {https://arxiv.org/abs/2410.12784} {{JudgeBench: A Benchmark for Evaluating LLM-Based Judges}}.
\newblock In \emph{International Conference on Learning Representations}.

\bibitem[{Trivedi et~al.(2022)Trivedi, Balasubramanian, Khot, and Sabharwal}]{trivedi2022musique}
Harsh Trivedi, Niranjan Balasubramanian, Tushar Khot, and Ashish Sabharwal. 2022.
\newblock \href {https://doi.org/10.1162/tacl_a_00475} {{MuSiQue: Multihop Questions via Single-hop Question Composition}}.
\newblock \emph{Transactions of the Association for Computational Linguistics}, 10:539--554.

\bibitem[{{TypeSafe AI}(2026)}]{typesafeai2026jev}
{TypeSafe AI}. 2026.
\newblock \href {https://typesafe.ai/blog/introducing-system-one-models-and-jev} {{Introducing System One Models \& Jev}}.

\bibitem[{Vu et~al.(2024)Vu, Iyyer, Wang, Constant, Wei, Wei, Tar, Sung, Zhou, Le, and Luong}]{vu2024freshllms}
Tu~Vu, Mohit Iyyer, Xuezhi Wang, Noah Constant, Jerry Wei, Jason Wei, Chris Tar, Yun-Hsuan Sung, Denny Zhou, Quoc Le, and Thang Luong. 2024.
\newblock \href {https://doi.org/10.18653/v1/2024.findings-acl.813} {{{F}resh{LLM}s: Refreshing Large Language Models with Search Engine Augmentation}}.
\newblock In \emph{Findings of the {A}ssociation for {C}omputational {L}inguistics: {ACL} 2024}, pages 13697--13720, Bangkok, Thailand. Association for Computational Linguistics.

\bibitem[{Wang et~al.(2024{\natexlab{a}})Wang, Ma, Zhang, Ni, Chandra, Guo, Ren, Arulraj, He, Jiang, Li, Ku, Wang, Zhuang, Fan, Yue, and Chen}]{wang2024mmlupro}
Yubo Wang, Xueguang Ma, Ge~Zhang, Yuansheng Ni, Abhranil Chandra, Shiguang Guo, Weiming Ren, Aaran Arulraj, Xuan He, Ziyan Jiang, Tianle Li, Max Ku, Kai Wang, Alex Zhuang, Rongqi Fan, Xiang Yue, and Wenhu Chen. 2024{\natexlab{a}}.
\newblock \href {https://doi.org/10.52202/079017-3018} {{MMLU-Pro: A More Robust and Challenging Multi-Task Language Understanding Benchmark}}.
\newblock In \emph{Advances in Neural Information Processing Systems}, volume~37, pages 95266--95290. Curran Associates, Inc.

\bibitem[{Wang et~al.(2024{\natexlab{b}})Wang, Dong, Delalleau, Zeng, Shen, Egert, Zhang, Sreedhar, and Kuchaiev}]{wang2024helpsteer2}
Zhilin Wang, Yi~Dong, Olivier Delalleau, Jiaqi Zeng, Gerald Shen, Daniel Egert, Jimmy~J. Zhang, Makesh~Narsimhan Sreedhar, and Oleksii Kuchaiev. 2024{\natexlab{b}}.
\newblock \href {https://doi.org/10.52202/079017-0047} {{HelpSteer 2: Open-source dataset for training top-performing reward models}}.
\newblock In \emph{Advances in Neural Information Processing Systems}, volume~37, pages 1474--1501. Curran Associates, Inc.

\bibitem[{Welbl et~al.(2017)Welbl, Liu, and Gardner}]{welbl2017sciq}
Johannes Welbl, Nelson~F. Liu, and Matt Gardner. 2017.
\newblock \href {https://doi.org/10.18653/v1/W17-4413} {{Crowdsourcing Multiple Choice Science Questions}}.
\newblock In \emph{Proceedings of the 3rd Workshop on Noisy User-generated Text}, pages 94--106, Copenhagen, Denmark. Association for Computational Linguistics.

\bibitem[{{WeMM-JEV Team}(2026)}]{wemmjevteam2026wemm}
{WeMM-JEV Team}. 2026.
\newblock \href {https://huggingface.co/Rover00/WeMM-JEV} {{WeMM-JEV}}.

\bibitem[{Yang et~al.(2025)Yang, Li, Yang, Zhang, Hui, Zheng, Yu, Gao, Huang, Lv, Zheng, Liu, Zhou, Huang, Hu, Ge, Wei, Lin, Tang, Yang, Tu, Zhang, Yang, Yang, Zhou, Zhou, Lin, Dang, Bao, Yang, Yu, Deng, Li, Xue, Li, Zhang, Wang, Zhu, Men, Gao, Liu, Luo, Li, Tang, Yin, Ren, Wang, Zhang, Ren, Fan, Su, Zhang, Zhang, Wan, Liu, Wang, Cui, Zhang, Zhou, and Qiu}]{yang2025qwen3}
An~Yang, Anfeng Li, Baosong Yang, Beichen Zhang, Binyuan Hui, Bo~Zheng, Bowen Yu, Chang Gao, Chengen Huang, Chenxu Lv, Chujie Zheng, Dayiheng Liu, Fan Zhou, Fei Huang, Feng Hu, Hao Ge, Haoran Wei, Huan Lin, Jialong Tang, and 41 others. 2025.
\newblock {Qwen3 Technical Report}.
\newblock \emph{arXiv preprint arXiv:2505.09388}.

\bibitem[{Yong(2026)}]{yong2026decidebench}
Cho~Yin Yong. 2026.
\newblock \href {https://github.com/choyiny/decidebench} {{DecideBench: Accuracy, Cost and Latency of Decision Models}}.

\bibitem[{Yu et~al.(2018)Yu, Zhang, Yang, Yasunaga, Wang, Li, Ma, Li, Yao, Roman, Zhang, and Radev}]{yu2018spider}
Tao Yu, Rui Zhang, Kai Yang, Michihiro Yasunaga, Dongxu Wang, Zifan Li, James Ma, Irene Li, Qingning Yao, Shanelle Roman, Zilin Zhang, and Dragomir Radev. 2018.
\newblock \href {https://doi.org/10.18653/v1/D18-1425} {{{S}pider: A Large-Scale Human-Labeled Dataset for Complex and Cross-Domain Semantic Parsing and Text-to-{SQL} Task}}.
\newblock In \emph{Proceedings of the 2018 Conference on Empirical Methods in Natural Language Processing}, pages 3911--3921, Brussels, Belgium. Association for Computational Linguistics.

\bibitem[{{Z.ai}(2026)}]{zai2026glm52}
{Z.ai}. 2026.
\newblock \href {https://z.ai/blog/glm-5.2} {{GLM-5.2: Built for Long-Horizon Tasks}}.

\bibitem[{Zellers et~al.(2019)Zellers, Holtzman, Bisk, Farhadi, and Choi}]{zellers2019hellaswag}
Rowan Zellers, Ari Holtzman, Yonatan Bisk, Ali Farhadi, and Yejin Choi. 2019.
\newblock \href {https://doi.org/10.18653/v1/P19-1472} {{{H}ella{S}wag: Can a Machine Really Finish Your Sentence?}}
\newblock In \emph{Proceedings of the 57th Annual Meeting of the {A}ssociation for {C}omputational {L}inguistics}, pages 4791--4800, Florence, Italy. Association for Computational Linguistics.

\bibitem[{Zhang et~al.(2015)Zhang, Zhao, and LeCun}]{zhang2015character}
Xiang Zhang, Junbo~Jake Zhao, and Yann LeCun. 2015.
\newblock {Character-level Convolutional Networks for Text Classification}.
\newblock In \emph{Advances in Neural Information Processing Systems}.

\bibitem[{Zhang et~al.(2024)Zhang, Fang, and Chen}]{zhang2024retrievalqa}
Zihan Zhang, Meng Fang, and Ling Chen. 2024.
\newblock \href {https://doi.org/10.18653/v1/2024.findings-acl.415} {{{R}etrieval{QA}: Assessing Adaptive Retrieval-Augmented Generation for Short-form Open-Domain Question Answering}}.
\newblock In \emph{Findings of the {A}ssociation for {C}omputational {L}inguistics: {ACL} 2024}, pages 6963--6975, Bangkok, Thailand. Association for Computational Linguistics.

\bibitem[{Zhao et~al.(2025)Zhao, Huang, Lv, Cui, Sun, Mao, Zhang, Xin, Yin, Li, and Wei}]{zhao2025mmlucf}
Qihao Zhao, Yangyu Huang, Tengchao Lv, Lei Cui, Qinzheng Sun, Shaoguang Mao, Xin Zhang, Ying Xin, Qiufeng Yin, Scarlett Li, and Furu Wei. 2025.
\newblock \href {https://doi.org/10.18653/v1/2025.acl-long.656} {{MMLU-CF: A Contamination-free Multi-task Language Understanding Benchmark}}.
\newblock In \emph{Proceedings of the 63rd Annual Meeting of the Association for Computational Linguistics (Volume 1: Long Papers)}, pages 13371--13391, Vienna, Austria. Association for Computational Linguistics.

\bibitem[{Zhong et~al.(2025)Zhong, Du, Zhang, Hu, and Tang}]{zhong2025complexfuncbench}
Lucen Zhong, Zhengxiao Du, Xiaohan Zhang, Haiyi Hu, and Jie Tang. 2025.
\newblock \href {https://arxiv.org/abs/2501.10132} {{ComplexFuncBench: Exploring Multi-Step and Constrained Function Calling under Long-Context Scenario}}.
\newblock \emph{arXiv preprint arXiv:2501.10132}.

\bibitem[{Zhu et~al.(2021)Zhu, Lei, Huang, Wang, Zhang, Lv, Feng, and Chua}]{zhu2021tatqa}
Fengbin Zhu, Wenqiang Lei, Youcheng Huang, Chao Wang, Shuo Zhang, Jiancheng Lv, Fuli Feng, and Tat-Seng Chua. 2021.
\newblock \href {https://doi.org/10.18653/v1/2021.acl-long.254} {{{TAT}-{QA}: A Question Answering Benchmark on a Hybrid of Tabular and Textual Content in Finance}}.
\newblock In \emph{Proceedings of the 59th Annual Meeting of the {A}ssociation for {C}omputational {L}inguistics and the 11th International Joint Conference on Natural Language Processing (Volume 1: Long Papers)}, pages 3277--3287, Online. Association for Computational Linguistics.

\bibitem[{Zhu et~al.(2024)Zhu, Cheng, Peng, Li, Peng, Liu, Qiu, and Huang}]{zhu2024inference}
Qin Zhu, Qinyuan Cheng, Runyu Peng, Xiaonan Li, Ru~Peng, Tengxiao Liu, Xipeng Qiu, and Xuanjing Huang. 2024.
\newblock \href {https://doi.org/10.18653/v1/2024.findings-emnlp.532} {{Inference-Time Decontamination: Reusing Leaked Benchmarks for Large Language Model Evaluation}}.
\newblock In \emph{Findings of the Association for Computational Linguistics: EMNLP 2024}, pages 9113--9129, Miami, Florida, USA. Association for Computational Linguistics.

\bibitem[{Zhu et~al.(2026)Zhu, Xu, He, Chai, Hu, Zheng, Luo, Li, Zhang, and Cheng}]{zhu2026intern}
Tong Zhu, Zhe Xu, Zefeng He, Yunkai Chai, Xuyang Hu, Dehong Zheng, Yun Luo, Yafu Li, Qi~Zhang, and Yu~Cheng. 2026.
\newblock {Intern-Decision}: Efficient multimodal decision making with autoregressive masked language modeling.
\newblock \emph{arXiv preprint arXiv:2610.22131}.

\end{thebibliography}
